\pdfoutput=1
\documentclass[nonblindrev]{informs3} % arXiv version -- author info shown
\newcommand\numberthis{\addtocounter{equation}{1}\tag{\theequation}}

\newcommand{\QED}{\hfill\ensuremath{\square}}%

\newcommand\MyBox[2]{
  \fbox{\lower0.75cm
    \vbox to 2cm{\vfil
      \hbox to 2cm{\hspace*{\fill}\parbox{1.8cm}{#1\\#2}\hfil}
      \vfil}%
  }%
}

\newcommand\circled[1]{\tikz[baseline=(char.base)]{
            \node[shape=circle,draw,inner sep=2pt] (char) {#1};}}

\usepackage[numbers]{natbib}
 \bibpunct[, ]{(}{)}{,}{a}{}{,}%
 \def\bibfont{\small}%

\graphicspath{{./}{Figures/}{Plots/}}

\usepackage{pifont}
\usepackage{amsmath, amssymb}

\usepackage{bbold}
\usepackage{rotating,multirow}
\usepackage{booktabs}
\usepackage{lscape}
\usepackage{caption}
\usepackage{bm}

\usepackage{tikz}
\usetikzlibrary{calc}
\usepackage{float}
\usepackage{subfig}

\usepackage{centernot}

\usepackage{caption}
\usepackage{array}

\usepackage{xcolor}

\usepackage[linesnumbered,ruled]{algorithm2e}
\TheoremsNumberedThrough     % Preferred (Theorem 1, Lemma 1, Theorem 2)
\EquationsNumberedThrough    % Default: (1), (2), ...
\usepackage{soul}
\usepackage{enumitem}
\usepackage[normalem]{ulem}
\usepackage{float}

\usepackage{hyperref}

\hypersetup{colorlinks=true}
\hypersetup{citecolor=blue}

\usepackage{bookmark}

\newtheorem{counterexample}{Counterexample}

\definecolor{amber}{rgb}{1.0, 0.75, 0.0}

\begin{document}
%%%%%%%%%%%%%%%%

%\setcounter{page}{1}
%\MONTH{\Vec{X}xxxx}% (month or a similar seasonal id)
%\DOI{10.1287/xxxx.0000.0000}%

\RUNAUTHOR{She, Wang, Ayer, and Chhatwal}

\RUNTITLE{TLRF for COVID-19 Outbreak Detection}

% \TITLE{Bundling Information Goods of Decreasing Value}
\TITLE{Small Area Estimation of Case Growths for Timely COVID-19 Outbreak Detection}
%\RUNTITLE{COVID-19 GRF}
\ARTICLEAUTHORS{%
\AUTHOR{Zhaowei She\thanks{Equal contribution co-first authors, listed alphabetically}}
\AFF{Singapore Management University, \EMAIL{zwshe@smu.edu.sg}, \URL{}}
\AUTHOR{Zilong Wang\footnotemark[1]}
\AFF{Georgia Institute of Technology, \EMAIL{zwang937@gatech.edu}, \URL{}}
\AUTHOR{Turgay Ayer\footnote{Corresponding author}}
\AFF{Georgia Institute of Technology, \EMAIL{ayer@isye.gatech.edu}, \URL{}}
\AUTHOR{Jagpreet Chhatwal}
\AFF{Massachusetts General Hospital, Harvard Medical School, \EMAIL{jagchhatwal@mgh.harvard.edu}, \URL{}}
} % end of the block
%\ta{I believe OR now follows a blinded review policy}
% \textcolor{red}{
% Points to make
% Transparent Interpretable Model
% Nice connection to R-effective
% Good results quantification
% Perform Very Well Numerically
% Practical Decision Support tool @ covidsim
%  A paragraph or two
%  -> Dataset and Algorithm publicly available
% Stick with 1 doubling week for Confusion Matrix
% -> Followup in appendix
% -> All outbreaks vs No Outbreaks (TPR)
% -> None outbreaks vs False Labels (FPR)
% -> Key Point: Of all outbreaks detected how many could've been detected in advance and prevented?
% -> Followup: What about false positive: Our recall rate is reasonable (side point, not main point)
% }

%Despite the growing optimism, the threat of resurgent COVID-19 outbreaks remains. 

% The COVID-19 pandemic has had a profound impact on the global economy and has caused significant loss of life. It is crucial to quickly identify and control new outbreaks of COVID-19, and estimating the case growth rate is an important epidemiological factor in achieving this goal.

%\textcolor{red}{WZL:TURGAY REVISIT MSOM STYLE FORMAT}
%\par
%\noindent
%\input{backup/Abstract_MS.tex}

\ABSTRACT{The COVID-19 pandemic has exerted a profound impact on the global economy and continues to exact a significant toll on human lives. The COVID-19 case growth rate stands as a key epidemiological parameter to estimate and monitor for effective detection and containment of the resurgence of outbreaks. A fundamental challenge in growth rate estimation and hence outbreak detection is balancing the accuracy-speed tradeoff, where accuracy typically degrades with shorter fitting windows. In this paper, we provide a transfer learning framework, which we call \textit{Transfer Learning Random Forest} (\texttt{TLRF}), for an effective implementation of the random forests algorithm that balances this accuracy-speed tradeoff. 
Specifically, we develop an identification strategy that converts the growth rate estimation problem into a regression task, which enables effective transfer learning across space and time through random forests' adaptive weighting mechanism. As such, through adaptively choosing fitting window sizes based on relevant day-level and county-level features affecting the disease spread, \texttt{TLRF} can accurately estimate case growth rates for counties with small sample sizes.  Out-of-sample prediction analysis shows that the \texttt{TLRF} outperforms established growth rate estimation methods. Furthermore, we conducted a case study based on outbreak case data from the state of Colorado and showed that the \texttt{TLRF} could improve timely detections of outbreaks up to $224\%$ when compared to the decisions made by Colorado's Department of Health and Environment (CDPHE). To demonstrate practical implementation, we developed a publicly available outbreak detection tool that operated from September 2020 through March 2023, receiving substantial attention from policymakers across all 50 states.}

\KEYWORDS{COVID-19, Epidemiology, Machine Learning, Outbreak Detection, Transfer Learning}

\maketitle

\section{Introduction}

% PARAGRAPH 1 FILLER (herd immunity is impossible)

The COVID-19 pandemic has wreaked havoc on economies and human lives globally. Although vaccination efforts have effectively curtailed the spread of the virus, the specter of recurrent COVID-19 surges continues to loom, driven by factors such as waning immunity, vaccine hesitancy, the emergence of new variants, and the evolution of global policies. In China, for instance, the abrupt shift away from a ``zero-COVID" policy led to a swift and widespread Omicron outbreak, infecting an estimated 90\% of the population by the end of December 2022 \citep{Goldberg2023}. In light of these dynamic conditions, accurate and timely estimation of new case growth rates remains paramount to prepare for potential outbreaks.

% Even though the U.S. COVID-19 case incidence and deaths have declined since initiating mass vaccination campaigns, the threat of resurgent waves of COVID-19 still remains due to waning of immunity, vaccine hesitancy, and virus variants. A recent national opinion poll shows that half of the Americans are still concerned about potential surges of new outbreaks \citep{PBS2021}. To better prepare for potential outbreaks, it is of utmost importance to estimate the growth rates of new cases as accurately and as timely as possible.\par

%$\bm{r}$ = 0 means a constant number of new cases, and a negative $\bm{r}$ implies decreasing new cases

%, and 
% Formally, it models the exponential growth of new COVID-19 cases, i.e., $\bm{I_{s}}=e^{\bm{\bm{r s}}}$, where $\bm{I_{s}}$ is the number of new cases on the day $\bm{s}$, and $e$ is a mathematical constant related to exponential growth

%This rate, defined as the expected logarithmic ratio of two consecutive days' infected case numbers (See \S\ref{3.1} for its precise definition), serves as an indicator of the daily rate of change in infection numbers. Positive and negative exponential growth rates denote daily increases and decreases in new cases, respectively. 

% As an illustration, if the number of infected cases on day $t$ and $t-1$ follow the identical exponential growth pattern, expressed as $e^{rt}$ and $e^{r(t-1)}$ respectively, then the logarithmic ratio $r=\ln(\frac{e^{rt}}{e^{r(t-1)}})$ would represent the exponential growth rate (for an elaborate definition, please see Section \ref{3.1} of this paper).

Since the beginning of the pandemic, the COVID-19 exponential growth rate has been a key epidemiological parameter closely monitored by public health officials \citep{anderson2020reproduction,NPR2020}. This rate is defined as the expected natural logarithmic ratio of two consecutive days' infected case numbers (for a formal definition, please see \S\ref{3.1} of this paper). As an illustration, if the infected case numbers on days $t$ and $t-1$ are generated from the same exponential growth model and are expressed as $e^{rt}$ and $e^{r(t-1)}$, respectively, the natural logarithmic ratio would represent the exponential growth rate, i.e.,  $r=\ln(\frac{e^{rt}}{e^{r(t-1)}})$. The exponential growth rate provides a direct measure of change in infection numbers, with a larger exponential growth rate indicating faster growth. Additionally, as further discussed in \S\ref{2.3}, other important epidemiological parameters such as the basic reproduction number ($R_0$) are derived from the exponential growth rate and play key roles in outbreak management   \citep{lipsitch2003transmission,ma2020estimating,bertozzi2020challenges}. Lastly, and perhaps most importantly for our study purposes, accurately estimating and understanding spatial and temporal heterogeneity in exponential growth rates might help identify potential outbreaks. As such, the exponential growth rate is considered one of the most important ``model-free" parameters in formulating a well-informed epidemic outbreak response \citep{chowell2003sars}.\par

% To formulate a well-informed epidemic outbreak response, it is important to accurately estimate the instantaneous county-level exponential growth rate $\bm{r_{t,c}}$, which is specific to county $c$ on day $t$.\par

% For small λ values, λ approximates the daily percentage growth in cases. 

% It's modeled using the equation N(t) = constant * e^(λt), where N is the number of cases, t is time in days, and e is a mathematical constant related to exponential growth. A positive λ indicates increasing new cases daily, λ = 0 means a constant number of new cases, and a negative λ implies decreasing new cases. For small λ values, λ approximates the daily percentage growth in cases. The relationship between λ and the disease's reproduction ratio (R) can be approximated by R = e^(λT), where T is the mean generation time​1​.

% is indicative of an epidemic's potential to expand or die out, and it quantifies the potential impact of an outbreak. A higher rate suggests a more rapid disease spread. Furthermore, other essential epidemiological parameters, such as the basic reproduction number ($R_0$), can be derived from the exponential growth rate under common epidemiological models.

% This is because an epidemic outbreak expands if its exponential growth rate of incident cases is positive, otherwise it dies out. In addition, the potential impact of an outbreak is measured by its exponential growth rate, as higher rates indicate more rapid disease spread. 

Nevertheless, it remains a challenge to obtain accurate small area exponential growth rate estimates of epidemic outbreaks \citep[c.f.][]{ma2014estimating}, especially for diseases like COVID-19 where frequent policy-level (e.g., social distancing guidelines or school closures) and behavioral response (e.g., reduced mask use) changes quickly alter the disease propagation dynamics. In rapidly changing environments like the COVID-19 pandemic, selecting an appropriate fitting window involves balancing two competing objectives: detection timeliness (speed) and estimation reliability (accuracy). The fitting window—defined as the number of days of historical data used in the estimation process—directly influences both objectives. A longer fitting window provides more data points, potentially improving estimation accuracy through reduced variance. However, it also creates a delay in detecting significant changes in transmission patterns. Conversely, a shorter fitting window enables more rapid detection of emerging trends, such as potential outbreaks, but may sacrifice estimation precision due to limited data. This tradeoff is particularly critical in outbreak surveillance, where early detection must be weighed against the reliability of the signals being detected.

% it becomes a challenge to choose a fitting window size that balances the speed and accuracy tradeoff. That is, depending on the length of the fitting window (i.e., the number of days of past data used for estimation), there is a tradeoff between how quickly (speed) and how accurately (accuracy) the algorithm can detect changes in COVID-19 transmission patterns. For instance, when case numbers begin to rise, a shorter window allows the algorithm to classify this trend as a potential outbreak sooner. In contrast, a longer fitting window requires more days of consistent data before it can make the outbreak classification. In other words, while a longer fitting window is preferable as it reduces variance in estimation due to a larger sample size, a shorter window is appealing for early and timely detection of outbreaks while they are still in controllable stages.

Historically, this fitting window size is treated as a hyperparameter that is either directly specified by the user \citep[c.f.][]{MEL1} or determined by cross-validation \citep[c.f.][]{chowell2007comparative} in the epidemiology literature. However, the constantly changing policy interventions and the public's behavioral responses have presented new challenges with regard to window size selection. More precisely, considering the same window size at different stages of the epidemic or for different regions/counties, as is common in established literature and current practice, fails to account for temporal and spatial heterogeneity. These limitations call for a more adaptive approach to accurately estimate the exponential growth rates across various jurisdictions and over time at a fast pace.\par

% PARAGRAPH 5 WHAT DOES OUR TOOL ACCOMPLISHES
To address these challenges, we develop a transfer learning framework, which we refer to as Transfer Learning Random Forest (\texttt{TLRF}), that dynamically balances the speed-accuracy tradeoff by adjusting the window sizes over time and across space in case growth rate estimation. Our key insight is that, under certain identification assumptions (detailed in \S\ref{MO}), the growth rate estimation problem can be converted into a regression task, enabling effective transfer learning across space and time through random forests' adaptive weighting mechanism. In other words, \texttt{TLRF} is able to adaptively choose fitting window sizes based on relevant day-level and county-level features affecting the disease spread, such as social distancing policies and the United States' Centers for Disease Control and Prevention's (CDC) Social Vulnerability Index. Moreover, for counties with insufficient data, \texttt{TLRF} pools together all relevant COVID-19 case data from counties with similar features across the US via transfer learning.\par

% can accurately estimate case growth rates for counties with small sample sizes.

% Specifically, to capture the heterogeneity of COVID-19 growth rates over time and across space, \texttt{TLRF} chooses an adaptive fitting window size for each U.S. county on each day based on features that affect the disease spread

%First, from a methodological standpoint, we demonstrate how random forests, a classical machine learning algorithm, can be effectively applied to balance the challenging accuracy-speed tradeoff.  Second, as further discussed in \S\ref{2.2}, our proposed \texttt{TLRF} algorithm sheds new light on the ``small sample problem" in COVID-19 outbreak detection \citep[c.f.][]{brookmeyer2004monitoring} and reduces model uncertainty in COVID-19 outbreak management  \citep[c.f.][]{holmdahl2020wrong,bertozzi2020challenges}. 

Our study makes several key contributions to the relevant existing literature. First, we provide an identification strategy that transforms instantaneous growth rate estimation, a challenging epidemiological parameter inference problem, into a tractable regression task. Second, we demonstrate how this formulation enables standard random forests to naturally perform transfer learning across space and time, addressing the ``small sample problem" in COVID-19 outbreak detection \citep[c.f.][]{brookmeyer2004monitoring} and reducing model uncertainty in COVID-19 outbreak management  \citep[c.f.][]{holmdahl2020wrong,bertozzi2020challenges}. Third, through carefully designed experiments using county-level COVID-19 data throughout the pandemic, we show that \texttt{TLRF} substantially improves case growth rate projections and can reduce the median Mean Absolute Error (MAE), defined as the average of the absolute difference between the predictions and the ground truth samples, and Root Mean Squared Error (RMSE), defined as the square root of the average squared differences between the predictions and the ground truth samples, by at least $27.4\%$ and $37.7\%$ respectively (see $\S$\ref{subsec.flexible} for details). 
In addition, applying \texttt{TLRF} to Colorado state data reveals that CDPHE could have targeted counties with the most severe outbreaks in 58.79\% of their investigations, a significant improvement over their historical decisions (18.13\%). 
This marks a substantial 224\% increase in the Positive Predictive Value (PPV) of CDPHE's decisions (see $\S$\ref{CS} for details). 
To make our findings accessible to the policy-makers and the general public,  we have further developed an online interactive local-level outbreak detection tool (\url{www.covid19sim.org}). 
Since its official launch in September 2020, the outbreak detection tool has been used by more than 20,000 authentic users from all 50 states by April 2022. 
We have also been consulted by public health officials from several states to help them in their policy-making decisions through the support of the \texttt{TLRF} algorithm.

The rest of this paper is organized as follows. \S\ref{LR} reviews the related literature in epidemiology, machine learning, and health operations management on COVID-19 outbreak detection and management, with an emphasis on current challenges and our contribution to this literature. \S\ref{MO} develops a statistical model that identifies the instantaneous county-level COVID-19 exponential growth rate, while \S\ref{ME} presents the proposed machine learning algorithm to estimate this model. \S\ref{PE} evaluates the performance of this machine learning algorithm against other fitting window size selection methods via out-of-sample prediction accuracy in 7-day ahead COVID-19 incident case number forecasts. \S\ref{benchmark_TL} benchmarks our \texttt{TLRF} against other transfer learning algorithms to further validate its effectiveness. \S\ref{CS} conducts a case study to demonstrate how \texttt{TLRF} can assist health authorities in prioritizing limited resources in outbreak investigations. \S\ref{DST} discusses the decision support tool implementation of \texttt{TLRF}. \S\ref{CF} concludes this paper with suggestions for future work.

%\S\ref{GRF} highlights the advantage of the proposed \texttt{TLRF} algorithm over other implementations of random forest algorithms in instantaneous growth rate estimations and compares their empirical performance.

%\S\ref{PALT} explores the potential applications and constraints of TLRF beyond the task of county-level exponential growth rate estimations. 

%Particularly, to address these problems, we modify the classic GRF algorithm \citep[c.f.][]{athey2019generalized} and extend it to models with variable treatment intensity \citep[c.f.][]{angrist1995two}.

%during the major wave of U.S. COVID-19 outbreaks in winter 2020, we were also able to conduct a case study  in terms of COVID-19 outbreak detection

\section{Literature Review}\label{LR}

In the global effort to combat COVID-19, various epidemiological models were developed by researchers from a wide range of academic disciplines. These models can be categorized into three main classes, i.e., descriptive, predictive and prescriptive models \citep{liu2020paradigms}. \S\ref{2.1} reviews relevant literature in epidemiology, machine learning and health operations management based on this model classification. \S\ref{2.2} and \S\ref{2.3} discuss some current challenges in these modeling approaches, and explain how our study contributes to the fast-growing literature on COVID-19 outbreak detection and management.\par

%explains why the exponential growth rate of COVID-19 cases is an important parameter to estimate for all three modeling approaches

\subsection{COVID-19 Outbreak Detection and Management Studies}\label{2.1}

A straightforward way to analyze COVID-19 outbreaks is through descriptive models such as clustering and hot spot analysis. The goal of these models is to identify spatial distribution patterns of COVID-19 infection clusters, which have been widely used for data visualization purposes by popular media outlets \citep{MayoClinic2021hotspots,NYTimes2021hotspots}. 
More advanced clustering techniques such as generalized K-Means and point process analysis were also employed to detect COVID-19 outbreak clusters \citep{zhang2021generalized, ruiz2021health,hohl2020daily}.\par

Taking data visualization one step further, several studies, particularly in the machine learning community, have developed predictive models to predict when a COVID-19 outbreak was likely to happen and how severe the impact would be. For example, boosting algorithms like XGBoost were used to forecast whether a U.S. county would be affected by COVID-19, i.e., has at least one infected case \citep[c.f.][]{Mehta2020}. Besides, CUSUM-based change detection and deep learning based anomaly detection techniques were also applied to detect aberrations in COVID-19 case growth patterns \citep[c.f.][]{braca2021quickest,soldi2021quickest,karadayi2020unsupervised}. In addition, many short-term forecasting models are developed to predict the potential impacts (e.g., cases, hospitalizations and deaths) of COVID-19 outbreaks, with methods including deep learning \citep[c.f.][]{rodriguez2020deepcovid,wieczorek2020neural}, time-series analysis \citep[c.f.][]{awan2020prediction,tandon2020coronavirus} and support vector machine \citep[c.f.][]{singh2020prediction}, among others. Starting from April 2020, the U.S. Centers for Disease Control and Prevention (CDC) formally established a COVID-19 Forecast Hub to generate ensemble forecasts by combining predictions from multiple epidemiological methods \citep{ray2020ensemble}.\par %\ta{This citation appears a bit odd, pls correct.}

%\tdelete{ These prediction models provide early warning signals to policy makers and the public before the outbreaks are fully developed.}\par

% In general, to predict an epidemic outbreak, one needs to predict ``an increase, often sudden, in the number of cases of a disease above what is normally expected in that population in that area" \citep[c.f.][]{dicker2006principles}.

% \cite{chen2020allocation} solve for optimal vaccine allocation policy to contain COVID-19 outbreaks through a compartmental model.

In addition to descriptive and predictive models, prescriptive models are often used to guide operational decisions such as how to allocate medical resources (e.g., ventilators and vaccines) and when to implement non-pharmaceutical interventions (NPIs) (e.g., social distancing and mask mandate). Many of these prescriptive models in COVID-19 outbreak management are developed by healthcare operations management researchers. For example, \cite{mehrotra2020model} developed a stochastic optimization model to allocate ventilators in the national stockpile across states to address COVID-19 outbreaks. The compartmental modeling approach, often utilizing either the susceptible-infected-recovered (SIR) model or the susceptible-exposed-infectious-recovered (SEIR) model, has been employed to investigate various decision-making challenges, such as the formulation of targeted lockdown policies to curb COVID-19 outbreaks \citep[e.g.,][]{birge2022controlling,palmer2020optimal,camelo2021quantifying,li2023forecasting}. By explicitly modeling population transitions between different disease states (e.g., susceptible
(S), exposed (E), infected (I), and recovered (R)), these compartmental models have played pivotal roles in elucidating the dynamics of COVID-19, especially with respect to the implementation of targeted policies and the behavioral responses they elicit \citep[e.g.,][]{chen2020allocation,bai2021no,chen2023hospital}. Last but not least, the global pandemic also catalyzes collaborations between researchers and practitioners. In the past two years, various practice-based COVID-19 outbreak management models have been developed to inform policy and clinical level decisions such as targeted border screening, patient triage, ICU capacity and nurse staffing planning \citep[c.f.][]{bastani2021efficient, Betcheva2021,bertsimas2021predictions,kaplan2020om,shi2021operations}. 

%Lastly, we remark that this is only a non-exhaustive summary of the rapid-growing literature on COVID-19 outbreak management.

% % PARAGRAPH 6 Literature Review of Similar Work

% \subsection{\tdelete{Some Current Challenges and Our Contribution}}\label{2.2}

% \tdelete{This section discusses two current challenges in COVID-19 outbreak detection and management, and explains how this paper helps to address these challenges.}

\subsection{The Small Sample Problem}\label{2.2}
A distinctive challenge in epidemic outbreak detection is the small sample problem. 
Specifically, to effectively contain potential outbreaks in their early stages, an outbreak detection algorithm must provide early warning signals of these outbreaks before they fully develop, when their case numbers are still small \citep{brookmeyer2004monitoring}. 
However, when case numbers are small, accurately training statistical or machine learning models becomes a challenge due to the small sample size problem. 
Unless managed carefully, models tend to overfit in such cases resulting in poor out-of-sample prediction and the level of overfitting increases with the model complexity \citep{nakip2020curse}.\par

In ML, a common approach to address this small sample problem is through transfer learning \citep[c.f.][]{pan2009survey,zhuang2020comprehensive}. Specifically, transfer learning refers to a class of ML algorithms that pool relevant proxy data to improve their ML models' performance on target data. For example, \cite{bastani2021predicting} propose a LASSO-based estimator to incorporate a large sample of proxy data in ML models to improve their prediction performance on target data. Following a similar spirit, \cite{farias2019learning} develop a tensor completion model to improve recommender systems' performance on target customers through pooling response data from other proxy customers. Based on the successful application of transfer learning in various settings, it is natural to expect that a similar technique can be employed to address the small sample problem in epidemic outbreak detection.\par

However, existing transfer learning techniques are not directly applicable to COVID-19 outbreak detection to address the small sample problem. Specifically, existing transfer learning methods with COVID-19 applications, like \cite{wang2021machine} and those surveyed by \cite{mukherjee2022epidemic}, follow the parameter-transfer approach, where a pre-trained model is adapted across domains through shared parameters or priors (See Figure 2 by \cite{zhuang2020comprehensive} or Table 3 by \cite{pan2009survey} for a complete categorization of transfer learning methods). However, this approach is unsuitable for our problem because COVID-19 growth rates vary continuously across time and space, making parameter transfer ineffective. Instead of transferring parameters, \texttt{TLRF} uses a data-based transfer learning approach, leveraging case data and features from source counties and dates to estimate the growth rate for a target county and date. To the best of our knowledge, this is the first study that applies a data-based transfer learning framework to epidemiological parameter estimation, highlighting a novel application of transfer learning techniques in this context.\par

Our transfer learning framework (\texttt{TLRF}) provides the necessary transformations, identification assumptions, and empirical validation required for an effective application of transfer learning in COVID-19 outbreak detection. Specifically, \texttt{TLRF} presents a new implementation of random forest algorithms \citep[c.f.][]{breiman2001random} that allows for proper identification of the county-level instantaneous COVID-19 exponential growth rates. The necessity for this implementation stems from the limitations of the classical random forest implementation for parameter estimation, e.g., the generalized random forests (GRF) by \cite{athey2019generalized}, which cannot effectively utilize time (or treatment) dependent features, such as the duration of face mask mandates in a given county. As demonstrated in Appendix \ref{GRF}, the population overlap assumption, a fundamental assumption of the GRF implementation, is violated when features are time-dependent. Consequently, the county-level instantaneous COVID-19 exponential growth rate cannot be identified as a conditional average partial effect, and thus cannot be estimated by the classical implementation.

\subsection{Model Uncertainty in COVID-19 Outbreak Management}\label{2.3}

Another challenge in COVID-19 outbreak management comes from model uncertainty \citep{holmdahl2020wrong}. Specifically, prescriptive models to guide COVID-19 outbreak management, as reviewed in \S\ref{2.1}, require accurate estimates of key epidemiological parameters such as the basic reproduction number and the exponential growth rate. However, the fact that these parameter estimates tend to vary from one outbreak to another leads to high model uncertainty in COVID-19 outbreak management \citep{bertozzi2020challenges}.\par

%\tdelete{To reduce this uncertainty, many prescriptive papers choose to estimate these epidemiological parameters using their own data, instead of borrowing estimates ``directly from incomplete or unrepresentative data-sources" (c.f.  \cite{Betcheva2021,bertsimas2021predictions,kaplan2020om,shi2021operations})}.\par

Among these epidemiological parameters, the epidemic exponential growth rate is viewed as one of the most important ``model-free" parameters in/for epidemic outbreak detection and management \citep{chowell2003sars,anderson2020reproduction}. Specifically, the epidemic exponential growth rate can be estimated through pure phenomenological models that are independent of disease transmission assumptions \citep{ma2014estimating}. Furthermore, for epidemiological parameters whose estimation depends on certain modeling assumptions, a more accurate estimate of the epidemic exponential growth rate would lead to more accurate estimates of these ``model-based" parameters as well. For example, when a mechanistic transmission model (e.g., an SIR or SEIR model) is assumed, the basic reproduction number ($R_0$) is a function of the exponential growth rate. More precisely, in an SIR model where the infectious period is exponentially distributed with a mean $\frac{1}{\gamma}$, its basic reproduction number ($R_0$) is determined by the epidemic exponential growth rate (r) as $R_0=\frac{r}{\gamma}+1$. In an SEIR model where the infectious and latent periods are exponentially distributed with means $\frac{1}{\gamma}$ and $\frac{1}{\sigma}$ respectively, its basic reproduction number ($R_0$) is determined by the epidemic exponential growth rate (r) as $R_0=\frac{(r+\gamma)(r+\sigma)}{\gamma \sigma}$ \citep{ma2020estimating}. Therefore, reducing uncertainty in epidemic exponential growth rate estimation would help in COVID-19 outbreak management regardless of their disease transmission assumptions.\par

To this aim, this paper proposes a new algorithm to improve the estimation accuracy of exponential growth rates in COVID-19 outbreaks. Specifically, this algorithm addresses a long-standing open problem in epidemiology regarding the speed and accuracy tradeoff in epidemic exponential growth rate estimation, and demonstrates consistently higher out-of-sample prediction accuracy compared to existing estimation methods.\par

\section{Instantaneous Exponential Growth Model and Identification}\label{MO}

This section first constructs an instantaneous COVID-19 exponential growth model (\S\ref{3.1}). Subsequently, we present and discuss the identification assumptions of the instantaneous county-level exponential growth rate ${r_{t,c}}$ from this model (\S\ref{3.2}). Lastly, we discuss the practical implications of our identification framework and particularly its advantages over the traditional ordinary least squares (OLS) framework (\S\ref{IAD}).\par

Throughout the paper, random variables, indices, and functions are represented in bold (e.g., $\bm{\Vec{X}_{t,c}}$). Vectors are expressed using vector notation (e.g., $\Vec{X}_{t,c}$), while matrices are denoted in calligraphic fonts (e.g., $\mathcal{X}$). In addition, we denote the set of positive integers, i.e., ${1,2,3,\ldots}$, as $\mathbb{Z}^+$ \citep{Weisstein2023}. Lastly, abbreviated notations such as $\mathbb{E}[\ln(\bm{I^t_{s,c}}) |\Vec{X}_{t,c}]$  are employed to represent the conditional expectations of random variables given relevant features at county $c$ on day $t$, e.g., $\mathbb{E}[\ln(\bm{I^t_{s,c}}) |\bm{\Vec{X}_{t,c}}=\Vec{X}_{t,c}]$.\par

\subsection{Instantaneous Exponential Growth Model}\label{3.1}

To capture the heterogeneity of COVID-19 case growth patterns, we employ a flexible potential growth model for each county's daily log incident case number, defined as follows: $\forall s, t\in \mathbb{Z}^+$,
\begin{equation}\label{POt}
    \ln(\bm{I^t_{s,c}})=\alpha(\Vec{X}_{t,c})+ r(\Vec{X}_{t,c}){s}+ \bm{\varepsilon^t_{s,c}}.
\end{equation}
Here, $\ln(\bm{I^t_{s,c}})$ represents the potential log incident case number of county $c$ on day ${s}$ if COVID-19 growth in county $c$ were to follow the day $t$ spreading patterns consistently. $\alpha: \mathbb{R}^d \to \mathbb{R}$ and $ r: \mathbb{R}^d \to \mathbb{R}$ are deterministic functions. ${\Vec{X}_{t,c}}\in \mathbb{R}^d$ is a realization of the random feature vector $\bm{\vec{X}_{s,c}}: \Omega \to \mathbb{R}^d$ on day $s=t$, capturing relevant features of county $c$ on day $t$ that affect COVID-19 spread. The error term $\bm{\varepsilon_{s,c}}$ accounts for the random noise, e.g., omitted variables and measurement errors. Appendix \ref{apdA} provides a detailed epidemiological definition of incident case numbers, while Appendix \ref{apdC} lists the relevant features used in this study.\par

%\numberthis \label{diagonal}
 This potential outcome framework can be illustrated through the following matrix:
\[
\begin{bmatrix}
\circled{$\ln(\bm{I^1_{1,c}})$} & \ln(\bm{I^1_{2,c}}) & \ln(\bm{I^1_{3,c}}) & \cdots \\
\ln(\bm{I^2_{1,c}}) & \circled{$\ln(\bm{I^2_{2,c}})$} & \ln(\bm{I^2_{3,c}}) & \cdots \\
\ln(\bm{I^3_{1,c}}) & \ln(\bm{I^3_{2,c}}) & \circled{$\ln(\bm{I^3_{3,c}})$} & \cdots \\
\vdots & \vdots & \vdots & \ddots 
\end{bmatrix}. 
\]
The diagonal entries in this matrix represent the observed log incident case numbers of county $c$ on day ${s}$. To simplify notations, we denote these observed incident case numbers as $\{\bm{I_{s,c}}\}_{s\in \mathbb{Z}^+, c\in C}$, which follows the relation: $\forall s\in \mathbb{Z}^+$ and $\forall c\in C$,
$$\ln(\bm{I^s_{s,c}})=\ln(\bm{I_{s,c}}).$$
The off-diagonal entries, i.e., $\ln(\bm{I^t_{s,c}})$ where ${s}\neq t$, denote the potential log incident case number of county $c$ on day ${s}$ as if it were generated by day $t$'s growth model, which are counterfactual outcomes and not observable.\par

As a consequence of this modeling framework, the actual growth model for each county's daily log incident case number can be expressed as: $\forall s\in \mathbb{Z}^{+}$,
\begin{equation}\label{observed}
    \ln(\bm{I_{s,c}})=\alpha(\bm{\Vec{X}_{s,c}})+ r(\bm{\Vec{X}_{s,c}}){s}+ \bm{\varepsilon_{s,c}}.
\end{equation}
Here, $\bm{I_{s,c}}$ represents the actual incident case number in county $c$ on day $s$. The intercept $\alpha(\bm{\Vec{X}_{s,c}})$ and slope $r(\bm{\Vec{X}_{s,c}})$ are random coefficients, whose realizations on day $s=t$ correspond to the intercept $\alpha({\Vec{X}_{t,c}})$ and slope $r({\Vec{X}_{t,c}})$ in the day $t$ potential growth model (\ref{POt}), respectively. $\bm{\varepsilon_{s,c}}$ is a random error that coincides with the random error of each day $t$ potential growth model (\ref{POt}) on the day $s=t$: $\forall t\in \mathbb{Z}^{+}$, $\bm{\varepsilon_{t,c}}=\bm{\varepsilon^t_{t,c}}.$
In particular, this potential outcome framework allows the random error $\bm{\varepsilon_{s,c}}$ to correlate with the random feature vector. 

% the day $t$: $\bm{\varepsilon_{t,c}}=\bm{\varepsilon^t_{t,c}}$.

% , with $\alpha(\cdot)$ and $r(\cdot)$ being deterministic functions on $\mathbb{R}^f \times \mathbb{R}$. $\bm{\Vec{X}_{s,c}}$ is a random feature vector with support on $\mathbb{R}^f$, capturing time and county-level variations affecting COVID-19 spread. The error term $\bm{\varepsilon{s,c}}$ accounts for random noise, including omitted variables and measurement errors, and may correlate with $\bm{\Vec{X}_{s,c}}$. Appendix \ref{apdA} provides a detailed epidemiological definition of incident case numbers, while Appendix \ref{apdC} lists the relevant features used in this study.
% This specification essentially permits a unique data generating process for each data point, reflecting the reality that COVID-19 spreading patterns can vary day by day and county by county.

We are now ready to define the instantaneous county-level exponential growth and characterize conditions under which potential growth models (\ref{POt}) follow this pattern:  {adp:synthetic}
:\\
\textbf{Assumption 1} \textit{(Instantaneous Exponential Growth)} 
$$\mathbb{E}[\ln(\bm{I^t_{s,c}})|\Vec{X}_{t,c} ]=\alpha(\Vec{X}_{t,c})+ r(\Vec{X}_{t,c}){s}+\mathbb{E}[\bm{\varepsilon^t_{t,c}}|\Vec{X}_{t,c}] \ \ \forall s\in \mathbb{Z}^+.$$
Importantly, Assumption 1 pertains specifically to the exponential growth pattern of the potential growth model (\ref{POt}). It does not impose an exponential growth constraint on the actual growth model (\ref{observed}). As illustrated in Figure \ref{AssumptionExamples}, the actual growth model (\ref{observed}) exhibits only local exponential growth, which diverges from the global exponential behavior of the potential growth models (\ref{POt}). More precisely, we establish in Lemma \ref{LemmaPotential} that the potential growth model follows an exponential growth pattern, where the expected daily logarithm of incident case numbers is a linear function of time.
\begin{lemma}\label{LemmaPotential}
The slope term of the potential growth model (\ref{POt}) defines the instantaneous county-level exponential growth rate:
$${r_{t,c}}:=r(\Vec{X}_{t,c}),$$
whose estimand is
\begin{equation}\label{cape}
r(\Vec{X}_{t,c})=\mathbb{E}\left[\ln(\bm{I^t_{t,c}})-\ln(\bm{I^t_{t-1,c}}) \bigg| \Vec{X}_{t,c}\right]. 
%\ \ \forall s\in \mathbb{Z}^+.
% \ln(\bm{I^t_{s,c}})-\ln(\bm{I^t_{s-1,c}}) \ \ \forall s\in \mathbb{Z}^+.
\end{equation}
\end{lemma}
For the sake of brevity, all the proofs are included in Appendix \ref{proof}.\par

Notably, while violations of Assumption 1 compromise the interpretation of $r(\Vec{X}_{t,c})$ as an exponential growth rate (essential for epidemiological models like SIR/SEIR), our synthetic experiments show that forecast accuracy degrades modestly under misspecification. This distinction between parameter interpretability and predictive robustness is examined in detail in Appendix \ref{adp:synthetic}.

In addition, the functional form of (\ref{POt}) does not guarantee instantaneous exponential growth. In fact, any growth model $\mathbb{E}[\ln(\bm{I^t_{s,c}})|\Vec{X}_{t,c}]={f(s)}$ can be derived from (\ref{POt}) by setting $\mathbb{E}[\bm{\varepsilon^t_{s,c}}|\Vec{X}_{t,c}]={f(s)}-r(\Vec{X}_{t,c}){s}-\alpha(\Vec{X}_{t,c})$. In addition, as demonstrated in Counterexample \ref{Counter1} of Appendix \ref{CounterExamples}, our identification assumptions do not imply the error term $\bm{\varepsilon^t_{s,c}}$ to be invariant of $s$. The key insight of Assumption 1 is that while the error term may vary with $s$, its conditional expectation must be invariant to $s$:
$\forall s\in \mathbb{Z}^+$,
$$\mathbb{E}[\bm{\varepsilon^t_{s,c}}|\Vec{X}_{t,c}]=\mathbb{E}[\bm{\varepsilon^t_{t,c}}|\Vec{X}_{t,c}].$$
In other words, the quality of the fit of the instantaneous exponential growth model depends only on the day $t$ whose spreading patterns are extrapolated and not on the specific day $s$ for which the incident case numbers are computed. Lastly, we note that Assumption 1 does not imply $E[\bm{\varepsilon_{s,c}}]=0$, a necessary condition to identify the intercept term of the actual growth model (\ref{observed}).

\subsection{Model Identification}\label{3.2}

This section characterizes conditions that enable the identification of the instantaneous county-level exponential growth rate ${r_{t,c}}$. Specifically, according to its definition (\ref{cape}), identifying ${r_{t,c}}$ requires imputing the unobservable ``counterfactual" incident case number $\{\bm{I^{t}_{s,c}}\}_{s\neq t}$ from the observable actual $\bm{I^t_{t,c}}$. We show in Theorem 1 that, under the Instantaneous Exponential Growth Assumption (Assumption 1), imputing $\bm{I^{t}_{t-1,c}}$, instead of $\{\bm{I^{t}_{s,c}}\}_{s\neq t}$, suffices to achieve this identification.

% This section presents the identification assumptions of the instantaneous county-level exponential growth rate in Theorem \ref{T1}. Subsequently, we proceed to examine these assumptions, demonstrating that they are less restrictive and allow for more adaptive estimators compared to OLS assumptions.\par

% To overcome this challenge, assumptions must be made about the data-generating process of (\ref{observed}), particularly with regards to its relationship to the potential case growth models (\ref{POt}) and (\ref{POt-1}). These assumptions are discussed in detail after being presented in Theorem \ref{T1}.

%\textbf{Assumption 1} (Instantaneous Exponential Growth)\\ and\\

Our identification requires a certain ``overlap" assumption between the potential growth models (\ref{POt}):\\
\textbf{Assumption 2} \textit{(Overlap)} 
$$\mathbb{E}[\ln(\bm{I^t_{t-1,c}})|\Vec{X}_{t,c}]=\mathbb{E}[\ln(\bm{I^{t-1}_{t-1,c}})|{\Vec{X}_{t-1,c}}].$$
Assumption 2 requires that the potential growth models of $\ln(\bm{I^t_{s,c}})$ and $\ln(\bm{I^{t-1}_{s,c}})$ overlap on the day $t-1$ in expectation. In the absence of this assumption, only \textit{a single data point} of the potential growth model $\ln(\bm{I^t_{s,c}})$ can be observed from the data generated by the actual growth model (\ref{observed}). With a single observation, it is not possible to identify any parameters, including ${r_{t,c}}$, from the potential growth model $\ln(\bm{I^t_{s,c}})$. Figure \ref{AssumptionExamples} illustrates a possible trajectory of $\mathbb{E}[\ln(\bm{I^s_{s,c}})|{\Vec{X}_{s,c}}]$ in accordance with Assumptions 1 and 2, serving as an explanatory example. Theorem \ref{T1} demonstrates that ${r_{t,c}}$ is identified under these two assumptions.
\begin{theorem}\label{T1}
The county-level instantaneous exponential growth rate is identified as
\begin{equation}\label{fw}
r(\Vec{X}_{t,c})=\mathbb{E}[\ln(\bm{I^t_{t,c}})|\Vec{X}_{t,c}]-\mathbb{E}[\ln(\bm{I^{t-1}_{t-1,c}})|{\Vec{X}_{t-1,c}}].
\end{equation}
\end{theorem} 
\noindent
Notably, Theorem \ref{T1} cannot be obtained by directly plugging in Assumption 2 (Overlap). We provide a counterexample (Counterexample \ref{Counter2}) in Appendix \ref{CounterExamples} to illustrate that Theorem 1 requires additional assumptions beyond Assumption 2 to ensure proper identification of ${r_{t,c}}$.\par

\begin{figure}[h!]
	\centering
\begin{tikzpicture}
\draw[->, thick] (0,0)--(5,0) node[right] {$s$};
\draw[->, thick] (0,0)--(0,4) node [above] {$\mathbb{E}[\ln(\bm{I_{s,c}})|{\Vec{X}_{s,c}}]$};
% Series 1 (amber): dotted
\draw[->, amber, thick, dotted, domain=0:2, samples=100] plot(\x, {1 + 0.25*\x});
% Series 2 (blue): long dashes
\draw[->, blue, thick, dash pattern=on 8pt off 4pt, domain=1:3, samples=100] plot(\x, {0.5 + 0.75*\x});
% Series 3 (purple): dash-dot
\draw[->, purple, thick, dash pattern=on 6pt off 2pt on 2pt off 2pt, domain=2:4, samples=100] plot(\x, {1.8 + 0.1*\x});
% Series 4 (green): medium dashes
\draw[->, green, thick, dash pattern=on 4pt off 3pt, domain=3:5, samples=100] plot(\x, {3.3 - 0.4*\x});
% amber: filled circle
\filldraw[amber] (1,1.25) circle (3pt);
% blue: filled square
\fill[blue] ($(2,2)+(-3pt,-3pt)$) rectangle ($(2,2)+(3pt,3pt)$);
% purple: filled diamond
\fill[purple] ($(3,2.1)+(0,4pt)$) -- ($(3,2.1)+(4pt,0)$) -- ($(3,2.1)+(0,-4pt)$) -- ($(3,2.1)+(-4pt,0)$) -- cycle;
% green: filled upward triangle
\fill[green] ($(4,1.7)+(0,4pt)$) -- ($(4,1.7)+(4pt,-3.5pt)$) -- ($(4,1.7)+(-4pt,-3.5pt)$) -- cycle;
\foreach \x in {0,1,2,3,4}
\draw[shift={(\x,0)},color=black] (0pt,2pt) -- (0pt,-2pt) node[below] {\x};
\foreach \y in {0,1,2,3}
\draw[shift={(0,\y)},color=black] (2pt,0pt) -- (-2pt,0pt) node[left] {\y};
\end{tikzpicture}
\medskip

% Add this for the common legend
\begin{tikzpicture}
% Define legends
    \def\legendposx{4}
    \def\legendposy{2.5}
    
    % Series 1 (amber): dotted line, circle marker
    \draw[->, amber, thick, dotted] (\legendposx, \legendposy) -- ++(0.5,0) node[right] {\textcolor{amber}{$\mathbb{E}[\ln(\bm{I^1_{s,c}})|{\Vec{X}_{1,c}}]$}};
    \filldraw[amber] (\legendposx, \legendposy-1) circle (3pt) node[right] {\textcolor{amber}{$\mathbb{E}[\ln(\bm{I^1_{1,c}})|{\Vec{X}_{1,c}}]$}};
    % Series 2 (blue): long-dash line, square marker
    \draw[->, blue, thick, dash pattern=on 8pt off 4pt] (\legendposx+6, \legendposy) -- ++(0.5,0) node[right] {\textcolor{blue}{$\mathbb{E}[\ln(\bm{I^2_{s,c}})|{\Vec{X}_{2,c}}]$}};
    \fill[blue] ($(\legendposx+6, \legendposy-1)+(-3pt,-3pt)$) rectangle ($(\legendposx+6, \legendposy-1)+(3pt,3pt)$);
    \node[right] at (\legendposx+6, \legendposy-1) {\textcolor{blue}{$\mathbb{E}[\ln(\bm{I^2_{2,c}})|{\Vec{X}_{2,c}}]$}};
    % Series 3 (purple): dash-dot line, diamond marker
    \draw[->, purple, thick, dash pattern=on 6pt off 2pt on 2pt off 2pt] (\legendposx, \legendposy-2) -- ++(0.5,0) node[right] {\textcolor{purple}{$\mathbb{E}[\ln(\bm{I^3_{s,c}})|{\Vec{X}_{3,c}}]$}};
    \fill[purple] ($(\legendposx, \legendposy-3)+(0,4pt)$) -- ($(\legendposx, \legendposy-3)+(4pt,0)$) -- ($(\legendposx, \legendposy-3)+(0,-4pt)$) -- ($(\legendposx, \legendposy-3)+(-4pt,0)$) -- cycle;
    \node[right] at (\legendposx, \legendposy-3) {\textcolor{purple}{$\mathbb{E}[\ln(\bm{I^3_{3,c}})|{\Vec{X}_{3,c}}]$}};
    % Series 4 (green): medium-dash line, triangle marker
    \draw[->, green, thick, dash pattern=on 4pt off 3pt] (\legendposx+6, \legendposy-2) -- ++(0.5,0) node[right] {\textcolor{green}{$\mathbb{E}[\ln(\bm{I^4_{s,c}})|{\Vec{X}_{4,c}}]$}};
    \fill[green] ($(\legendposx+6, \legendposy-3)+(0,4pt)$) -- ($(\legendposx+6, \legendposy-3)+(4pt,-3.5pt)$) -- ($(\legendposx+6, \legendposy-3)+(-4pt,-3.5pt)$) -- cycle;
    \node[right] at (\legendposx+6, \legendposy-3) {\textcolor{green}{$\mathbb{E}[\ln({I^4_{4,c}})|\bm{\Vec{X}_{4,c}}]$}};
    
    \end{tikzpicture}
\caption{Each potential growth model, i.e., {\color{amber}$\ln(\bm{I^1_{s,c}})$ (dotted)}, {\color{blue}$\ln(\bm{I^2_{s,c}})$ (long-dashed)}, {\color{purple}$\ln(\bm{I^3_{s,c}})$ (dash-dot)}, and {\color{green}$\ln(\bm{I^4_{s,c}})$ (dashed)}, can exhibit distinct patterns under Assumptions 1 and 2. The unobservable components of these potential growth models are indicated by non-solid lines. The observable components of these potential growth models, which coincide with the actual growth model $\ln(\bm{I_{s,c}})$, are depicted as solid markers.}
\label{AssumptionExamples}
\end{figure}

% \textmd{Figure (a): Both Assumptions are not satisfied;\\
% Figure (b): Assumption 1 is not satisfied but Assumption 2 is satisfied;\\
% Figure (c): Assumption 1 is satisfied but Assumption 2 is not satisfied;\\
% Figure (d): Both Assumptions are satisfied}}

%$\gamma_{t^-,c}\in \mathbb{R}^+\ \ \text{and}\ \ \sum^t_{t^-=t-\delta+1}\gamma_{t^-,c}=1$,

\subsection{Discussion of Identification Assumptions}\label{IAD} 

In this section, we explore the practical implications of our identification assumptions by examining how the actual growth model (\ref{observed}) can be identified within the conventional OLS framework. This analysis not only highlights the limitations of the OLS approach but also serves to motivate the development of our \texttt{TLRF} estimator (\ref{T3r}) in the subsequent section, which generalizes the OLS estimator to accommodate the complexities of our model.\par

Identifying the actual growth model (\ref{observed}) is challenging due to the temporal and spatial heterogeneity of the exponential growth rate $r(\bm{\Vec{X}_{s,c}})$. The traditional OLS approach addresses this challenge by imposing strong homogeneity assumptions over a $\delta$-day fitting window:
$\forall t^-\in \{t-\delta+1,\dots,t\}$,
\begin{equation}\label{Hom}
    {r}({\Vec{X}_{t,c}})={r}({\Vec{X}_{t^-,c}}).
\end{equation}
These stringent assumptions result in inflexible OLS estimators. Specifically, we show in Appendix \ref{vsOLS} that any OLS estimators of ${r_{t,c}}$ with a $\delta$-day fitting window can be expressed as a convex combination of $\delta-1$ two-point estimators in the form of (\ref{fw}), i.e.,
\begin{equation}\label{rtcOLS}
    \bm{\hat{r}^{ols(\delta)}_{t,c}}=\sum^t_{t^-=t-\delta+2}\hat{\mu}_{t^-,c}\left(\ln(\bm{I_{t^-,c}})-\ln(\bm{I_{t^--1,c}}) \right),
\end{equation}
where $\{\hat{\mu}_{t^-,c}\}_{t^-\in \{t-\delta+2,\dots,t\}}$ are non-negative weights that sum to 1, i.e., 
\begin{equation}\label{olsweight}
    \hat{\mu}_{t^-,c}=  \frac{\sum^t_{i=t^-}(i-\frac{2t-\delta+1}{2})}{\sum^t_{i=t-\delta + 1}(i-(t-\delta+1))(i-\frac{2t-\delta+1}{2})}.
\end{equation}
In other words, OLS identification assumptions introduce unnecessary constraints in OLS estimators and, thus, inflexible and non-adaptive estimation. In contrast, our approach offers more flexibility through the potential growth framework, where each data point from the actual growth model (\ref{observed}) has a unique data-generating process (\ref{POt}). This framework eliminates the need for the homogeneity assumption in (\ref{Hom}) and allows for more flexible and adaptive estimation. In particular, we demonstrate in \S\ref{ME} that our identification framework allows the \texttt{TLRF} estimator to leverage a fully adaptive weighting scheme to replace the fixed OLS weights (\ref{olsweight}). This flexibility is essential for capturing COVID-19's heterogeneous transmission patterns across time and space.\par

Having established the model identification, we now turn to the derivation of our \texttt{TLRF} estimator in \S\ref{ME}.

\section{Model Estimation}\label{ME}

This section discusses how we estimate the instantaneous county-level exponential growth rate ${r_{t,c}}$, identified by (\ref{fw}). To this end, \S\ref{4.1} first derives the conditional mean outcome and an initial estimate of ${r_{t,c}}$. Subsequently, \S\ref{4.2} improves this initial estimate by incorporating relevant proxy data through transfer learning enabled by \texttt{TLRF}. Lastly, \S\ref{ITLRF} discusses the implementation details of \texttt{TLRF}.

\subsection{Initial Estimator}\label{4.1}

Corollary \ref{T2} derives the initial estimator for ${r_{t,c}}$ identified under Assumptions 1 and 2. 

%we derive the local moment equation of the estimation problem in Corollary \ref{T2}.

\begin{corollary}\label{T2}
For any county $c \in C$ on any day $t\in \mathbb{Z}^+\backslash\{1\}$ with realized features $x_{t,c}$, its instantaneous county-level exponential growth rate ${r_{t,c}}$ can be estimated from data  $\{\ln({I_{t,c}}), \ln({I_{t-1,c}}), \Vec{X}_{t,c}\}$ via the conditional mean outcome
\begin{equation}\label{OLSID}
r({\Vec{X}}_{t,c})=\mathbb{E}[(\ln(\bm{I_{t,c}})-\ln(\bm{I_{t-1,c}}))|\Vec{X}_{t,c}],
%    \mathbb{E}[\left(\bm{Y_{t,c}}(\bm{s})-r(\bm{\Vec{X}}){W_{t,c}}(\bm{s})\right){W_{t,c}}(\bm{s})|\bm{\Vec{X}}=\Vec{X}_{t,c}]=0, 
%\ \ \text{for all}\ \ x\in \text{Range}(\Vec{X}_i)
\end{equation}
which yields the following initial estimator:
\begin{equation}\label{T2r}
\bm{\hat{r}_{t,c}}=\ln(\bm{I_{t,c}})-\ln(\bm{I_{t-1,c}}).
\end{equation}
\end{corollary}
\noindent
We observe that (\ref{T2r}) is the same as the two-point estimator (\ref{fw}), which is equivalent to an OLS estimator with a 2-day fitting window as in (\ref{rtcOLS}). This estimator best captures the case growth rate of this county on this day but is noisy. To improve this initial estimator further, we propose our \texttt{TLRF} estimator in \S\ref{4.2}, which adaptively chooses fitting windows by pooling data across space and time through transfer learning.

\subsection{Transfer Learning through Random Forests}\label{4.2}

In \S\ref{4.1}, we showed that the target data $\{\ln({I_{t,c}}), \ln({I_{t-1,c}}),\Vec{X}_{t,c}\}$ is sufficient to produce the initial estimator (\ref{T2r}). To further improve this initial estimator, we incorporate relevant proxy data $\{\ln({I_{t',c'}}), \ln({I_{t'-1,c'}}),{\Vec{X}_{t',c'}}\}$ from other counties $c'\neq c$ and days $t'\neq t$ into the estimation through transfer learning.\par

For any county $c \in C$ on any day $t\in \mathbb{Z}^+$ with relevant features $\Vec{X}_{t,c}$, we construct a similarity measure $\bm{\gamma_{t',c'}}(\Vec{X}_{t,c})$ between the target data $\{\ln({I_{t,c}}), \ln({I_{t-1,c}}),\Vec{X}_{t,c}\}$ and the proxy data $\{\ln({I_{t',c'}}), \ln({I_{t'-1,c'}}),{\Vec{X}_{t',c'}}\}$ from other counties $c'\neq c$ and days $t'\neq t$ using random forests. Following the adaptive kernel interpretation of random
forests \citep[c.f.][]{athey2019generalized,friedberg2020local},  this algorithm grows a forest $B:=\{1,\dots, |B|\}$, where for each tree indexed by $b\in B$, the algorithm recursively partitions the training sample through an axis-aligned cut of relevant features based on the similarity of their underlying instantaneous county-level exponential growth rates.
As such, training samples whose features fall into the same leaf as that of the target data, denoted as $L_b(\Vec{X}_{t,c})$, are effectively used to estimate ${r_{t,c}}$. The similarity measure, which we denote ${\gamma_{t',c'}}(\Vec{X}_{t,c})$, is thus constructed as follows
\begin{equation}\label{weight}
    {\gamma_{t',c'}}(\Vec{X}_{t,c}):=\frac{1}{|B|}\sum\limits_{b=1}^{|B|}\frac{\mathbb{1} (\{{\Vec{X}_{t',c'}}\in L_b(\Vec{X}_{t,c})\})}{|L_b(\Vec{X}_{t,c})|},
\end{equation}
which measures how frequently each training sample $\{\ln(\bm{I_{t',c'}}), \ln(\bm{I_{t'-1,c'}}),{\Vec{X}_{t',c'}}\}$ is in the same leaf as the target data $\{\ln({I_{t,c}}), \ln({I_{t-1,c}}),\Vec{X}_{t,c}\}$, i.e., ${\Vec{X}_{t',c'}}\in L_b(\Vec{X}_{t,c})$. We note that this similarity measure essentially weights the proxy data based on its similarity to the target data, which resembles classical variance reduction techniques such as importance sampling \citep{sugiyama2008direct}.\par

%\textbf{Assumption 4} \textit{For any county $c \in C$ on any day $t\in \mathbb{Z}^+$ with realized features $\Vec{X}_{t,c}$, $\mathbb{E}[{Y_{t,c}}(\bm{s})|\bm{\Vec{X}}=\Vec{X}_{t,c}]$ is Lipschitz continuous in $\Vec{X}_{t,c}$.}\\

Given these similarity measures, we now present \texttt{TLRF} estimator for the instantaneous county-level exponential growth rate ${r_{t,c}}$. We start with an assumption commonly made for adaptive kernel methods:\\
% \begin{assumption}\label{A3} 
\textbf{Assumption 3} \textit{For any county $c \in C$ on any day $t\in \mathbb{Z}^+\backslash\{1\}$ with realized features $\Vec{X}_{t,c}$, $\mathbb{E}[\ln(\bm{I^t_{t,c}})-\ln(\bm{I^t_{t-1,c}})|\Vec{X}_{t,c}]$ is Lipschitz continuous in $\Vec{X}_{t,c}$.}\\
% \end{assumption}
\noindent
Together with Assumptions 1 and 2, Assumption 3 implies that $r({\Vec{X}}_{t,c})$ is changing continuously with relevant features, which implies that proxy data whose features are close enough to that of the target data have similar underlying instantaneous county-level exponential growth rates. As such, Assumption 3 allows us to estimate ${r_{t,c}}$ using the entirety of data up to day $t$, i.e.,
\[
    \{\ln(\bm{I_{t',c'}}), \ln(\bm{I_{t'-1,c'}}),{\Vec{X}_{t',c'}}\}_{t'\in [t],c'\in C} \text{ where } [t]:= \{t'\in \{1,\dots,t\}| t \equiv t'\ (\textrm{mod}\ 2)\},
\] 
where mod 2 is the standard modulo 2 operation such that
\[
    [t]:=\begin{cases}
    \{1,3,5,\dots,t\} & \text{, if $t$ is odd}\\
    \{2,4,6,\dots,t\} & \text{, if $t$ is even}.
    \end{cases}
\]
\noindent
Under these assumptions, the \texttt{TLRF} estimator, as described in Theorem \ref{T3}, is obtained.
\begin{theorem}\label{T3}
For any county $c, c' \in C$ on any day $t, t'\in \mathbb{Z}^+$ with realized features $\Vec{X}_{t,c}$ and $\vec{X}_{t',c'}$ such that $\Vec{X}_{t',c'}\in \underset{b\in B}{\cup} L_b(\Vec{X}_{t,c})$, its instantaneous county-level exponential growth rate ${r_{t,c}}$ can be consistently estimated from data $ \{\ln({I_{t',c'}}), \ln({I_{t'-1,c'}}),\Vec{X}_{t',c'}\}_{t'\in [t],c'\in C}$ via the weighted conditional mean outcome:
\begin{equation}\label{GRFID}
r({\Vec{X}}_{t,c})= \sum_{t'\in [t]}\sum_{c'\in C}\gamma_{t',c'}(\Vec{X}_{t,c})\mathbb{E}[(\ln(\bm{I_{t',c'}})-\ln(\bm{I_{t'-1,c'}}))|\Vec{X}_{t,c}]\\
% \sum_{t'\in [t]}\sum_{c'\in C}\gamma_{t',c'}(X_{t,c}) \mathbb{E}\left[\left({Y}_{t',c'}(\bm{s})-{r}(x_{t,c}){W}_{t',c'}(\bm{s})\right) {W}_{t',c'}(\bm{s})|\bm{\Vec{X}} =X_{t',c'}\right]=0, 
%\ \ \text{for all}\ \ x\in \text{Range}(\Vec{X}_i)
%$\{\{\bm{Y}_{t',c'}(t^-),{W}_{t',c'}(t^-), \bm{\Vec{X}_{t',c'}}\}_{t^-\in \{t'-1,t'\}}\}_{t'\in [t],c'\in C}$ via
\end{equation}
resulting in the following \texttt{TLRF} estimator:
\begin{equation}\label{T3r}
  \bm{\hat{r}^{TLRF}_{t,c}}= \sum\limits_{t'\in [t]}\sum\limits_{c'\in C}  {\gamma_{t',c'}}(\Vec{X}_{t,c}) \left(\ln(\bm{I_{t',c'}})-\ln(\bm{I_{t'-1,c'}})\right).
\end{equation}
\end{theorem}
\noindent
We note that the initial estimator presented in Theorem \ref{T2} is a special case of the \texttt{TLRF} estimator derived in Theorem \ref{T3}. Specifically, if no proxy data is in the same leaf as the target data for all trees in the forest, i.e., $\underset{b\in B}{\cup} L_b(\Vec{X}_{t,c})=\{\Vec{X}_{t,c}\}$, we would have a similarity measure concentrated at the target data, i.e., $\gamma_{t,c}(\Vec{X}_{t,c})=1$ and $\gamma_{t',c'}(\Vec{X}_{t,c})=0$ for all $t'\neq t$ or $c'\neq c$. In that case, the weighted conditional mean outcome \eqref{GRFID} and the \texttt{TLRF} estimator \eqref{T3r} reduce to the conditional mean outcome \eqref{OLSID} and the initial estimator \eqref{T2r}, respectively. Lastly, we would like to clarify that since (\ref{T3r}) uses features $\Vec{X}_{t,c}$ in a fully nonparametric way, it is difficult to isolate the causal effect of any single feature. Therefore, the feature vector $\Vec{X}_{t,c}$ in the \texttt{TLRF} estimator does not have a causal interpretation.\par

To better understand the flexibility of the \texttt{TLRF} estimator \eqref{T3r}, we compare it with OLS estimators (\ref{rtcOLS}) with $\delta$-day fitting windows. It is noteworthy that both estimators can be expressed as a convex combination of two-point estimators (\ref{fw}), i.e., 
\begin{align*}
    \bm{\hat{r}^{TLRF}_{t,c}}&= \sum\limits_{t'\in [t]}\sum\limits_{c'\in C}  {\gamma_{t',c'}}({\Vec{X}_{t,c}}) \left(\ln(\bm{I_{t',c'}})-\ln(\bm{I_{t'-1,c'}})\right) \numberthis\label{GRFAD}
    % &=\sum\limits_{t'\in [t]}\sum\limits_{c'\in C}  \gamma_{t',c'}(x) \bm{r_{t',c'}} \numberthis\label{GRFAD}
\end{align*}
and
\begin{align*}
   \bm{\hat{r}^{ols(\delta)}_{t,c}}= \sum^t_{t'=t-\delta+2}\hat{\mu}_{t',c} \left(\ln(\bm{I_{t',c}})-\ln(\bm{I_{t'-1,c}})\right) \numberthis\label{OLSNAD} 
\end{align*}
The two estimators differ along two key dimensions. First, the OLS estimator (\ref{OLSNAD}) pools observations only over time within the same county, while \texttt{TLRF} estimator (\ref{GRFAD}) pools data across both time and counties. Second and more importantly, the weights of the \texttt{TLRF} estimator (\ref{GRFAD}) are adaptive to the relevant features, while the weights of the OLS estimators (\ref{OLSNAD}), as specified in (\ref{olsweight}), are non-adaptive. In other words, by allowing for more flexible weighting, the \texttt{TLRF} estimator \eqref{T3r} generalizes the OLS estimators with non-adaptive fitting windows. A formal analysis of how these two fitting window selection approaches affect the bias-variance tradeoff of their corresponding estimators is presented in
Appendix \ref{AppendixBiasVariance}.

\subsection{Implementation of TLRF}\label{ITLRF}

% The \texttt{TLRF} method incorporates a customized data transformation tailored for estimating instantaneous growth rates. Unlike the direct use of raw data (e.g., Table \ref{GRFdata}), \texttt{TLRF} preprocesses the data as follows:
%     \[\{\ln(\bm{I_{t',c'}}), \ln(\bm{I_{t'-1,c'}}),{\Vec{X}_{t',c'}}\}_{t'\in [t],c'\in C} \text{ where } [t]:= \{t'\in \{1,\dots,t\}| t \equiv t'\ (\textrm{mod}\ 2)\}. \numberthis \label{TLGRFTransformation}\]
% As discussed in \S\ref{4.2}, this transformation enables the \texttt{TLRF} estimator (\ref{T3r}) to eliminate dependency on the non-adaptive OLS weights (\ref{olsweight}) inherent in the direct \texttt{causal\_forest()} implementation. Moreover, this preprocessing facilitates the use of both the \texttt{causal\_forest()} and \texttt{regression\_forest()} functions for county-level instantaneous exponential growth rate estimation, addressing the limitations of the direct GRF implementation.

\begin{table}[ht]
\centering
 \begin{tabular}{|c|c|}
\hline
Dependent Variable  & Feature  \\
\hline
$\ln(I_{2,1})-\ln(I_{1,1})$ & $\Vec{X}_{2,1}$ \\\hline
$\ln(I_{4,1})-\ln(I_{3,1})$ & $\Vec{X}_{4,1}$ \\\hline
\vdots & \vdots \\
\hline
\end{tabular}
\caption{\texttt{TLRF} through random regression forests \citep[c.f.][]{breiman2001random} Implementation}
\label{TLGRFDataRegression}
\end{table}

%\subsubsection{TLRF \texttt{regression\_forest()} Implementation} 
Based on Corollary \ref{T2}, our parameter of interest $r({\Vec{X}}_{t,c})$ can be identified through the conditional mean:
\[
{r_{t,c}}=\mathbb{E}[\ln(\bm{I_{t,c}})-\ln(\bm{I_{t-1,c}})|\Vec{X}_{t,c}].
\]
This allows for estimation using the random regression forest method \citep[c.f.][]{breiman2001random}:
\begin{align*}
\bm{\hat{r}^{RF}_{t,c}}=\frac{1}{|B|}\sum^{|B|}_{b=1}\bm{\hat{r}^b_{t,c}}, \numberthis \label{RFP}
\end{align*}
where each tree's prediction is given by:
$$\bm{\hat{r}^b_{t,c}}:=\sum\limits_{t'\in [t]}\sum\limits_{c'\in C} (\ln(\bm{I_{t',c'}})-\ln(\bm{I_{t'-1,c'}}))\frac{\mathbb{1} ({{\Vec{X}_{t',c'}}\in L_b(\Vec{X}_{t,c})})}{|L_b(\Vec{X}_{t,c})|}.$$
More importantly, we note that 
\begin{proposition}\label{GRFEquivalent}
The random forest predictor (\ref{RFP}) can be cast as an adaptive locally weighted estimator, and is thus equivalent to the \texttt{TLRF} estimator (\ref{T3r}):
$$\bm{\hat{r}^{RF}_{t,c}}=\bm{\hat{r}^{TLRF}_{t,c}}.$$
\end{proposition}
Therefore, the \texttt{TLRF} method can be implemented through standard regression forests, with a typical input represented by Table \ref{TLGRFDataRegression}. Pseudocode of \texttt{TLRF} is presented in Appendix \ref{apd1.TLRF}. A detailed discussion of this algorithm's stability with regard to its key hyperparameters is provided in Appendix \ref{apd:TLGRF_Hyperparameter_Tuning}.\par

In our implementation, \texttt{TLRF} is deployed on a high-performance compute server equipped with 1TB of memory and 127 AMD EPYC-Rome processors. For the complete U.S. dataset (with each county-day representing a single observation), model retraining with 200 trees requires approximately 20 minutes when incorporating one additional day of data across all counties, inclusive of exhaustive hyperparameter selection. Once trained, the model performs estimation and forecasting inference for all counties in approximately 10 seconds per day, enabling near real-time epidemiological monitoring.

%If this were to be run from scratch across ~2+ years, it would take around 25 hours.

To summarize, in order to improve the estimation, our proposed algorithm adaptively chooses fitting windows by pooling proxy data across time and counties via the \texttt{TLRF} similarity measures. In \S\ref{PE}, we evaluate the numerical performance of our estimator by benchmarking it against traditional estimators with both non-adaptive fitting windows and fitting windows chosen via time series cross-validation.

\section{Performance Evaluation Against Other Fitting Window Size Selection Methods}\label{PE}
This section empirically assesses the performance of the \texttt{TLRF} algorithm. Specifically, we compare the prediction of the \texttt{TLRF} algorithm against true case numbers 7 days ahead and report two well-established prediction accuracy measures in the epidemiology and operations management literature, i.e., the mean absolute error (MAE) and root-mean-squared error (RMSE) \citep{bertsimas2021predictions}. That is, to predict the COVID-19 incident case number $\bm{\hat{I}_{t+7,c}}$ at county $c$ on day $t+7$, we first estimate the instantaneous county-level growth rate  at county $c$ on day $t$, and generate the prediction as follows
\begin{equation}\label{forecast}
    \bm{\hat{I}_{t+7,c}}=e^{\ln(\bm{I_{t,c}})+7 \bm{\hat{r}_{t,c}}}.
\end{equation}
Therefore, the prediction accuracy depends solely on the instantaneous county-level growth rate estimate $\bm{\hat{r}_{t,c}}$.\par

%\cite{chowell2017fitting} shows that an
% $\{w_{t-\delta+1,c'}\}_{\delta\in \{1,\dots,t\}, c'\in C}$ are assigned 
% in the following weighted least square (WLS) estimation (c.f. ), i.e.
% For example, (\ref{wls}) 

%
% first present a unified framework of fitting window size selection in this class of problems and then explain our inclusion and exclusion criteria of benchmarking algorithm selection. To begin with, we note that fitting window size choice in exponential growth rate estimation is essentially determined by how weights are assigned in exponential growth rate estimation problems. For example, an OLS estimator for $\bm{r_{t,c}}$ based on 14-day fitting window can be obtained from  
% \begin{equation*}
% \{\bm{\hat{\alpha}_{t,c}},\bm{\hat{r}_{t,c}}\}=\underset{\{\bm{\alpha_{t,c}},\bm{r_{t,c}}\}}{argmin}\{\sum^{|C|}_{c'=1}\sum^t_{\delta=1}w_{t-\delta+1,c'}(\bm{\alpha_{t,c}}+\bm{r_{t,c}}(t-\delta+1)-\ln(\bm{I_{t-\delta+1,c'}}))^2\}
% \end{equation*}
% by assigning weights as $w_{t-\delta+1,c'}=\frac{1}{14}$ for $\delta\in \{1,\dots,14\}\ c'=c$ and $w_{t-\delta+1,c'}=0$ otherwise.

In line with the paper's central premise, our comparison primarily concentrates on varying techniques for estimating exponential growth rates, particularly emphasizing the differences in fitting window selection. Specifically, selecting fitting windows essentially involves determining the weights $\{w_{t',c'}\}_{t'\in \{2,\dots,t\},c'\in C}$ for estimating ${r_{t,c}}$:
\begin{align}
    \underset{c'\in C}{\sum}\sum^t_{t'=2}w_{t',c'} \left(\ln(\bm{I_{t',c'}})-\ln(\bm{I_{t'-1,c'}})\right).\numberthis \label{General}
\end{align}
For example, we demonstrate in \S\ref{IAD} that any OLS estimators of ${r_{t,c}}$ with a $\delta$-day fitting window can be expressed as (\ref{General}) with the weight assignment $w_{t',c}=\hat{\mu}_{t',c}\ \ \forall t'\in \{t-\delta+2,\dots,t\}$ are defined in accordance with (\ref{olsweight}), while $w_{t',c'}=0\ \ \forall t'\in \{2,\dots,t-\delta+1\}\ \text{or}\ \forall c'\in C\backslash\{c\}$. In this section, we compare the weight assignment scheme of \texttt{TLRF}, as specified in Theorem \ref{T3}, against three classes of weight assignment methods. First, we demonstrate that \texttt{TLRF} consistently outperforms all fixed window size choices in $\S$\ref{subsec.fixed}. Then, in $\S$\ref{subsec.flexible}, we show that \texttt{TLRF} also outperforms dynamic window size algorithms that pool data and conduct cross validation across time. Finally, in \S\ref{kmeans}, we conduct a comparison between the empirical performance of \texttt{TLRF} and K-Means Dynamic Window size algorithms. Similar to \texttt{TLRF}, this algorithm is capable of pooling data and performing cross validation across both time and space. Through this analysis, we provide evidence that the superiority of \texttt{TLRF} stems not only from its capacity to pool data across time and space, but also from its adaptive weighting schemes as outlined in Theorem \ref{T3}. In contrast, we note that alternative fitting window choice algorithms rely, at least partially, on non-adaptive OLS weight assignments (\ref{olsweight}). Properties of the benchmarks against which we compare \texttt{TLRF} are summarized in Table \ref{tab:Inclusion_Exclusion_Table_OLS}. 
\begin{table}[ht]
\centering
\begin{tabular}{|c|c|c|c|c|}
\hline
\multirow{2}{*}{\bfseries (Estimator)} & \multirow{2}{*}{\bfseries\begin{tabular}[c]{@{}c@{}}Cross Validation\\ Across Space\end{tabular}} & \multirow{2}{*}{\bfseries\begin{tabular}[c]{@{}c@{}}Cross Validation\\ Across Time\end{tabular}} & \multirow{2}{*}{\bfseries\begin{tabular}[c]{@{}c@{}} Weighting\\ Schemes \end{tabular}} \\
 & & & \\
\hline
\bfseries Fixed Windows & & & \textbf{OLS} \\
\bfseries Dynamic Windows & & \checkmark & \textbf{OLS} \\
\textbf{K-Means Dynamic Windows} & \checkmark  & \checkmark & \textbf{OLS}\\
%\bfseries Classical GRF & \checkmark & & \checkmark \\
\bfseries TLRF & \checkmark & \checkmark & \textbf{TLRF} \\
\hline
\end{tabular}
\caption{Comparison of exponential growth rate estimators: Fixed Window Size Method, Dynamic Window Size Method, i.e., Time-based Cross Validation (tcv) and County and Time-based Cross Validation (ctcv), K-means Dynamic Window Size Method, and \texttt{TLRF}}
\label{tab:Inclusion_Exclusion_Table_OLS}
\end{table}
%\footnote{The closest paper to ours is \cite{altieri2020curating}, who proposed a new exponential predictor for county-level COVID death forecast.}

Lastly, we remark that while there are many COVID-19 prediction algorithms and time series clustering methods \citep[c.f.][]{wynants2020prediction,ma2019learning}, few of them aim to improve the estimation of exponential growth rates, which is the main focus of this paper. In this section, we only include those benchmarking algorithms that are capable of estimating exponential growth rates for comparison.

\subsection{Performance Comparison Against Fixed Window Size Methods}\label{subsec.fixed}

In this section, we compare the empirical performance of \texttt{TLRF} against fixed window size methods. As established in \S\ref{IAD}, any OLS estimators of ${r_{t,c}}$ with a $\delta$-day fitting window can be expressed as
\begin{equation}\label{ols5}
\bm{\hat{r}^{ols(\delta)}_{t,c}}=  \sum^t_{t'=t-\delta+2}\hat{\mu}_{t',c} \left(\ln(\bm{I_{t',c}})-\ln(\bm{I_{t'-1,c}})\right),
\end{equation}
where $\{\hat{\mu}_{t',c}\}_{t'\in \{t-\delta+2,\dots,t\}}$ are the OLS weights (\ref{olsweight}).  

%\footnote{\cite{bergman2020oscillations} reported that the oscillation in U.S. COVID-19 case counts can in part be explained by increased testing during some week days and test backlogs during the weekend. To account for daily variations in reported cases caused by changing testing rates, we compute and work with the 4-day moving averages} Additional results for performance evaluation are available at Appendix \ref{apdD}. 

%that while our proposed algorithm initially performs inferior to some of the fixed window-size algorithms due to the lack of proxy data

\begin{figure}[htpb!]
  \includegraphics[width=\linewidth, height=0.6\textwidth]{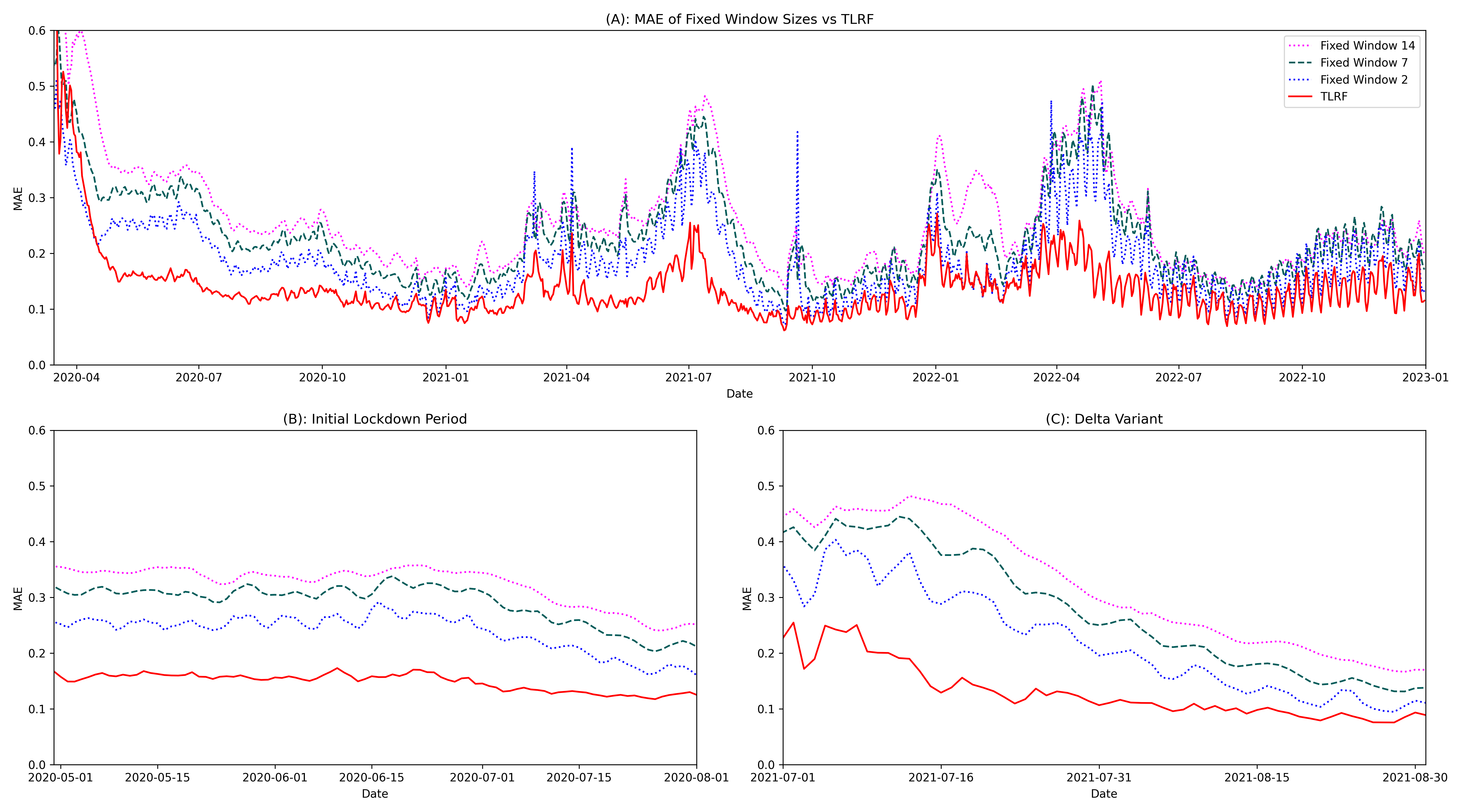}
  \caption{MAE plot of prediction accuracy (\texttt{TLRF} vs. Fixed 2-, 7-, and 14-Day Fitting Window): Plot A (Top) shows the MAE comparison across the study period. Plot B (bottom left) zooms in on the initial lockdown period. Plot C (bottom right) zooms in on the period during the outbreak of the Delta variant.}
  \label{fig:MAE4_together}
\end{figure}

Figure \ref{fig:MAE4_together} characterizes the performance of \texttt{TLRF} against algorithms with fixed 2-, 7-, and 14-fitting window sizes, starting from the first day of recorded COVID-19 cases in NYTimes COVID-19 dataset \citep{NYTimes2020}. The results summarized in this figure show that \texttt{TLRF} consistently outperforms all the fixed window-size algorithms after April 2020, when there were enough proxy data to conduct transfer learning. In addition, Table \ref{tab:fixed_windows} demonstrates that \texttt{TLRF} achieves MAE $0.126$ and RMSE $0.195$, which are respectively $27.4\%$ and $42.5\%$ lower than their closest competitors. Therefore, \texttt{TLRF} reliably estimates the instantaneous county-level growth rates, including those counties with small sample sizes, and outperforms all fixed window-size algorithms with the help of transfer learning. Additional results on the comparison between \texttt{TLRF} and fixed window size algorithms are available in Appendix \ref{supplementary_fixed}.\par

% Next, in  we quantify the median MAEs and RMSEs of. The findings revealed in Table \ref{tab:fixed_windows} are consistent with those of Figure \ref{fig:MAE4_together}. Specifically,

%\tdelete{Figure \ref{fig:MAE4_together} shows that \texttt{TLRF} consistently outperforms all the fixed window-size algorithms after April 2020, when there were enough proxy data to conduct transfer learning. In other words, \texttt{TLRF} reliably estimates the instantaneous county-level growth rates, including those counties with small sample sizes, and outperform all fixed window-size algorithms with the help of transfer learning.}\par

In addition, this experiment also shows that there is no clear winner among the fixed window size algorithms considered, indicating that optimal window sizes for different counties are different and change over time. To see this, please observe that even though Table \ref{tab:fixed_windows} shows Fixed Window 2 has the lowest median MAE and RMSE compared to the other fixed window-size algorithms, Figure \ref{fig:MAE4_together} reveals that Fixed Window 2 has been outperformed by others fitting window size choices on many different days. This is because, at different time points of the pandemic, the rate of spread varies, and no fixed time window algorithm has the flexibility to capture this changing nature of disease epidemiology. \par 

\begin{table}[htpb!]
\centering
  \begin{tabular}{lcc}
  \toprule
  \bfseries Method &  \bfseries MAE &  \bfseries RMSE \\
  \midrule
Fixed Window 14 & 0.240 & 0.373 \\
Fixed Window 7  & 0.213 & 0.365 \\
Fixed Window 2  & 0.175 & 0.339 \\
\textbf{TLRF}  & \textbf{0.126} & \textbf{0.195} \\
  \bottomrule
  \end{tabular}
    \caption{Median MAE and RMSE of \texttt{TLRF} vs. Fixed Window 2-, 7-, and 14-Day Fitting Window}
    \label{tab:fixed_windows}%
\end{table}

\newpage
% Merge Both Subsections 
% \subsection{Performance Gain through Transfer Learning across Time and Space}\label{subsec.two_way}
% % Across Time: tcv ad ctcv
% % Across Counties: GRF
% % Across Time and County: TLRF
% The gain in performance of \texttt{TLRF} comes from its ability to adaptively choose fitting window sizes by pooling proxy data across both time and space. In $\S$\ref{subsec.flexible}, we compare \texttt{TLRF} against exponential growth rate estimators where the fitting window is chosen dynamically over time but not over space. Then, in $\S$\ref{subsec.vsGRF}, we demonstrate that being able only to pool proxy data across space but not across time, the classical GRF algorithm performs significantly worse than \texttt{TLRF}. Lastly, \ref{subsec.vsOLSweight} shows that the superior performance of \texttt{TLRF} is attributed not only to its capacity for integrating proxy data across time and space, but also to its fully adaptive approach to weight assignments. 

%'s ability to pool data across both time and space, as opposed to just one or the other in the case of , results in massive performance gains.

\subsection{Performance Comparison Against Dynamic Window Methods}\label{subsec.flexible}
%\textcolor{red}{WZL: TURGAY Read Results + Case Study + Website + Conclusion}
%\wzladd{CTCV and TCV figures and Tables updated, now update the numbers and percentages in the text}
While most existing literature is predominantly occupied by fixed window-size methods, relatively recent epidemiology literature has started to treat the window size as a hyperparameter that could dynamically change over time \citep{MEL1,chowell2007comparative}. Motivated by these studies, we adopt a similar approach in this section and treat the window size as a hyperparameter that can change over time. More precisely, our approach involves dynamically selecting the appropriate window size $\delta_t$ using cross validation and applying it uniformly across all U.S. counties. We refer to this approach as the Time-based Cross Validation (\texttt{tcv}) algorithm, which yields the following estimator:
\begin{equation}\label{tcv}
    \bm{\hat{r}^{ols(\delta_t)}_{t,c}}=  \sum^t_{t'=t-\delta_t+2}\hat{\mu}_{t',c} \left(\ln(\bm{I_{t',c}})-\ln(\bm{I_{t'-1,c}})\right),
\end{equation}
where $\{\hat{\mu}_{t',c}\}_{t'\in \{t-\delta_t+2,\dots,t\}}$ are the OLS weights (\ref{olsweight}). Notably, the primary distinction between the \texttt{tcv} estimator (\ref{tcv}) and the fixed window size estimator (\ref{ols5}) is that the \texttt{tcv} algorithm dynamically selects the window size $\delta_t$ through cross-validation, whereas the fixed window size estimator (\ref{ols5}) employs a predetermined fixed window size $\delta$. Further details and implementation specifics of the Dynamic Window estimation algorithms are provided in Appendix \ref{apd1.dynamic}.

In addition, we consider another benchmarking algorithm that adjusts fitting window sizes dynamically at the county level, which we refer to as County Time-based Cross Validation (\texttt{ctcv}). That is, this algorithm dynamically selects the appropriate fitting window size $\delta_{t,c}$ for each county using cross validation, and yields the following estimator:
\begin{equation}\label{ctcv}
    \bm{\hat{r}^{ols(\delta_{t,c})}_{t,c}}=  \sum^t_{t'=t-\delta_{t,c}+2}\hat{\mu}_{t',c} \left(\ln(\bm{I_{t',c}})-\ln(\bm{I_{t'-1,c}})\right).
\end{equation}
Again, $\{\hat{\mu}_{t',c}\}_{t'\in \{t-\delta_{t,c}+2,\dots,t\}}$ represent the OLS weights (\ref{olsweight}). Unlike the \texttt{tcv} estimator (\ref{tcv}), where a uniform fitting window size $\delta_t$ is selected for all U.S. counties each day, the \texttt{ctcv} estimator (\ref{ctcv}) chooses a distinct fitting window size $\delta_{t,c}$ for each county on a daily basis by cross validating at the county-level.\par

%For more comprehensive information and implementation details of the \texttt{ctcv} estimation algorithm, please refer to Appendix \ref{apd1.ctcv}.

%This section compares the 7-day ahead prediction performance of these dynamic window-size tuning algorithms (\texttt{tcv} and \texttt{ctcv}) against that of \texttt{TLRF}.\par

%enables learning about the window-size of a given county via data from other similar counties through transfer learning. 

%This section hence helps to quantify the value of additional information gained through transfer learning.\par

%\input{Plots/updated_tcv_ctcv_mae}
\begin{table}[ht]
\centering
  \begin{tabular}{lcc}
  \toprule
  \bfseries Method &  \bfseries MAE &  \bfseries RMSE \\
  \midrule
  \texttt{tcv}  & 0.175 & 0.337\\
  \texttt{ctcv} & 0.175 & 0.314 \\
  \textbf{TLRF}  & \textbf{0.126} & \textbf{0.195}\\
  \bottomrule
  \end{tabular}
    \caption{Median MAE and RMSE of \texttt{TLRF} vs \texttt{tcv} and \texttt{ctcv}}
    \label{tab:updated_flexible}%
\end{table}

%, and Figures \ref{fig:updated_tcv_ctcv_mae_appendix} and \ref{fig:updated_tcv_ctcv_rmse_appendix} in Appendix\ref{supplementary_tcvctcv}.

The performance comparison results between \texttt{tcv}/\texttt{ctcv} and \texttt{TLRF} are summarized in Table \ref{tab:updated_flexible}. Specifically, \texttt{TLRF} outperforms \texttt{tcv} and \texttt{ctcv} with respect to both median MAE and RMSE, with \texttt{TLRF} outperforming them by over $27.4\%$ in terms of median MAE and $37.7\%$ in terms of RMSE.  These findings overall indicate that the ability to transfer learning across counties enables \texttt{TLRF} to improve its performance against other dynamic window-size tuning algorithms. 

%Please refer to Appendix \ref{supplementary_tcvctcv} for additional results on the comparison of performance between \texttt{TLRF} and \texttt{tcv}/\texttt{ctcv}.

%\tadd{ and that the performance gain of \texttt{TLRF} not only comes from dynamic window-size choices but also through its ability to transfer ``learning'' across different counties.}

\subsection{Performance Comparison Against K-Means Dynamic Window Methods}\label{kmeans}

This section investigates a method that selects the best fitting windows between two extremes: the \texttt{tcv} and \texttt{ctcv} algorithms. Specifically, the \texttt{tcv} algorithm performs cross validation at the national level, thereby determining a single dynamic fitting window, $\delta_t$, applicable to all counties on each day $t$. In contrast, the \texttt{ctcv} algorithm implements cross validation exclusively at the county level, thus establishing a unique dynamic fitting window, $\delta_{t,c}$, for each individual county $c$ on each day $t$. Between these two extremes, a spectrum of fitting window size selection methods exists.\par

Investigation of this spectrum can be achieved through the utilization of k-means clustering methods. This involves partitioning all the U.S. counties into $K$ clusters, with counties within each cluster sharing similar time-invariant characteristics, such as the geographical coordinates of county centroids. For each resulting cluster $C_k\in \{C_1,\dots,C_K\}$, cross validation can be performed internally to dynamically determine a fitting window size $\delta_{t,k}$, which will be applicable to all counties in the specific cluster on each day $t$. This procedure is denoted as the K-Means Dynamic Window algorithm. The resultant estimator is defined as follows: $\forall c\in C_k$,
\begin{equation}\label{kmeanstcv}
    \bm{\hat{r}^{ols(\delta_{t,k})}_{t,c}}=  \sum^t_{t'=t-\delta_{t,k}+2}\hat{\mu}_{t',c} \left(\ln(\bm{I_{t',c}})-\ln(\bm{I_{t'-1,c}})\right).
\end{equation}
Again, $\{\hat{\mu}_{t',c}\}_{t'\in \{t-\delta_{t,k}+2,\dots,t\}}$ represent the OLS weights (\ref{olsweight}). Notably, when there is only 1 cluster, the K-Means Dynamic Window estimator (\ref{kmeanstcv}) reduces to the \texttt{tcv} estimator (\ref{tcv}). In contrast, when each county forms its own cluster, the K-Means Dynamic Window estimator (\ref{kmeanstcv}) becomes the \texttt{ctcv} estimator (\ref{ctcv}). Additional implementation details are documented in Appendix \ref{apd1.kdw}.\par

The performance comparison results between the K-Means Dynamic Window method and \texttt{TLRF} are presented in Table \ref{tab:best_kmeans_table}. As highlighted in Table \ref{tab:best_kmeans_table}, \texttt{TLRF} surpasses the K-Means Dynamic Window estimators in both median MAE and RMSE metrics, exceeding their performance by over $26.6\%$ in median MAE and $37.7\%$ in RMSE.  More detailed performance comparisons can be found in Appendix \ref{apd1.kdw}.  

%Furthermore, Figures \ref{fig:best_kmeans_MAE} and \ref{fig:best_kmeans_RMSE} indicate that \texttt{TLRF}'s superior performance over the K-Means Dynamic Window method is sustained throughout the study period.

%\input{Plots/Best_Kmeans_MAE}

%\input{Plots/Best_MAE_Kmeans_Daily}
\begin{table}[ht]
\captionsetup{justification=raggedright,singlelinecheck=false}
\centering
\begin{tabular}{ccc}
\hline
\bfseries Method & \bfseries MAE & \bfseries RMSE \\
\hline
%K=1 (\texttt{tcv}) &  0.175 & 0.337 \\
K=800 &  0.173 & 0.333 \\
K=3100 &  0.175 & 0.313 \\
%K=3136 (\texttt{ctcv}) &  0.175 & 0.314 \\
\textbf{TLRF} &  0.126 & 0.195 \\
\hline
\end{tabular}
    \caption{Median MAE and RMSE of \texttt{TLRF} vs. K-Means Dynamic Window: The K-Means Dynamic Window estimators achieve the lowest median MAE and RMSE when $K=800$ and $K=3100$, respectively.}
    \label{tab:best_kmeans_table}%
\end{table}

These findings indicate that the enhanced effectiveness of \texttt{TLRF} is attributed not only to its ability to aggregate data and perform cross-validation across both time and space but also to its adaptive weighting schemes. It is noteworthy that all benchmarking methods examined in this section rely on non-adaptive OLS weights (\ref{olsweight}). For example, K-Means Dynamic Window method alternates between different OLS weight assignments $\{\hat{\mu}_{t',c}\}_{t'\in \{t-\delta_{t,k}+2,\dots,t\}}$ through varying the fitting window size hyperparameter $\delta_{t,k}$. In contrast, the weights of \texttt{TLRF} estimator, i.e., $\{{\gamma_{t',c'}}({\Vec{X}_{t,c}})\}_{t'\in [t],c\in C}$, are fully adaptive and are not constrained by the OLS weighting schemes. The feature adaptive nature of \texttt{TLRF} provides an additional edge for its comparison with other transfer learning algorithms, an aspect into which we delve deeper in \S\ref{benchmark_TL}.

%By way of contrast, the performance evaluation of \texttt{TLRF} using exclusively time-varying features is presented in \S\ref{TLGRFTO}. This analysis underscores the importance of such features in the estimation process.

% \input{Tables/Source_of_Benefit_GRFs_Table}

% The properties of these estimators are compared to those of \texttt{TLRF} in Table \ref{tab:Source_of_Benefit_GRFs}. 

%\zw{There is no direct implementation of GRF}

% This section estimates the parameter of interest, i.e., $\bm{r_{t,c}}$, by directly implementing the classical GRF R-package \citep[c.f.][]{grfpackage}. Specifically, the estimation uses the function \textit{causal\_forest(\Vec{X}, Y, W)} in this package, where $\Vec{X}$ are the relevant features (e.g., $\bm{\Vec{X}_{t,c}}$ in our case); $Y$ is the dependent variable (e.g., $\ln(\bm{I_{s,c}})$  in our case); $W$ is the independent variable (e.g., $\bm{s}$ in our case). Particularly, when $W$ takes non-binary value, as is $\bm{s}$ in our case, the function \textit{causal\_forest(\Vec{X}, Y, W)} estimates the conditional average partial effect, i.e., 
% $$\frac{\mathbb{Cov}[Y,W|\Vec{X}]}{\mathbb{Var}[W|\Vec{X}]}$$

% overlap assumption \citep[c.f.][]{d2021overlap}

% concrete implementation of the classical GRF \citep[c.f.][]{athey2019generalized}

%This section elaborates on the distinct strengths of \texttt{TLRF} by providing a detailed comparison with other GRF implementations in instantaneous growth rate estimations.\par 

\section{Performance Evaluation Against Other Transfer Learning Methods}\label{benchmark_TL}

% Our identification framework also enables a direct comparison between \texttt{TLRF} and other transfer learning methods. Specifically, Corollary \ref{T2} establishes that our parameter of interest $r({\Vec{X}}_{t,c})$ is identified by the observable conditional mean (\ref{OLSID}):
% \[
% {r({\Vec{X}}_{t,c})}=\mathbb{E}[\ln(\bm{I_{t,c}})-\ln(\bm{I_{t-1,c}})|\Vec{X}_{t,c}].
% \]
% As such, we are able to compare \texttt{TLRF} with any transfer learning methods whose target tasks can be transformed into the conditional mean format (\ref{OLSID}).\par

The identification framework developed in this paper enables comprehensive comparison between \texttt{TLRF} and alternative transfer learning approaches. Through Corollary \ref{T2}, we establish that our parameter of interest $r({\Vec{X}}_{t,c})$ is identified by the observable conditional mean (\ref{OLSID}):
\[
{r({\Vec{X}}_{t,c})}=\mathbb{E}[\ln(\bm{I_{t,c}})-\ln(\bm{I_{t-1,c}})|\Vec{X}_{t,c}]
\]
This formulation allows direct comparison with any transfer learning method whose target task can be transformed into this conditional mean format. In this section, we leverage this framework to conduct such a comparison to demonstrate the robustness of our \texttt{TLRF} method.\par

% with the LASSO-based estimator (\texttt{LASSOTL}) introduced by \cite{bastani2021predicting}, a prominent method in the transfer learning literature.\par

%For this comparison, we transformed the target task of \texttt{LASSOTL} into the observable conditional mean format (\ref{OLSID}).

We benchmark \texttt{TLRF}  against a prominent transfer learning method: the LASSO-based estimator (\texttt{LASSOTL}) proposed by \cite{bastani2021predicting}. Specifically, by denoting 
$${Y_{gold}}=\ln({I_{t,c}})-\ln({I_{t-1,c}}),\ \ \Vec{X}_{gold}=\Vec{X}_{t,c},$$
and
$${\vec{Y}_{proxy}}=[\ln({I_{t',c'}})-\ln({I_{t'-1,c'}})]_{(t',c')\neq (t,c)},\ \ \mathcal{X}_{proxy}=[\Vec{X}_{t',c'}]_{(t',c')\neq (t,c)},$$
we implemented the \texttt{LASSOTL} estimator in two stages:
\begin{align*}
\mathbf{Step\ 1:}&\ \text{given regularization parameter $\mu$,}\\
&\hat{\vec{\beta}}_{proxy} = \text{arg}\min_{\vec{\beta}} \left\{
    \frac{1}{t|C|-1} \left\| {\vec{Y}_{proxy}} - \mathcal{X}^T_{proxy} \vec{\beta} \right\|_2^2 + \mu\|\vec{\beta}\|_{1}
\right\} \\
\mathbf{Step\ 2:}&\ \text{given regularization parameter $\lambda$,}\\
&\hat{\vec{\beta}}_{joint} = \text{arg}\min_{\vec{\beta}} \left\{
    \left\| {Y_{gold}} - \Vec{X}^T_{gold} \vec{\beta} \right\|_2^2
    + \lambda \left\| \vec{\beta} - \hat{\vec{\beta}}_{proxy} \right\|_1
\right\},
\end{align*}
and obtain the \texttt{LASSOTL} estimator:
$${\hat{r}^{LASSOTL}_{t,c}}=\Vec{X}^T_{gold} \hat{\vec{\beta}}_{joint}.$$
To evaluate performance, we generate 7-day-ahead predictions of log incident cases $\ln(\hat{I}^{LASSOTL}_{t,c})$ and assess accuracy using RMSE and MAE metrics, maintaining consistency with our established evaluation framework. Table \ref{tab:LASSOTL} presents the comparative results. Additional details of the implementation and tuning process can be found in Appendix \ref{apd:LASSOTL}.
\begin{table}[ht]
  \centering
\begin{tabular}{ccc}
    \hline
    \textbf{Method} & \textbf{MAE} & \textbf{RMSE} \\
    \hline
    \texttt{LASSOTL} & 0.167 & 0.230 \\
    \texttt{TLRF} & \textbf{0.126} & \textbf{0.195} \\
    \hline
    \end{tabular}
\caption{Median MAE and RMSE of \texttt{TLRF} vs. \texttt{LASSOTL}}
\label{tab:LASSOTL}
\end{table}

The empirical results demonstrate \texttt{TLRF}'s superior performance in COVID-19 exponential growth rate estimation tasks. This advantage stems from \texttt{TLRF}'s adaptive weighting mechanism, which proves particularly effective for dynamic, heterogeneous epidemiological data. In contrast, \texttt{LASSOTL}'s reliance on feature sparsity assumptions may constrain its ability to capture the full complexity of epidemic dynamics.\par

This section shows that feature-adaptive \texttt{TLRF} performs competitively against other transfer learning algorithms. We conduct a case study in \S\ref{CS}, which embeds these \texttt{TLRF} growth-rate estimates within a compartmental model to assess whether the statistical improvements translate into better mechanistic forecasts and policy-relevant metrics.

\section{Case Study: Allocating Resources for COVID-19 Outbreak Investigations in Colorado}\label{CS}

This section presents a case study demonstrating how \texttt{TLRF} can assist policymakers in their outbreak responses. Specifically, we compare the outbreak investigation decisions made by the Colorado Department of Public Health and Environment (CDPHE) \citep[c.f.][]{dphe_2021} with those recommended by \texttt{TLRF}. Through these analyses, we intend to demonstrate how \texttt{TLRF} can help quantify and operationalize CDPHE's investigation priorities.

%we chose Colorado for this case study because Colorado has an excellent track record of outbreak investigation reporting \citep[c.f.][]{dphe_2021}, allowing for an extensive evaluation of our methodology.

% The case study involves comparing the outbreak investigation decisions made by the Colorado Department of Public Health and Environment (CDPHE) with the outbreak investigation recommendations provided by TLRF. We will also explain

% to illustrate how \texttt{TLRF} can aid policymakers in expediting their response to outbreaks. Specifically, this case study compares 

% for opening investigations for potential outbreaks in many counties in that state. Specifically, we identify those counties where an outbreak investigation was opened by the CDPHE and compare them against those other counties where our algorithm predicted that the potential risk of an outbreak is higher. We report positive predictive value and true negative detection rates of the decisions made by the CDPHE as well as those recommended by \texttt{TLRF}.\par

%thus giving us the greatest range of time to evaluate our methodology.

\subsection{Analyzing CDPHE's Investigation Decisions}

We begin by illustrating the outbreak investigation process of CDPHE. By law, ``all suspected outbreaks must be reported to Public Health (CDPHE) as soon as possible" in Colorado \citep{epc2020covid}. 
Once CDPHE receives the reports, they must determine which outbreaks to investigate. 
In general, CDPHE would prioritize their investigations for outbreaks with ``concerning characteristics" such as ``a high attack rate, the introduction of a new variant of concern, higher than expected severity among cases"\citep{Outbreak_Report_Form_2023}. 
Additionally, CDPHE tends to prioritize outbreak investigations for at-risk populations, such as those in healthcare and correctional settings \citep{covid19colorado_priority}.

% We chose Colorado for this case study because Colorado has an excellent track record of case data reporting and outbreak record-keeping, allowing for a more extensive evaluation of our methodology.
% Specifically, 

\begin{figure}[!htpb]
\centering
\includegraphics[width=\linewidth]{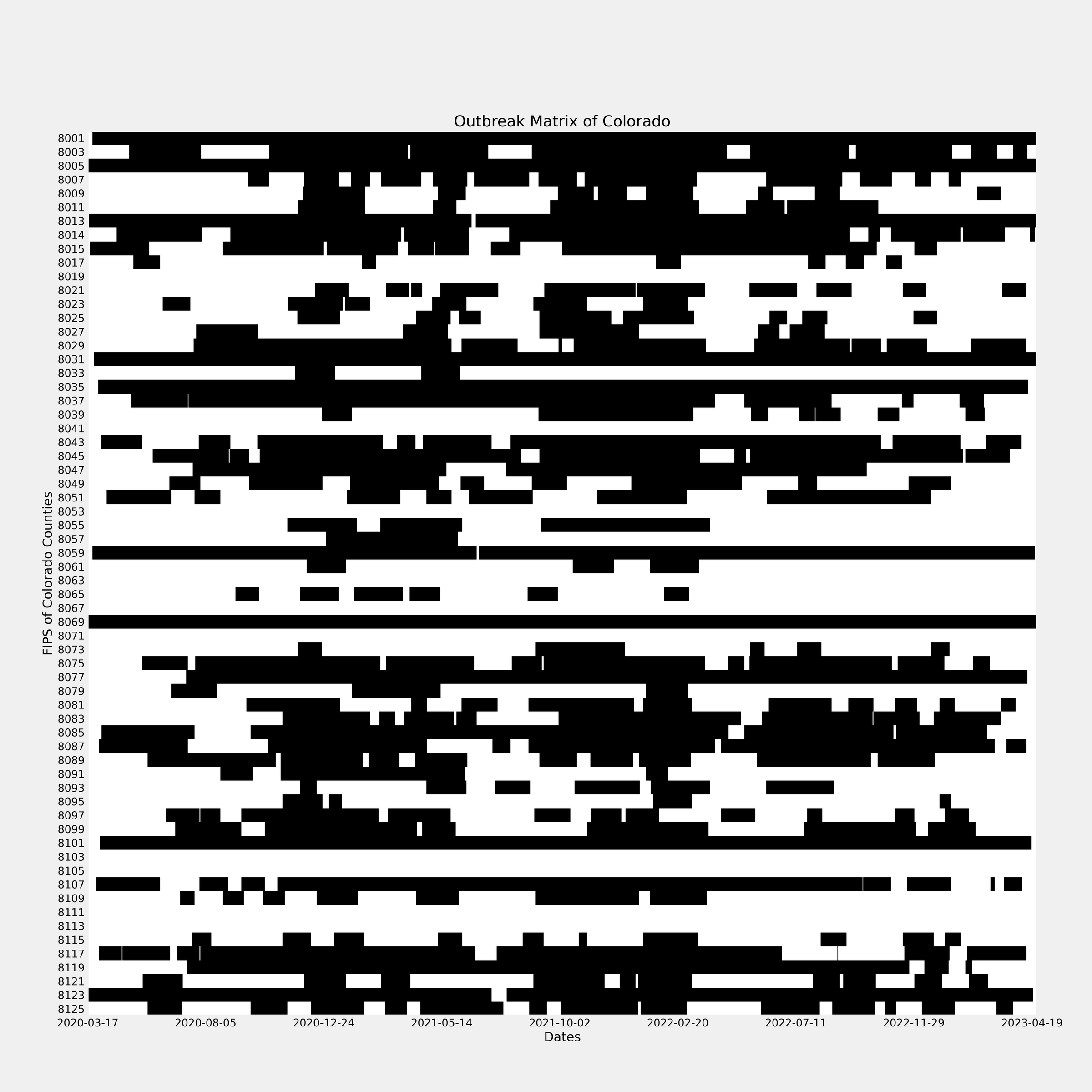}
\caption{Outbreak Matrix of Colorado Counties, where the rows are each county (indicated by its FIPS code) and each column is a point in time. Black indicates that the county is under investigation of a suspected outbreak and white otherwise.}
\label{fig:Outbreak_Matrix}
\end{figure}

To analyze CDPHE's investigation decisions, we first summarize its investigation record \citep[c.f.][]{dphe_2021} in Figure \ref{fig:Outbreak_Matrix}, which indicates the start and end dates of each outbreak investigation conducted. For our study purposes, we are interested in the days when an investigation starts in
a county, marked by a transition from white to black in Figure \ref{fig:Outbreak_Matrix}. We refer to these days as \textit{decision points}. As demonstrated in Appendix \ref{apd:Case_Study_DP}, out of the 1,020 days in our updated study period (from 2020-03-17 to 2022-12-31), only 159 days are decision points when CDPHE initiated new investigations. Additionally, out of these 159 decision points, CDPHE initiated only one investigation 86\% (137/159) of the time.

The observed sparse pattern in CDPHE's investigation frequency suggests a potential investigation resource constraint. In fact, as part of the procedure for investigating new outbreaks, a rapid response team often needs to be dispatched by CDPHE to conduct on-site testing \citep{metz2022investigation}. 
However, the number of rapid response teams contracted by CDPHE is limited and thus cannot be dispatched on a daily basis \citep{BidNet_2022}. 
In other words, an investigation decision can be made only on specific days when relevant resources are available.

% CDPHE did not initiate new investigations on a daily basis. 
% More precisely, out of the 1,020 days in our updated study period (from 2020-03-17 to 2022-12-31), CDPHE only conducted new investigations on 159 days. 

% At each decision point, we observed that they alloted $n \geq 1$ investigations

It is thus reasonable to model CDPHE's outbreak investigation decisions under a resource allocation framework. Specifically, given its investigation priority and resource constraints, one can evaluate whether CDPHE's investigation decisions were justifiable in a data-driven way. If they are not, it may be possible to improve these decisions by prioritizing another county with the help of \texttt{TLRF}. In the following section, we explore these pertinent questions and demonstrate how \texttt{TLRF} can add value in this context.\par

\subsection{Operationalizing CDPHE's Investigation Priorities}

This section demonstrates how \texttt{TLRF} can help CDPHE to operationalize one of its main investigation priorities, i.e., outbreaks with ``a high attack rate". Specifically, we will examine whether the counties that CDPHE chose to investigate actually had a higher increase in incident cases compared to other counties in the following week. Should this not be the case, we will proceed to evaluate the effectiveness of \texttt{TLRF} in accurately predicting the counties with higher attack rates for the upcoming week. Such prediction could provide valuable insight to CDPHE in terms of allocating their resources more efficiently for investigation purposes. Lastly, we also discuss the potential impact of \texttt{TLRF}'s investigation recommendations on CDPHE's other investigation priorities.

Formally, we formulate CDPHE's investigation resource allocation as a binary classification problem. Specifically, a True Positive (TP) result means that the investigated county did indeed have the most serious outbreak. In contrast, a False Positive (FP) result means that the investigated county was not the one with the most serious outbreak. That is, there was at least another county with a similar profile of at-risk populations but a greater increase in incident cases in the following week but was not chosen by the CDPHE for investigation. For the negative predictions, a False Negative (FN) result means that the county which was not investigated turned out to be the one with the most serious outbreak, and a True Negative (TN) result means that the non-investigated counties were not among those with the greatest increase in incident cases in the following week. On times when more than one county was investigated, the corresponding overlap between the ground truth and decisions are categorized into the various TP, FP, TN, FN as above.

\begin{table}[!htpb]
\begin{minipage}{.5\linewidth}
\centering
\begin{tabular}{c >{\bfseries}r @{\hspace{0.7em}}c @{\hspace{0.4em}}c @{\hspace{0.7em}}l}
  \multirow{10}{*}{\rotatebox{90}{\parbox{1.4cm}{\bfseries\centering Investigated?}}} & 
    & \multicolumn{2}{c}{\bfseries Most Serious Outbreak?}\\
  & & \bfseries Yes & \bfseries No \\
  & Yes & \MyBox{TP}{33} & \MyBox{FP}{149} \\[2.4em]
  & No & \MyBox{FN}{149} & \MyBox{TN}{1956}\\
\end{tabular}
\label{tab:stage1_confusion}
\end{minipage}%
\quad
\begin{minipage}{.5\linewidth}
\centering
\begin{tabular}{c >{\bfseries}r @{\hspace{0.7em}}c @{\hspace{0.4em}}c @{\hspace{0.7em}}l}
  \multirow{10}{*}{\rotatebox{90}{\parbox{1.4cm}{\bfseries\centering Investigated?}}} & 
    & \multicolumn{2}{c}{\bfseries Most Serious Outbreak?}\\
  & & \bfseries Yes & \bfseries No \\
  & Yes & \MyBox{TP}{107} & \MyBox{FP}{75} \\[2.4em]
  & No & \MyBox{FN}{75} & \MyBox{TN}{2030}\\
\end{tabular}
\label{tab:stage2_confusion}
\end{minipage}
\caption{Comparison of CDPHE's decisions (left) vs \texttt{TLRF}'s recommendations (right) as confusion matrices.}
\label{tab:combined_confusion}
\end{table}

%Our decision increased the True Positive Rate (TPR) from $21.74\%$ to $57.39\%$ and the True Negative Rate (TNR) from $88.43\%$ to $93.70\%$. Our decisions hence obtained a relative increase of $164\%$ for TPR and $5.96\%$ for TNR over Colorado DPHE's decisions

The evaluation results of CDPHE's decisions are summarized by the left confusion matrix of Table \ref{tab:combined_confusion}. Specifically, out of 182 counties CDPHE chose to investigate, only 33 of them turned out to experience the most serious outbreaks in the following week, corresponding to a positive predictive value (PPV) of $18.1\%$. In other words, at approximately four out of five investigation attempts, CDPHE did not allocate its scarce resources optimally to investigate the most serious outbreaks.\par

Next, we demonstrate how \texttt{TLRF} can improve CDPHE's investigation decisions.
Specifically, at each decision point, we first use \texttt{TLRF} to estimate the rate of change in incident cases, i.e.,
\begin{equation}\label{DecisionCreteria}
    \frac{dI_{t,c}}{dt} = \hat{r}_{t,c}^{TLRF} I_{t,c},
\end{equation}
for all counties currently not under investigation. This rate (\ref{DecisionCreteria}) is determined by multiplying \texttt{TLRF}'s estimated instantaneous exponential growth rate ($\hat{r}_{t,c}^{TLRF}$) by the current reported incident case numbers ($I_{t,c}$). Subsequently, we choose the county to investigate based on CDPHE's investigation capacity at this decision point. For example, if CDPHE has initiated one investigation at this decision point, we recommend investigating the county with the highest estimated rate (\ref{DecisionCreteria}). Similarly, on decision points when CDPHE investigates two or three counties, \texttt{TLRF}'s recommendation would be to investigate the counties with the top two or three highest estimated rates (\ref{DecisionCreteria}). As such, \texttt{TLRF} recommends the same number of investigations that CDPHE actually conducts at each decision point. Therefore, \texttt{TLRF}'s recommendations are comparable to CDPHE's decisions.\par

%Similarly, if CDPHE has initiated two investigations at this decision point, we recommend investigating the counties with the highest and second highest estimated rates (\ref{DecisionCreteria}).

% the number of investigations recommended by \texttt{TLRF} always matches the number of investigations actually conducted by CDPHE at each decision point. Therefore, \texttt{TLRF}'s recommendations always respect CDPHE's capacity constraint and are thus comparable to CDPHE's decisions.\par 

% rank these rate estimates of (\ref{DecisionCreteria}) and see if 

% for all counties not under investigation at this decision point. Lastly, if CDPHE's choices

% compare the estimates of (\ref{DecisionCreteria}) of CDPHE's choices with these estimates of other  Lastly, we replace CDPHE's choices with counties exhibiting highest difference 

% prioritize investigations on counties demonstrating the highest estimated

As shown in the right confusion matrix of Table \ref{tab:combined_confusion}, CDPHE's decisions can be substantially improved upon with the help of \texttt{TLRF}. Specifically, if CDPHE chose to investigate the county recommended by \texttt{TLRF}, its detection rate could be improved by $224\%$ (from 33 to 107 out of 182). We remark that while our estimation of FP (also TP) rate for the CDPHE is exact, the estimation for the CDPHE's FN rate is a lower bound. This is because at each decision point, if CDPHE chooses to investigate a county that does not have the highest growth rate in the following week (i.e., a FP case), it implies that CDPHE fails to investigate at least one more county with a higher growth rate in the following week (i.e., at least one or more FN cases). In other words, we assume that for each FP case, there is exactly one FN case while in reality we only know that there is \textit{at least} one FN case. Hence, our estimation for CDPHE's FN rate is likely an underestimation and TN rate is an overestimation. Yet, even with this conservative approach, we observe that in addition to improving the number of true positives by $224\%$, the TLRF improves the true negative rate by $3.51\%$ (from 1956 to 2030 out of 2105).\par

We further observe that, in addition to improving the overall prediction accuracy, \texttt{TLRF} also improves the prediction time of the most serious outbreaks. For example, on July 5th 2020, CDPHE recommended an investigation of Jefferson County, which uncovered an outbreak at a Chick-fil-A (an American fast food chain) with 2 confirmed cases and no deaths. In contrast \texttt{TLRF} recommended an investigation of Mesa County, whereas CDPHE did choose to investigate Mesa one week later. This county-level outbreak was later linked to a local-level outbreak in the St Mary's Medical Center and the Grandview Care Lodge Assisted Living facility \citep{zorn_2020} with 13 confirmed cases and 2 deaths. One week later on July 13th 2020, CDPHE did decide to investigate Mesa County, while \texttt{TLRF} continued to recommend Mesa County's investigation.\par

% \subsection{analyzing CDPHE's Other Investigation Priorities}

\begin{table}[htpb]
\centering
\begin{tabular}{cccc}
\toprule
Column & Total Population & Nursing Home Beds & Inmates \\
\midrule
CDPHE mean          & 36322.53 & 103.29 & 379.17 \\
TLRF mean          & 45167.48 & 72.37 & 182.37 \\
\midrule
t-statistic         & -1.02 & 1.01 & 1.60 \\
Standard Error      & 8684.54 & 30.50 & 122.73 \\
$95\%$ CI & (-25866.66, 8176.76) & (-28.86, 90.69) & (-43.74, 437.34) \\
\midrule
p-value             & 0.31 & 0.31 & 0.11 \\
\bottomrule
\end{tabular}
\caption{To compare the at-risk population profile of counties selected for investigation by CDPHE and \texttt{TLRF}, this table presents the means of nursing homes, prisons and county population of their investigation choices, respectively. Results for two sample $t$-tests between these means are also presented, where a large $p$-value (typically greater than $0.05$) means that CDPHE and \texttt{TLRF}'s investigation choices are not statistically different in terms of the at-risk population profile. $n=182$ for both CDPHE and \texttt{TLRF}}
\label{tab:ranking_table}
\end{table}

Next, we discuss the impact of \texttt{TLRF}'s recommendations on CDPHE's other investigation priorities. Table \ref{tab:ranking_table} shows that counties recommended by \texttt{TLRF} are similar to those investigated by CDPHE in terms of demographic characteristics such as nursing homes, correctional facilities, and total populations. Therefore, \texttt{TLRF}'s recommendations are consistent with CDPHE's investigation prioritization, which focuses resources on outbreaks among at-risk populations \citep[c.f.][]{dphe_2021}. A more detailed comparison of the decisions made by CDPHE and \texttt{TLRF} can be found in Appendix \ref{apd:Case_Study_populous}.

Finally, we estimate the basic reproduction number, $R_0$, for our Colorado case study by incorporating the $r_{t,c}$ estimates from \texttt{TLRF} into a compartmental model. The results presented in Table $\ref{tab:R0_estimates}$ demonstrate that our TLRF method effectively identifies counties with higher transmission potential, showing a mean $R_0$ of 2.37 in TLRF-investigated counties compared to the state-wide average of 1.14. This significant difference validates our method's effectiveness in early outbreak detection, as it successfully identifies counties with more intensive disease transmission patterns than those identified through traditional surveillance methods (CDPHE: $R_0 = 1.45$).

% The results, summarized in Table $\ref{tab:R0_estimates}$, indicate that while the counties investigated by CDPHE generally exhibit a higher $R_0$ than the state average, the estimates derived from \texttt{TLRF} tend to exceed even those values on average. These findings further support the efficacy of our proposed investigation policy.

\begin{table}[h!]
\centering
\begin{tabular}{|c|c|}
\hline
\textbf{Counties}                          & \textbf{Mean $R_0$} \\ \hline
All counties in Colorado                 & 1.14               \\ \hline
Counties investigated by \texttt{TLRF}  & 2.37                \\ \hline
Counties investigated by CDPHE           & 1.45                \\ \hline
\end{tabular}
\caption{Mean $R_0$ Estimates for Colorado Counties derived from the SEIR-\texttt{TLRF} benchmark}
\label{tab:R0_estimates}
\end{table}

% Further implementation details and supplementary results are provided .

In summary, we have shown that \texttt{TLRF} serves not only as a robust growth rate estimator and incident case forecaster but also as a practical tool to assist decision-makers in evaluating and containing the disease spread. Additional analyses of this case study are provided in the Appendix. In Appendix \ref{apd:Case_Study_threshold}, we show how \texttt{TLRF} can be utilized to configure an absolute threshold policy for outbreak detections. In Appendix \ref{apd:coronaSEIR_R0}, we further demonstrate how the improved $r_{t,c}$ estimates obtained from \texttt{TLRF} can be utilized to enhance the predictions of compartmental models.

\section{A Decision Support Tool for Implementation: COVID-19 Outbreak Detection Tool}\label{DST}
We expanded on our research by developing a public decision support tool based on the \texttt{TLRF} algorithm presented in this study. The ``COVID-19 Outbreak Detection Tool" offers an interactive choropleth displaying county-specific doubling times, showcasing the duration required for daily incident COVID-19 case numbers in a county to double. Additionally, a user-friendly data explorer is available for users to access, examine, and download the data, estimates, and predictions utilized in our study. Released on September 15, 2020, the web platform (\url{www.covid19sim.org}) experienced consistent traffic from all 50 states in the US. Notably, public health officials from various U.S. states also utilized this platform. The tool was actively maintained and updated until March 2023, when the New York Times discontinued their county-level COVID-19 case reporting \citep{NYTimes2020}. While no longer actively updated, the platform remains accessible as a demonstration of how machine learning methods can be effectively deployed for public health decision support. During its active period, the tool served over 20,000 unique users and informed policy decisions across multiple states.\par

A detailed description of the tool's functionalities and its impact, along with specific usage details, is provided in Appendix \ref{apd:covid19_outbreak_detection_appendix}. This tool served as an essential resource, attracting significant user engagement, and was instrumental in assisting public health authorities in their decision-making processes during the ongoing pandemic.

%and peaked at 2011 visitors. Within the first 3 months, during the fall and winter 2020 outbreak period, the site averaged 102.8 daily visitors, with a max of 339. For the next 3 months, when the COVID situation is alleviated due to vaccination, the site averaged 35.9 daily visitors and peaked at 51, and overall, within the first year, once the pandemic has transitioned to an endemic status and new variants appeared, we averaged 25.5 daily visitors while peaking at 263 visitors. 

% What is the public feedback?

% Site visit statistics
% - Tool launched (September 15th 2020) before major outbreak in Fall 2020

% Write it out as a paragraph
% First two months of launch, the number is around
% On average how many, peak how many
% First 3 months ( Major outbreak )
% First 6 months (COVID situation is alleviated due to vaccination)
% First year (Pandemic to Endemic transition, outbreak detection)

% Mention in passing e.g.,"To our knowledge, we have been contacted blah blah blah"

% MERLOT
% Recognized by certain organizations 
% Cite the merlot website

% Maryland schools
% Used as a monitoring tool by the state of Maryland

\section{Discussion and Conclusion}\label{CF}
In this work, we developed a transfer learning algorithm, i.e., \texttt{TLRF}, which obtains reliable small area estimates for COVID-19 growth rates. 
By pooling proxy data across time and space, \texttt{TLRF} addresses a key challenge in small area case growth estimates, i.e., choosing the fitting window to adequately balance the speed-accuracy tradeoff in estimation. As showcased in our Colorado case study, \texttt{TLRF} can serve to substantially improve detection rates of the most serious outbreaks within a much earlier time frame. With more reliable estimates of small area case growth rates, public health officials could better allocate their limited resources in outbreak investigations.\par

In our ongoing commitment to keep the public informed and assist timely decision making by health and governing authorities, we have developed an online, interactive decision support tool to disseminate our findings. 
This tool has provided the doubling time of COVID-19 incident case numbers in each county since September 2020 and has attracted significant public attention. 
Our outbreak detection tools and resulting findings have been used by multiple states to track the spread of COVID-19 in different jurisdictions by policy-makers.\par

Beyond small-area estimation, \texttt{TLRF} serves as a general template for locally flexible, model-based prediction. Applications include real-time decision support in public health and environmental monitoring \citep[c.f.][]{fraser2007estimating,cori2013new}, as well as the estimation problem of dose-response models with subpopulation heterogeneity \citep[c.f.][]{slob2002dose}. Importantly, \texttt{TLRF} complements mechanistic models: when embedded within compartmental frameworks (SEIR), \texttt{TLRF} supplies data-driven, locally adaptive growth inputs while preserving the interpretability and policy relevance of the underlying dynamics, as demonstrated in our Colorado case study. This synergy helps delineate the method’s boundaries and its broader operational value.\par

Nevertheless, we emphasize that \texttt{TLRF} is an algorithm to estimate model-based parameters, such as the instantaneous local exponential growth rate. Its validity, therefore, hinges on the premise that the model is correctly specified. When the model is misspecified,  the estimate loses interpretability and any downstream predictions based on the estimated model are compromised.

%\acks{REDACTED for review stage.}

% Acknowledgments here
\ACKNOWLEDGMENT{%
This research was supported by the Ministry of Education, Singapore, under the Academic Research Fund (AcRF) Tier 1 (Grant No. 001548-00001), and by the Agency for Science, Technology and Research (A*STAR) National Science Scholarship.
}% Leave this (end of acknowledgment)

%\newpage

% Appendix here
% Options are (1) APPENDI\Vec{X} (with or without general title) or 
%             (2) APPENDICES (if it has more than one unrelated sections)
% Outcomment the appropriate case if necessary
%
% \begin{APPENDI\Vec{X}}{<Title of the Appendix>}
% \end{APPENDI\Vec{X}}
%
%   or 
%
% \begin{APPENDICES}
% \section{<Title of Section A>}
% \section{<Title of Section B>}
% etc
% \end{APPENDICES}

% References here (outcomment the appropriate case) 

% CASE 1: BiBTe\Vec{X} used to constantly update the references 
%   (while the paper is being written).
\bibliographystyle{informs2014} % outcomment this and next line in Case 1
\bibliography{mypaper} % if more than one, comma separated

% CASE 2: BiBTe\Vec{X} used to generate mypaper.bbl (to be further fine tuned)
%\input{mypaper} % outcomment this line in Case 2

\newpage

\renewcommand{\theHsection}{A\arabic{section}}
\begin{APPENDICES}
\section*{Appendix}
The appendix provides additional implementation details for the methods discussed in the main text, supplementary benchmarking results, and further analyses conducted on the case study.
It begins with Appendix \ref{apd:TLGRF_Analysis}, which details the implementation of \texttt{TLRF}, compares it with alternative approaches, and analyzes the robustness of our chosen method. Appendix \ref{GRF} analyzes the limitations of the generalized random forests (GRF) by \cite{athey2019generalized} in our context. Appendix \ref{proof} follows with additional proofs and analytical results for the lemmas and theorems introduced in the main text. \ref{adp:synthetic} presents the details of our synthetic experiment to assess the robustness of \texttt{TLRF} against model misspecification. In Appendix \ref{apd1}, we outline the implementation details of our benchmarks.
Appendix \ref{apd:dataset} provides information on the datasets used in our study and includes an analysis of the most important features.
Appendix \ref{apd:Case_Study} expands on the case study with additional policies and use cases for the \texttt{TLRF} estimator.
Finally, Appendix \ref{apd:covid19_outbreak_detection_appendix} discusses the COVID-19 Outbreak Detection Tool, an online decision support tool made available to the public.

%discussed in \S\ref{GRF} of the main text.

\section{Further Details Regarding TLRF: Fine-tuning and Analyses}\label{apd:TLGRF_Analysis}
This section provides additional details on the \texttt{TLRF} algorithm and investigation into other potential choices of implementation.
In Appendix \ref{apd1.TLRF}, we present the pseudo-code needed to implement \texttt{TLRF}.
Then, in Appendix \ref{apd:other_GRF}, we present, derive, and compare alternative TLRF-based implementations. Finally, Appendix \ref{apd:TLGRF_Hyperparameter_Tuning} examines the robustness of \texttt{TLRF} by analyzing how properties of the underlying estimator, such as leaf sizes and maximum tree depth, vary with different hyperparameter choices.

\subsection{Transfer Learning Random Forests (\texttt{TLRF}) Pseudo-code}\label{apd1.TLRF}
This subsection presents the pseudo-code required to implement \texttt{TLRF}.
The process begins with the normalization and initial estimation step, outlined in Algorithm \ref{alg:block}.
Following this, the transfer learning step is performed using a standard regression random forest algorithm \texttt{regression\_forest()} (Algorithm \ref{alg:grfreg}).

\ \\
\bigskip
\begin{algorithm}[H]
\label{alg:block}
\caption{Normalization and Initial Estimates}
\KwIn{$\{(I_{t',c'},{\Vec{X}_{t',c'}})\}_{t'\in \{1,\dots,t\}, c\in C}$}
\KwOut{$\{Feature[t',c']\}_{t'\in \{1,\dots,t\}, c'\in C}$, $\{Y[t',c',1], Y[t',c',0]\}_{t'\in \{1,\dots,t\}, c'\in C}$}
\For{$c' \leftarrow 1$ \KwTo $|C|$}{
%$I_{-1,c}\leftarrow 1$\;
    \For{$t' \leftarrow 1$ \KwTo $t$}{
        \tcc{Normalize the incident case numbers for each block}
        $Y[t',c',1] \leftarrow ln(I_{t',c'})-ln(I_{t'-1,c'})$\;
        $Y[t',c',0] \leftarrow 0$\;
        $W[t',c',1] \leftarrow 1$\;
        $W[t',c',0] \leftarrow 0$\;
        \tcc{Generate the initial OLS estimates for each block}
        $Dep \leftarrow \{Y[t',c',1],Y[t',c',0]\}$\;
        $Ind \leftarrow  \{W[t',c',1],W[t',c',0]\}$\;
        $\{r^{ols}_{t',c'}, \alpha^{ols}_{t',c'}\} \leftarrow OLS(Dep \sim Ind )$\;
        \tcc{Append the feature data for each block}
        $Feature[t',c'] \leftarrow \{ {\Vec{X}_{t',c'}}, r^{ols}_{t',c'}, \alpha^{ols}_{t',c'}, t \}$\;
        }
    }
\end{algorithm}

% \begin{algorithm}[H]
% \label{alg:grf}
% \caption{\texttt{TLRF} through the \texttt{causal\_forest()} implementation}
% \KwIn{$\{Feature[t',c']\}_{t'\in \{1,\dots,t\}, c'\in C}$, $\{Y[t',c',1], Y[t',c',0]\}_{t'\in \{1,\dots,t\}, c'\in C}$}
% \KwOut{$\{r^{GRF}_{t,c'}\}_{c'\in C}$}
% %\For{$t \leftarrow 1$ \KwTo $|T|$}{
%     \tcc{Compute the congruence classes} 
%     $[t]\leftarrow \{t'\in \{1,\dots,t\}| t \equiv t'\ (\textrm{mod}\ 2)\}$\;
%     %$[t-1]\leftarrow \{t'\in \{0,\dots,t\}| t-1 \equiv t'\ (\textrm{mod}\ 2)\}$\;
%     \tcc{Assign the feature variable}
%     %$x_{t'} = Feature[t',c]$ \textbf{ if }$t' \in [t]$ \textbf{ else }$Feature[t'+1,c]$\;
%     $\Vec{X} \leftarrow \{Feature[t',c'],Feature[t',c']\}_{t'\in [t],c'\in C}$\;
%     \tcc{Assign the outcome variable}
%     %$y_{t'}= Y[t',c,1] \textbf{ if } t' \in [t] \textbf{ else } Y[t',c,0]$\; 
%     $Y \leftarrow \{Y[t',c',1],Y[t',c',0]\}_{t'\in [t],c'\in C}$\;
%     \tcc{Assign the treatment variable}
%     %$w_{t'} = 1 $ \textbf{ if } $t' \in [t]$ \textbf{ else } 0 \;
%     $W \leftarrow \{W[t',c',1],W[t',c',0]\}_{t'\in [t],c'\in C}$\;
%     \tcc{Feed ($\Vec{X}$,Y,W) into the GRF algorithm and obtain final estimates}
%     $\{r^{GRF}_{t,c'}\}_{c'\in C} \leftarrow GRF(\Vec{X},Y,W)$\;
% \end{algorithm}

\begin{algorithm}[H]
\label{alg:grfreg}
\caption{\texttt{TLRF} through the \texttt{regression\_forest}() implementation}
\KwIn{$\{Feature[t',c']\}_{t'\in \{1,\dots,t\}, c'\in C}$, $\{Y[t',c',1], Y[t',c',0]\}_{t'\in \{1,\dots,t\}, c'\in C}$}
\KwOut{$\{r^{TLRF}_{t,c'}\}_{c'\in C}$}
%\For{$t \leftarrow 1$ \KwTo $|T|$}{
    \tcc{Compute the congruence classes} 
    $[t]\leftarrow \{t'\in \{1,\dots,t\}| t \equiv t'\ (\textrm{mod}\ 2)\}$\;
    %$[t-1]\leftarrow \{t'\in \{0,\dots,t\}| t-1 \equiv t'\ (\textrm{mod}\ 2)\}$\;
    \tcc{Assign the feature variable}
    %$x_{t'} = Feature[t',c]$ \textbf{ if }$t' \in [t]$ \textbf{ else }$Feature[t'+1,c]$\;
    $\Vec{X} \leftarrow \{Feature[t',c']\}_{t'\in [t],c'\in C}$\;
    \tcc{Assign the outcome variable}
    %$y_{t'}= Y[t',c,1] \textbf{ if } t' \in [t] \textbf{ else } Y[t',c,0]$\; 
    $Y \leftarrow \{Y[t',c',1]\}_{t'\in [t],c'\in C}$\;
    % \tcc{Assign the treatment variable}
    % %$w_{t'} = 1 $ \textbf{ if } $t' \in [t]$ \textbf{ else } 0 \;
    % $W \leftarrow \{W[t',c',1]\}_{t'\in [t],c'\in C}$\;
    \tcc{Feed ($\Vec{X}$,Y) into the random forest algorithm and obtain final estimates}
    $\{r^{TLRF}_{t,c'}\}_{c'\in C} \leftarrow regression\_forest(\Vec{X},Y)$.
\end{algorithm}

\ \\

\subsection{Other TLRF Based Implementations}\label{apd:other_GRF}
This subsection provides supplementary results for alternative implementation choices within the \texttt{TLRF} framework. We present analytical derivations, implementation details, and empirical comparisons against \texttt{TLRF}. First, in Appendix \ref{supplementary_time_only}, we investigate whether \texttt{TLRF}'s improvement over fixed and dynamic time-window methods stems solely from its advanced time-based cross-validation techniques or also from its integration of spatial information. 
To test this, we benchmark against the \texttt{TLRF Time Only} estimator, which excludes spatial information, and show that spatial information is critical for improved performance.
Second, in Appendix \ref{TLGRFDelta}, we examine the impact of varying the initial estimator's fitting window size ($\delta$), where \texttt{TLRF} uses the minimal size of $\delta=2$. 
We derive alternative estimators for other larger $\delta$ values, referred to as \texttt{TLRF} $\delta$, and empirically demonstrate that $\delta=2$ offers the best performance.

%Finally, Appendix \ref{LLF} evaluates the Local Linear Forest (LLF) as an alternative second-step estimator for \texttt{TLRF} following the data normalization step in Algorithm \ref{alg:block}. 
% We empirically compare the LLF-based implementation against \texttt{causal\_forest()} and \texttt{regression\_forest()} versions of \texttt{TLRF}.

%\clearpage
\subsubsection{TLRF vs TLRF Time Only Comparison}\label{supplementary_time_only}

We investigate whether the improvement of \texttt{TLRF} over fixed/dynamic time window methods is attributable solely to its advanced time-based cross-validation techniques or to its additional integration of spatial information.
To that end, we consider another potential choice of implementation, \texttt{TLRF Time Only}, that restricts \texttt{TLRF}'s access to spatial information.
By comparing this new benchmark with \texttt{TLRF} and the existing methods, we find that incorporating spatial information is essential for \texttt{TLRF}'s superior performance to fixed/dynamic time window methods.\\

We now describe the \texttt{TLRF Time Only} benchmark. 
This benchmark removes access to spatial information by having a \texttt{TLRF} trained separately for each county. That is, each county's incidence case history and time-varying features are fed separately into a \texttt{TLRF} training process, generating a unique estimator for every county every day. With over 1,000 days and 3,000 counties in our study sample, this experiment results in over 3 million separately trained \texttt{TLRF} estimators in total. We discuss and analyze its performance below.\\

\begin{table}[ht]
\centering
\begin{tabular}{ccc}
\hline
\bfseries Method & \bfseries MAE & \bfseries RMSE \\
\hline
 \texttt{TLRF Time Only} & 0.216 & 0.291 \\
\bfseries TLRF & \bfseries 0.126 & \bfseries 0.195 \\
\hline
\end{tabular}
    \caption{Median MAE and RMSE of \texttt{TLRF} vs. \texttt{TLRF Time Only}}
    \label{tab:TLGRF_Time_Only_Appendix}%
\end{table}

As indicated in Table \ref{tab:TLGRF_Time_Only_Appendix}, the performance superiority of \texttt{TLRF} over fixed/dynamic time window methods is not replicated by \texttt{TLRF Time Only}. 
Specifically, when the access to spatial information is restricted, \texttt{TLRF Time Only} is outperformed by both dynamic window approaches (i.e., \texttt{tcv} and \texttt{ctcv}) in terms of median RMSE.
In contrast, with access to spatial information, \texttt{TLRF} outperforms the dynamic window approaches in both metrics. Therefore, the improvement of \texttt{TLRF} over the fixed/dynamic window approaches is not solely due to smarter pooling over time, but from its access to spatial information.\\

To further investigate \texttt{TLRF Time Only}'s decrease in performance with respect to \texttt{TLRF}, we decided to inspect the mean and median depths of each tree in both \texttt{TLRF} and \texttt{TLRF Time Only}.\par

As seen in Figure \ref{fig:TLGRF_vs_TLGRF_Time_Only_Tree_Depth}, \texttt{TLRF Time Only}'s trees are considerably more shallow than that of \texttt{TLRF}'s, where the latter's depths are typically 10 times that of the former.
It is clear that the historical data of one county, which is what \texttt{TLRF Time Only} has access to, contains significantly less information than that of the entire country, which is utilized by \texttt{TLRF}.
This comparison shows that the lack of access to spatial information hinder \texttt{TLRF Time Only}'s ability to properly conduct transfer learning, resulting in decreased performance.\par

\begin{figure}[H]
  \centering
\includegraphics[width=\textwidth]{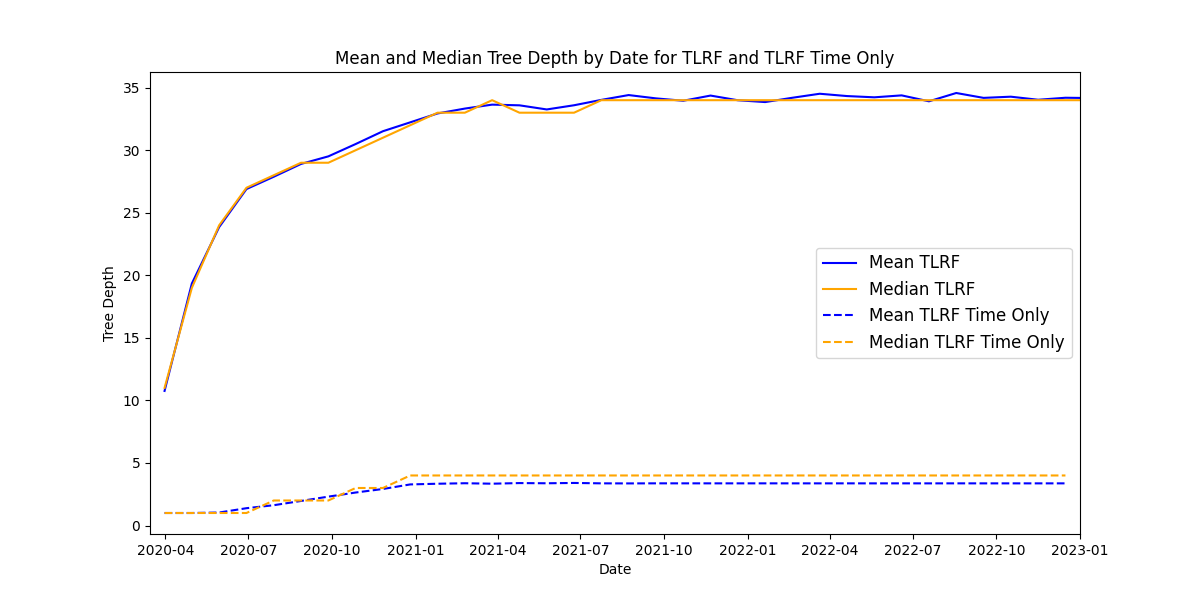}
  \caption{Mean and median tree depth by date (\texttt{TLRF} vs. \texttt{TLRF Time Only})}
   \label{fig:TLGRF_vs_TLGRF_Time_Only_Tree_Depth}
\end{figure}

%\subsubsection{TLRF vs TLRF $\delta$ Comparison}\label{supplementary_delta}

\subsubsection{TLRF $\delta$ Estimators}\label{TLGRFDelta}

To derive the \texttt{TLRF} $\delta$ estimators (\ref{GRFADDelta}), this section follows the same steps as in \S\ref{ME}. Specifically, we first characterize the conditional mean outcome corresponding to the initial estimator and subsequently the weighted conditional mean outcome that yields the \texttt{TLRF} $\delta$ estimators.\par

First, Lemma \ref{E5OLS} characterizes the conditional mean outcome of an initial estimator with a $\delta$-day fitting window. 
\begin{lemma}\label{E5OLS}
Under OLS Assumptions 1 and 2, an instantaneous county-level exponential growth rate ${r_{t,c}}$ with realized features $\Vec{X}_{t,c}$ can be estimated from data $ \{\ln({I_{t^-,c}}),\ln({I_{t^--1,c}}),\Vec{X}_{t^-,c}\}_{t^-\in \{t-\delta+2,\dots,t\}}$ using the OLS estimator with a $\delta$-day fitting window (\ref{rtcOLST5}). The corresponding conditional mean outcome of this estimator is
\begin{equation}\label{lcmE5}
   r({\Vec{X}}_{t,c})=\sum^{t}_{t^-=t-\delta+2} \hat{\mu}_{t^-,c} \mathbb{E}[(\ln(\bm{I_{t,c}})-\ln(\bm{I_{t-1,c}}))|\Vec{X}_{t,c}],
\end{equation}
where the weights $\{\hat{\mu}_{t^-,c}\}_{t^-\in \{t-\delta+2,\dots,t\}}$ are defined in (\ref{sampleweight}).
\end{lemma}
\noindent
Proof of Lemma \ref{E5OLS}:\\
By Theorems \ref{L1} and \ref{Restrict}, we can rewrite the conditional mean outcome corresponding to the OLS estimator with a $\delta$-day fitting window (\ref{OLSEQ}) as (\ref{lcmE5}).\QED

Second, Theorem \ref{E5T3} characterizes the weighted conditional mean outcomes that yield the \texttt{TLRF} $\delta$ estimators.
\begin{theorem}\label{E5T3}
For any county $c, c' \in C$ on any day $t'\in [t^-]$ with realized features $\Vec{X}_{t^-,c}$ and $\Vec{X}_{t',c'}$ such that $\Vec{X}_{t',c'}\in \underset{b\in B}{\cup} L_b(\Vec{X}_{t^-,c})$, its instantaneous county-level exponential growth rate ${r_{t,c}}$ can be estimated from data $\{ \{\ln({I_{t',c'}}),\ln({I_{t'-1,c'}}),\Vec{X}_{t',c'}\}_{t'\in [t^-],c'\in C}\}_{t^-\in \{t-\delta+2,\dots,t\}}$ via the weighted conditional outcomes:
\begin{equation}\label{E5GRFID}
     r({\Vec{X}}_{t,c})=\sum^{t}_{t^-=t-\delta+2} \hat{\mu}_{t^-,c}\sum_{c'\in C}\sum_{t'\in [t^-]}\gamma_{t',c'}(\Vec{X}_{t^-,c}) \mathbb{E}[(\ln(\bm{I_{t',c'}})-\ln(\bm{I_{t'-1,c'}}))|\Vec{X}_{t^-,c}], 
%\ \ \text{for all}\ \ x\in \text{Range}(\Vec{X}_i)
%$\{\{\bm{Y}_{t',c'}(t^-),{W}_{t',c'}(t^-), \bm{\Vec{X}_{t',c'}}\}_{t^-\in \{t'-1,t'\}}\}_{t'\in [t],c'\in C}$ via
\end{equation}
resulting in the following \texttt{TLRF} $\delta$ estimator:
\begin{equation}\label{E5T3r}
  \bm{\hat{r}^{TLRF(\delta)}_{t,c}}=  \sum^{t}_{t^-=t-\delta+2} \hat{\mu}_{t^-,c}\sum_{c'\in C}\sum_{t'\in [t^-]}\bm{\gamma_{t',c'}}({\Vec{X}_{t^-,c}}) \left(\ln(\bm{I_{t',c'}})-\ln(\bm{I_{t'-1,c'}})\right),
    % \hat{r}^{GRF}_{t,c}=\frac{\sum\limits_{t'\in [t]}\sum\limits_{c'\in C}\gamma_{t',c'}(x)\left(W_{t',c'}(t')-\frac{\sum\limits_{t'\in [t]}\sum\limits_{c'\in C} \gamma_{t',c'}(x)W_{t',c'}(t')}{2}\right)\left(\bm{Y}_{t',c'}(t')-\frac{\sum\limits_{t'\in [t]}\sum\limits_{c'\in C} \gamma_{t',c'}(x)\bm{Y}_{t',c'}(t')}{2}\right)}{\sum\limits_{t'\in [t]}\sum\limits_{c'\in C}\gamma_{t',c'}(x)\left(W_{t',c'}(t')-\frac{\sum\limits_{t'\in [t]}\sum\limits_{c'\in C}\gamma_{t',c'}(x) W_{t',c'}(t')}{2}\right)^2}.
\end{equation}
where weights $\{\hat{\mu}_{t^-,c}\}_{t^-\in \{t-\delta+2,\dots,t\}}$ are defined in (\ref{sampleweight}).
\end{theorem}
Proof of Theorem \ref{E5T3}:\\
By Theorem \ref{T3}, $\forall t^-\in \{t-\delta+2,\dots,t\}$, ${r_{t^-,c}}$ can be estimated from data $ \{\ln({I_{t',c'}}),\ln({I_{t'-1,c'}}),\Vec{X}_{t',c'}\}_{t'\in [t^-],c'\in C}$ via the weighted conditional mean outcome:
\begin{equation}\label{E5T3Eq1}
    r({\Vec{X}_{t^-,c}})=\sum_{c'\in C}\sum_{t'\in [t^-]}\gamma_{t',c'}(\Vec{X}_{t^-,c}) \mathbb{E}[(\ln(\bm{I_{t',c'}})-\ln(\bm{I_{t'-1,c'}}))|\Vec{X}_{t^-,c}].
\end{equation}
In addition, (\ref{E5T3Eq1}) implies (\ref{E5GRFID}) by Theorem \ref{Restrict}. Lastly, (\ref{E5T3r}) is the sample counterpart of (\ref{E5GRFID}). \QED

%We now present the MAE and RMSE plots of $\texttt{TLRF}(\delta=2)$ vs. $\delta\in\{3, 4,7,14\}$ mentioned in \S\ref{subsec.vsOLSweight}

%\input{Plots/TLGRF_Delta_MAE}

%\input{Plots/TLGRF_Delta_RMSE}

\textbf{Performance Comparison Against \texttt{TLRF} $\delta$}%\label{subsec.vsOLSweight}

We can thus extend Theorem \ref{T2} and Theorem \ref{T3} by allowing initial estimators to have fitting window sizes exceeding $\delta=2$. This extension essentially merges OLS weights with \texttt{TLRF} weighting schemes. Specifically, under the additional assumptions specified in Theorem \ref{E5OLS}, the initial estimator of \texttt{TLRF} $\delta$ becomes an OLS estimator with a $\delta$-day fitting window, i.e.,
\begin{equation}\label{TLGRFDeltaInitial}
    \bm{\hat{r}^{ols(\delta)}_{t,c}}=\sum^t_{t^-=t-\delta+2}\hat{\mu}_{t^-,c}\left(\ln(\bm{I_{t^-,c}})-\ln(\bm{I_{t^--1,c}}) \right),
\end{equation}
which coincides with the \texttt{TLRF} initial estimator (\ref{T2r}) only when $\delta=2$. Additionally, Theorem \ref{E5T3} demonstrates that these initial estimators yield the subsequent \texttt{TLRF} $\delta$ estimator, i.e., $\forall \delta\in \{2,\dots, t\}$,
\begin{equation}\label{GRFADDelta}
    \bm{\hat{r}^{TLRF(\delta)}_{t,c}}=  \sum^{t}_{t^-=t-\delta+2} \hat{\mu}_{t^-,c}\sum_{c'\in C}\sum_{t'\in [t^-]}\bm{\gamma_{t',c'}}(\bm{\Vec{X}_{t^-,c}}) \left(\ln(\bm{I_{t',c'}})-\ln(\bm{I_{t'-1,c'}})\right).
\end{equation}
where the non-adaptive weights $\{\hat{\mu}_{t^-,c}\}_{t^-\in \{t-\delta+2,\dots,t\}}$ are specified in (\ref{olsweight}).
Notably, the \texttt{TLRF} $\delta$ estimator (\ref{GRFADDelta}) reduces to the \texttt{TLRF} estimator \eqref{T3r} only when $\delta=2$.
%Detailed derivations of (\ref{GRFADDelta}) are presented in Appendix \ref{TLGRFDelta}.

As depicted in Table \ref{tab:TLGRF_wsize_table}, the median RMSE and MAE values of the \texttt{TLRF} algorithm are presented for various selections of $\delta$. 
Additionally, Figure \ref{fig:TLGRF_Delta_MAE_2x4} illustrates the daily mean MAE values corresponding to different choices of $\delta=2$ vs $\delta=4$ for clarity. 
%(see comparisons with other choices of $\delta$ in Appendix \ref{supplementary_delta}).
The numerical findings provide evidence of \texttt{TLRF}'s capacity to generalize effectively across diverse weighting schemes, with the results indicating that the most adaptable scheme, namely $\delta=2$, exhibits the most favorable empirical performance.
\begin{table}[ht]
  \centering
\begin{tabular}{ccc}
    \hline
    \textbf{Method} & \textbf{MAE} & \textbf{RMSE} \\
    \hline
  \texttt{TLRF}  14 & 0.173 & 0.237 \\
   \texttt{TLRF} 7 & 0.153 & 0.226 \\
  \texttt{TLRF}  4 & 0.138 & 0.217 \\
   \texttt{TLRF} 3 & 0.132 & 0.203\\
    \textbf{TLRF} ($\delta$=2) & \textbf{0.126} & \textbf{0.195} \\
    \hline
  \end{tabular}
  \caption{Median MAE and RMSE of \texttt{TLRF} vs. \texttt{TLRF} $\delta$: \texttt{TLRF} corresponds to \texttt{TLRF} $\delta$ with $\delta=2$.}
  \label{tab:TLGRF_wsize_table}
\end{table}
\begin{figure}[!htpb]
    \centering
    \resizebox{\linewidth}{!}{\includegraphics{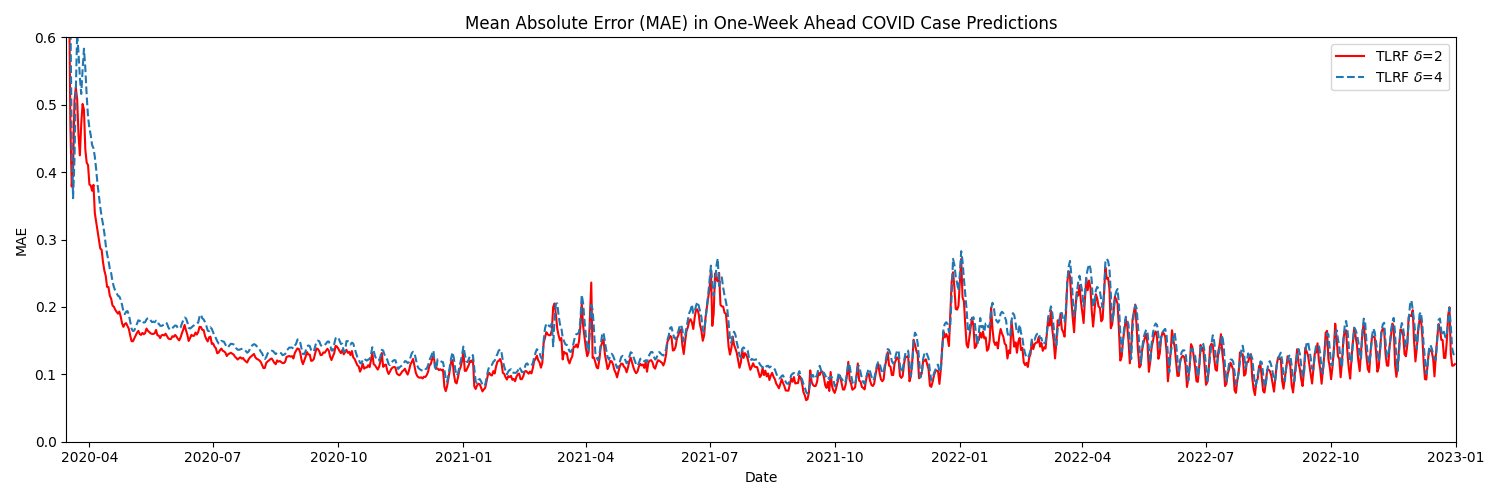}}
    \caption{MAE plot of prediction accuracy (\texttt{TLRF} with choices $\delta=2$ vs $\delta=4$)}
    \label{fig:TLGRF_Delta_MAE_2x4}
\end{figure}

\subsection{Robustness and Stability of the \texttt{TLRF} Algorithm}\label{apd:TLGRF_Hyperparameter_Tuning}
This subsection examines the robustness of the \texttt{TLRF} regarding its key hyperparameters and performance indicators, i.e., fitting window size for the initial estimator, number of trees, leaf node sizes, and tree depth.
% When investigating the stability of the \texttt{TGLRF} algorithm, the key properties that we are looking out for can be broken down into two categories, that is those that relate to: \zw{shorten this sentence.}
% \begin{itemize}
%     \item fitting window size for the initial estimator ($\delta$),
%     % \begin{itemize}
%     %     \item $\delta$
%     % \end{itemize}

%         \item Number of trees,
%         \item Leaf node sizes,
%         \item and Tree Depth.
%         %\item Honesty fraction
% \end{itemize}

\textbf{Fitting Window Size for the Initial Estimator ($\delta$)}\\
Results of extensive computational experiments on the impact of $\delta$ on \texttt{TLRF}'s performance are presented and discussed in Appendix \ref{TLGRFDelta}. These experiments reveal that the choice $\delta=2$ achieves the best empirical performance.\par

% For the fitting window sizes i.e., $\delta \geq 2$, we established in \S\ref{subsec.vsOLSweight} that our approach relies heavily on the adaptivity of the weighting scheme utilized, and we demonstrated that the most flexibly scheme possible is achieved under $\delta=2$. Empirical results obtained from trying out increasingly larger sizes of $\delta \geq 2$ as displayed in Table \ref{tab:TLGRF_wsize_table} lend further support to our analysis that $\delta=2$ is the optimal setting for the fitting window size for \texttt{TLRF}.\par

\textbf{Number of Trees}\\
For the number of trees in \texttt{TLRF}, we originally followed the default setting (i.e., $\texttt{num.trees}=2000$) in the \texttt{grf} package \citep[c.f.][]{athey2019generalized}. 
To seek more efficient implementations of \texttt{TLRF}, we have conducted additional computational experiments and found that there was no substantial improvement in the algorithm's empirical performance beyond 200 trees. 
Therefore, in the weekly update of our online decision support tool, we decided to keep the number of trees at 200 to alleviate the computational burden.\par

\textbf{Leaf Node Sizes}\\
Our \texttt{TLRF} algorithm adopts the default leaf node size figuration (i.e., $\texttt{min.node.size=5}$) in the \texttt{grf} package \citep[c.f.][]{athey2019generalized}. 
To investigate the potential impact of this hyperparameter on our algorithm's behavior, we have conducted additional computational experiments in this revision.
Specifically, we examine whether this minimum node size hyperparameter forces the split terminations in \texttt{TLRF}.
As illustrated in Figure \ref{fig:TLGRF_Leaf_Sizes_appendix}, we have not found any evidence of this nature.
Particularly, both the mean and median leaf node sizes observed in \texttt{TLRF} significantly surpass the default minimum node size, which suggests that this hyperparameter does not affect the quality of \texttt{TLRF}'s splits. Therefore, the performance of \texttt{TLRF} is not affected by the default setting of minimum node size.\par

\textbf{Tree Depth}\\
Sufficient tree depth within the \texttt{TLRF} algorithm is important to effectively identify meaningful subpopulations for splitting. 
However, excessively deep trees may risk overfitting to specific subpopulations. 
This section focuses on the evolution of trees in \texttt{TLRF} with increasing training data volume and assesses whether an adequate tree depth is achieved by this algorithm.
Figure \ref{fig:Tree_Depth_TLGRF_appendix} portrays a progression in tree depth of \texttt{TLRF}, initially averaging at a depth of 11 and eventually stabilizing around a depth of 33 as the dataset expanded. 
Additionally, Figure \ref{fig:TLGRF_Leaf_Sizes_appendix} reveals a consistency in both mean and median leaf node sizes of \texttt{TLRF} throughout our study period.
Observations from Figures \ref{fig:TLGRF_Leaf_Sizes_appendix} and \ref{fig:Tree_Depth_TLGRF_appendix} highlight that the splitting behavior of \texttt{TLRF} remains consistent even with increasing data fed into the algorithm. 
Such evidence supports the notion that \texttt{TLRF} has achieved an adequate tree depth, allowing it to accurately classify large volumes of data.\par

\begin{figure}[H]
  \centering
\includegraphics[width=0.8\textwidth]{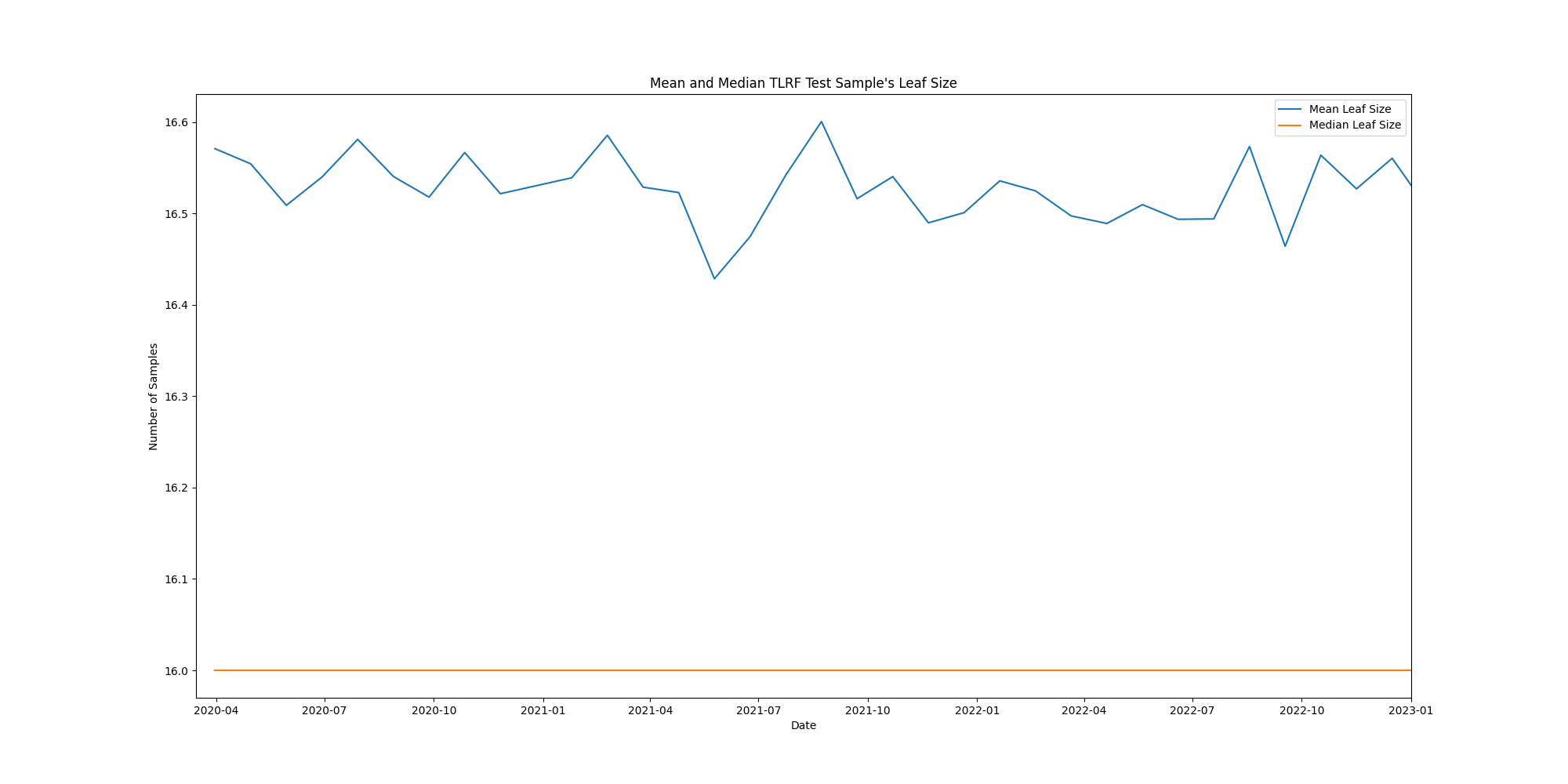}
  \caption{Mean and Median Leaf Sizes of \texttt{TLRF}}
   \label{fig:TLGRF_Leaf_Sizes_appendix}
\end{figure}

\begin{figure}[H]
  \centering
\includegraphics[width=0.8\textwidth]{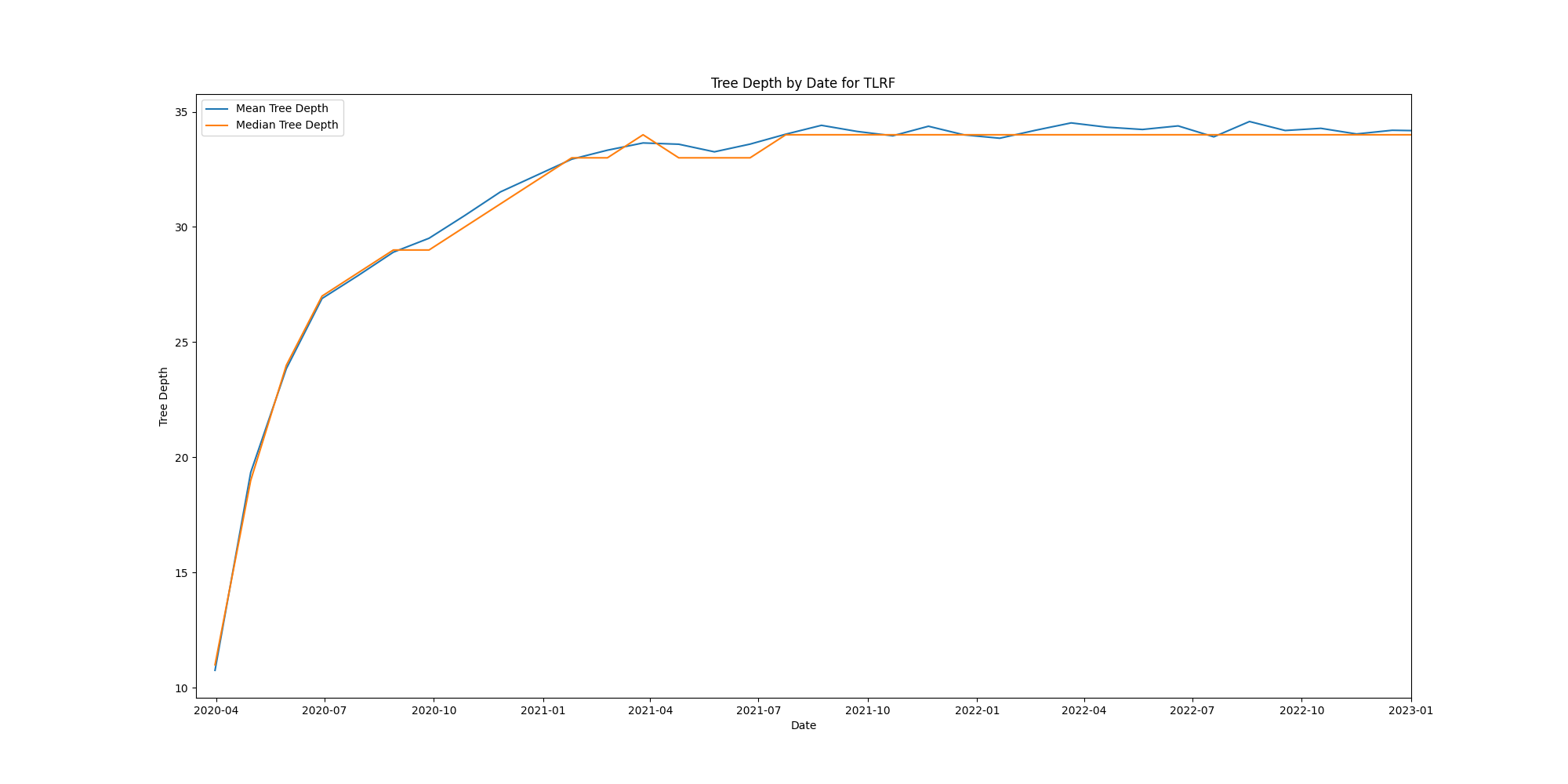}
  \caption{Mean and Median Tree Depth of \texttt{TLRF}}
   \label{fig:Tree_Depth_TLGRF_appendix}
\end{figure}

\section{Estimation of ${r_{t,c}}$ through Direct GRF Implementation}\label{GRF}

% \subsection{Direct Implementation Using the GRF Package}\label{subsec.vsGRF}

% The GRF framework, as implemented in the R package \texttt{grf}  \citep[c.f.][]{grfpackage}, supports two primary types of tasks: prediction tasks and estimation tasks. For prediction tasks, such as estimating conditional means of observable outcomes, functions like \texttt{regression\_forest()} implement the classical random forest methodology developed by \cite{breiman2001random}. For estimation tasks involving unobservable parameters identified through conditional mean outcomes, functions like \texttt{causal\_forest()} target conditional average partial effects (CAPEs) and other causal estimands.\par

% The county-level instantaneous exponential growth rate, ${r_{t,c}}$, is a latent parameter that is not directly observable in the data. Consequently, it is unsuitable for estimation using standard regression forest methods like \texttt{regression\_forest()}. 

This section estimates ${r_{t,c}}$ using an off-the-shelf GRF function \texttt{causal\_forest()}. Specifically, to estimate the actual growth model (\ref{observed}), a typical input to the \texttt{causal\_forest()} function is summarized by Table \ref{GRFdata}. The output of this function is the CAPE:
\[\frac{\mathbb{Cov}[\ln(\bm{I_{s,c}}), \bm{s}|{\Vec{X}_{t,c}}]}{\mathbb{Var}[\bm{s}|{\Vec{X}_{t,c}}]}, \numberthis \label{GRFcape}\]
representing the slope term estimate of the actual growth model (\ref{observed}). However, as we demonstrate both empirically and theoretically in the rest of this section, the estimate (\ref{GRFcape}) only identifies ${r_{t,c}}$ when features remain constant over time: $\Vec{X}_{t,c}=\Vec{X}_{c},\ \ \forall t\in \mathbb{Z}^+$.

\begin{table}[h!]
\centering
 \begin{tabular}{|c|c|c|}
\hline
\multirow{3}{*}{\begin{tabular}[c]{@{}c@{}}Dependent\\ Variable\\ ($\ln(\bm{I_{s,c}})$) \end{tabular}} & \multirow{3}{*}{\begin{tabular}[c]{@{}c@{}}Independent\\ Variable\\ ($\bm{s}$) \end{tabular}} & \multirow{3}{*}{\begin{tabular}[c]{@{}c@{}} \\ Feature \\ $(\bm{\Vec{X}_{s,c}})$ \end{tabular}}  \\
 & & \\
  & & \\
\hline
$\ln(I_{1,1})$ & $1$ & $\Vec{X}_{1,1}$ \\
\hline
$\ln(I_{2,1})$ & $2$ & $\Vec{X}_{2,1}$ \\\hline
$\ln(I_{3,1})$ & $3$ & $\Vec{X}_{3,1}$ \\
\hline
$\ln(I_{4,1})$ & $4$ & $\Vec{X}_{4,1}$ \\\hline
\vdots & \vdots & \vdots \\
\hline
% $\ln(I_{t,1})$ & $t$ & $x_{t,1}$ \\\hline
% $\ln(I_{1,2})$ & $1$ & $x_{1,2}$ \\\hline
% $\ln(I_{2,2})$ & $2$ & $x_{2,2}$ \\\hline
% \vdots & \vdots & \vdots \\
% \hline
\end{tabular}
\caption{Direct \texttt{causal\_forest()} Implementation}
\label{GRFdata}
\end{table}

First, we provide empirical evidence that a direct \texttt{causal\_forest()} implementation summarized by Table \ref{GRFdata} does not effectively utilize time-varying features. As illustrated in Figure \ref{fig:grf_tlgrf_mae} and Table \ref{tab:GRF_vs_TLGRF}, incorporating time-varying features into a direct \texttt{causal\_forest()} implementation yields only a marginal enhancement in performance. Moreover, it remains ambiguous whether this improvement is attributable to the effective utilization of temporal information. This ambiguity arises because all features employed in this study, whether time-invariant or time-varying, inherently contain spatial information. For example, the county-level vaccination coverage rate is a time-varying feature, yet its dynamics are significantly shaped by state-specific policies, which reveal geographic information. Consequently, the performance improvement observed when time-varying features such as vaccination coverage rate are included could also be ascribed to the spatial information embedded within these features.\par

%For more results on the performance comparison between \texttt{TLRF} and the direct \texttt{causal\_forest()} implementation, please refer to Appendix  \ref{apd1.GRF}.\par

\begin{figure}[!htpb]
    \centering
    \includegraphics[width=\textwidth]{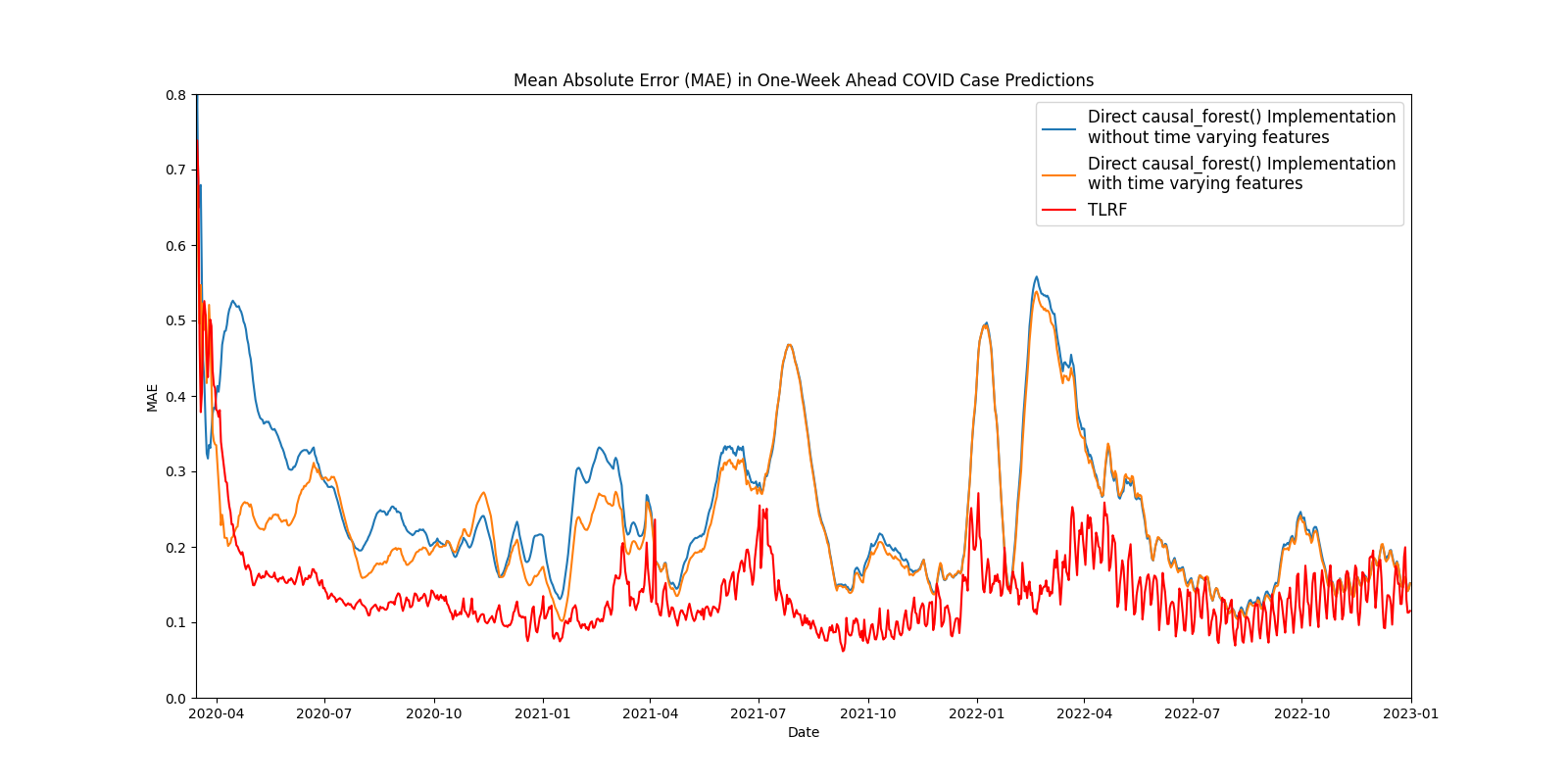}
    \caption{MAE plot of prediction accuracy (\texttt{TLRF} vs. Direct \texttt{causal\_forest()} Implementation with/without Time-varying Features)}
    \label{fig:grf_tlgrf_mae}
\end{figure}

\begin{table}[htpb!]
\centering
  \begin{tabular}{lcc}
  \toprule
  \bfseries Method &  \bfseries MAE &  \bfseries RMSE \\
  \midrule
Direct \texttt{causal\_forest()} Implementation without time varying features & 0.218 & 0.291 \\
Direct \texttt{causal\_forest()} Implementation with time varying features & 0.204 & 0.275\\
\textbf{TLRF}  & \textbf{0.126} & \textbf{0.195}\\
  \bottomrule
  \end{tabular}
    \caption{Median MAE and RMSE of \texttt{TLRF} vs.  Direct \texttt{causal\_forest()} Implementation with/without Time-varying Features}
    \label{tab:GRF_vs_TLGRF}%
\end{table}

Second, we theoretically reveal why the direct \texttt{causal\_forest()} implementation faces these limitations. Specifically, we note that
\begin{proposition}\label{GRFRestrict}
    The CAPE (\ref{GRFcape}) can be decomposed as
\begin{equation}\label{CAPEDecompose}
    \frac{\mathbb{Cov}[\ln(\bm{I_{s,c}}), \bm{s}|\bm{\Vec{X}}=\Vec{X}_{t,c}]}{\mathbb{Var}[\bm{s}|\bm{\Vec{X}}=\Vec{X}_{t,c}]}=\sum^t_{t^-=2}{\omega_{t^-,c}}\left( \mathbb{E}[\ln(\bm{I_{s,c}})|\bm{s}=t^-,\bm{\Vec{X}}=\Vec{X}_{t,c}]-\mathbb{E}[\ln(\bm{I_{s,c}})|\bm{s}=t^--1,\bm{\Vec{X}}=\Vec{X}_{t,c}]\right),
\end{equation}
where $\{{\omega_{t^-,c}}\}_{t^-\in \{1,\dots,t\}}$ are fixed weights. 
\end{proposition}
When the feature realization is time-varying, i.e., $\Vec{X}_{t',c}\neq \Vec{X}_{t,c}$, the conditional expectation $\mathbb{E}[\ln(\bm{I_{s,c}})|\bm{s}=t',\bm{\Vec{X}}=\Vec{X}_{t,c}]$ in (\ref{CAPEDecompose}) is not identifiable as the relevant data, i.e., $(\ln(I_{t',c}), t', \Vec{X}_{t,c})$, is not observable in Table \ref{GRFdata}. Consequently, incorporating time-varying features essentially compromises the assumption of population overlap, i.e., $\forall t'\in \{1,\dots,t\}$,
\begin{equation}\label{popoverlap}
    0<\mathbb{P}[\bm{s}=t',\bm{\Vec{X}}=\Vec{X}_{t,c}]<1,
\end{equation}
which is a key identification assumption of causal inference methods like \texttt{causal\_forest()} \citep{d2021overlap}. Therefore, certain constraints on the incorporated features are necessary to enable a direct \texttt{causal\_forest()} implementation for county-level instantaneous exponential growth rate estimations.\par 

%, when fed with data from Table \ref{GRFdata}

Notably, our experiment results from Figure \ref{fig:grf_tlgrf_mae} and Table \ref{tab:GRF_vs_TLGRF} suggest that the direct \texttt{causal\_forest()} implementation inherently constrains features to be time-invariant, i.e., $\forall c\in C$, $\forall t'\in \{1,\dots,t\}$, $\Vec{X}_{t',c}=\Vec{X}_{c}$. As such, the estimator corresponding to the direct \texttt{causal\_forest()} implementation is obtained by weighing the sample counterpart of CAPE (\ref{CAPEDecompose}) with the GRF similarity measure $\{\alpha_c'(\Vec{X}_{c})\}_{c'\in C}$, i.e.,
\begin{align*}
    \bm{\hat{r}^{GRF}_{t,c}}
    &=\sum\limits_{c'\in C}  \bm{\alpha_{c'}}(\Vec{X}_{c})\sum\limits^t_{t'=2}\hat{\mu}_{t',c}\left(\ln(\bm{I_{t',c'}})-\ln(\bm{I_{t'-1,c'}})\right), \numberthis \label{GRFEstimator}
\end{align*}
where 
\begin{equation*}
    \bm{\alpha_{c'}}(\Vec{X}_{c}):=\frac{1}{|B|}\sum\limits_{b=1}^{|B|}\frac{\mathbb{1} (\{{\Vec{X}_{c'}}\in L_b(\Vec{X}_{c})\})}{|L_b(\Vec{X}_{c})|}.
\end{equation*}
The formulation (\ref{GRFEstimator}) reveals that the direct \texttt{causal\_forest()} implementation is limited in its adaptability. Particularly, the OLS weights $\{\hat{\mu}_{t',c}\}_{t'\in \{2,\dots,t\}}$ inherited from the CAPE estimate limits the flexibility of the GRF framework.\\

\noindent
\textbf{Proof of Proposition \ref{GRFRestrict}}:\\
By denoting $\forall t^-\in \{t-\delta+2,\dots,t\}$,
\begin{equation*}
    \bm{r_{t^-,c}}:=\mathbb{E}[\ln(\bm{I_{s,c}})|\bm{s}=t^-,\bm{\Vec{X}_{t,c}}]-\mathbb{E}[\ln(\bm{I_{s,c}})|\bm{s}=t^--1,\bm{\Vec{X}_{t,c}}],
\end{equation*}
the CAPE (\ref{GRFcape}) can be rewritten as 
\begin{align*}
\mathbb{Cov}[\ln(\bm{I_{s,c}}),\bm{s}|{\Vec{X}_{t,c}}]&=\mathbb{E}[(\ln(\bm{I_{s,c}})-\mathbb{E}[\ln(\bm{I_{s,c}})|\bm{\Vec{X}_{t,c}}])(\bm{s}-\mathbb{E}[\bm{s}|\bm{\Vec{X}_{t,c}}])|\bm{\Vec{X}_{t,c}}]\\
    &=\mathbb{E}[\ln(\bm{I_{s,c}})(\bm{s}-\mathbb{E}[\bm{s}|\bm{\Vec{X}_{t,c}}])|\bm{\Vec{X}_{t,c}}]\\
    &=\mathbb{E}[\mathbb{E}[\ln(\bm{I_{s,c}})|\bm{s},\bm{\Vec{X}_{t,c}}](\bm{s}-\mathbb{E}[\bm{s}|\bm{\Vec{X}_{t,c}}])|\bm{\Vec{X}_{t,c}}]\\
    &=\mathbb{E}[(\bm{r_{i,c}}+\mathbb{E}[\ln(\bm{I_{s,c}})|\bm{s}=i-1,\bm{\Vec{X}_{t,c}}])(\bm{s}-\mathbb{E}[\bm{s}|\bm{\Vec{X}_{t,c}}])|\bm{\Vec{X}_{t,c}}]\\
    &=\mathbb{E}[(\sum^{i}_{t^-=t-\delta+2}\bm{r_{t^-,c}}+\mathbb{E}[\ln(\bm{I_{s,c}})|\bm{s}=t-\delta+1,\bm{\Vec{X}_{t,c}}])(\bm{s}-\mathbb{E}[\bm{s}|\bm{\Vec{X}_{t,c}}])|\bm{\Vec{X}_{t,c}}]\\
    &= \sum^t_{i=t-\delta+2}\mathbb{P}[\bm{s}=i|\bm{\Vec{X}_{t,c}}]\left(\sum^{i}_{t^-=t-\delta+2}\bm{r_{t^-,c}}+\mathbb{E}[\ln(\bm{I_{s,c}})|\bm{s}=t-\delta+1,\bm{\Vec{X}_{t,c}}]\right)(i-\mathbb{E}[\bm{s}|\bm{\Vec{X}_{t,c}}])\\
&+\mathbb{P}[\bm{s}=t-\delta+1|\bm{\Vec{X}_{t,c}}]\mathbb{E}[\ln(\bm{I_{s,c}})|\bm{s}=t-\delta+1,\bm{\Vec{X}_{t,c}}](t-\delta+1-\mathbb{E}[\bm{s}|\bm{\Vec{X}_{t,c}}])\\
&= \sum^t_{i=t-\delta+2}\sum^{i}_{t^-=t-\delta+2}\mathbb{P}[\bm{s}=i|\bm{\Vec{X}_{t,c}}]\bm{r_{t^-,c}}(i-\mathbb{E}[\bm{s}|\bm{\Vec{X}_{t,c}}])\\
&+ \sum^t_{i=t-\delta+1}\mathbb{P}[\bm{s}=i|\bm{\Vec{X}_{t,c}}] \mathbb{E}[\ln(\bm{I_{s,c}})|\bm{s}=t-\delta+1,\bm{\Vec{X}_{t,c}}](i-\mathbb{E}[\bm{s}|\bm{\Vec{X}_{t,c}}])\\
&= \sum^t_{t^-=t-\delta+2}\sum^{t}_{i=t^-}\mathbb{P}[\bm{s}=i|\bm{\Vec{X}_{t,c}}]\bm{r_{t^-,c}}(i-\mathbb{E}[\bm{s}|\bm{\Vec{X}_{t,c}}])\\
&+  \mathbb{E}[\ln(\bm{I_{\bm{s},c}})|\bm{s}=t-\delta+1,\bm{\Vec{X}_{t,c}}] \mathbb{E}[\bm{s}-\mathbb{E}[\bm{s}|\bm{\Vec{X}_{t,c}}]|\bm{\Vec{X}_{t,c}}]\\
&= \sum^t_{t^-=t-\delta+2}\bm{r_{t^-,c}} \sum^{t}_{i=t^-}\mathbb{P}[\bm{s}=i|\bm{\Vec{X}_{t,c}}](i-\mathbb{E}[\bm{s}|\bm{\Vec{X}_{t,c}}]).
\end{align*}
Lastly, by defining $\forall t^-\in \{t-\delta+2,\dots,t\}$,
$$\omega_{t^-,c}=\sum^{t}_{i=t^-}\mathbb{P}[\bm{s}=i|\bm{\Vec{X}_{t,c}}](i-\mathbb{E}[\bm{s}|\bm{\Vec{X}_{t,c}}]),$$
we have (\ref{CAPEDecompose}).\QED\\

\section{Proofs and Additional Theoretical Results}\label{proof}
This section provides additional proofs, assumptions, and theoretical justifications for the theorems in the main text.
It begins with Appendix \ref{CounterExamples}, which presents three counterexamples illustrating the role of Assumptions 1 and 2 in Theorem \ref{T1} for the identification purpose.
Next, Appendix \ref{apdMainTextProof} includes proofs for the main lemmas and theorems.
Following this, Appendix \ref{vsOLS} demonstrates how our identification assumptions are more flexible than the standard OLS assumptions.
Finally, Appendix \ref{AppendixBiasVariance} analyzes how the bias-variance tradeoff for the growth rate estimator depends on the fitting window size.

\subsection{Counterexamples}\label{CounterExamples}

\begin{counterexample}\label{Counter1}
Consider the following specifications of a potential growth model $(\ref{POt})$:
\begin{equation}\label{C1T4E1}
   \ln(\bm{I^2_{s,c}}) = 0.5 + 0.75 {s} + \bm{\varepsilon^2_{s,c}} 
\end{equation}
and 
\begin{equation}\label{C1T4E2}
    \ln(\bm{I^1_{s,c}}) =1 + 0.25{s}+ \bm{\varepsilon^1_{s,c}},
\end{equation}
where $\bm{\varepsilon^2_{s,c}}:=0.75 {s}- r(\bm{\Vec{X}_{s,c}}) {s}$ and $\bm{\varepsilon^1_{s,c}}:=0.25 {s}-  r(\bm{\Vec{X}_{s,c}}) {s}$. More precisely, the parameters in this counterexample are specified as ${r(\Vec{X}_{2,c})}=\mathbb{E}[r(\bm{\Vec{X}_{s,c}})|\bm{\Vec{X}_{s,c}}={\Vec{X}_{2,c}}]=0.75$, ${\alpha(\Vec{X}_{2,c})}=\mathbb{E}[{\alpha}(\bm{\Vec{X}_{s,c}})|\bm{\Vec{X}_{s,c}}={\Vec{X}_{2,c}}]=0.5$, ${r(\Vec{X}_{1,c})}=\mathbb{E}[{r}(\bm{\Vec{X}_{s,c}})|\bm{\Vec{X}_{s,c}}={\Vec{X}_{1,c}}]=0.25$,  ${\alpha(\Vec{X}_{1,c})}=\mathbb{E}[{\alpha}(\bm{\Vec{X}_{s,c}})|\bm{\Vec{X}_{s,c}}={\Vec{X}_{1,c}}]=1$. It is clear that error terms $\bm{\varepsilon^1_{s,c}}$ and $\bm{\varepsilon^2_{s,c}}$ explicitly depend on $s$ and $\bm{\Vec{X}_{s,c}}$.\par

We now verify that this specification satisfies \textbf{Assumptions 1}, i.e.,
$$\mathbb{E}[\ln(\bm{I^t_{s,c}})|\Vec{X}_{t,c}]=\alpha(\Vec{X}_{t,c})+ r(\Vec{X}_{t,c}){s}+\mathbb{E}[\bm{\varepsilon^t_{s,c}}|\Vec{X}_{t,c}] \ \ \forall s\in \mathbb{Z}^+,$$
and \textbf{Assumption 2}, i.e.,
$$\mathbb{E}[\ln(\bm{I^t_{s,c}})| \Vec{X}_{t,c}]=\mathbb{E}[\ln(\bm{I^{t-1}_{s,c}})|{\Vec{X}_{t-1,c}}].$$
First, we note that $\forall t^-\in \{1,2\}$,
$$\mathbb{E}[\bm{\varepsilon^{t^-}_{s,c}}| {\Vec{X}_{t^-,c}}]=0,$$
because
\begin{align*}
    \mathbb{E}[\bm{\varepsilon^2_{s,c}}| {\Vec{X}_{2,c}}]= \mathbb{E}[0.75{s}-r(\bm{\Vec{X}_{s,c}}){s}| \bm{\Vec{X}_{s,c}}={\Vec{X}_{2,c}}]=0.75{s}-0.75{s}=0
\end{align*}
and
\begin{align*}
    \mathbb{E}[\bm{\varepsilon^1_{s,c}}| {\Vec{X}_{1,c}}]= \mathbb{E}[0.25{s}-r(\bm{\Vec{X}_{s,c}}){s}| \bm{\Vec{X}_{s,c}}={\Vec{X}_{1,c}}]=0.25{s}-0.25{s}=0
\end{align*}
% \begin{align*}
%     \mathbb{E}[\bm{\varepsilon^1_{s,c}}|\bm{s}, {\Vec{X}_{1,c}}]= \mathbb{E}[0.25\bm{s}-r(\bm{\Vec{X}_{s,c}})\bm{s}|\bm{s}, {\Vec{X}_{1,c}}]=\mathbb{E}[\mathbb{E}[r(\bm{\Vec{X}_{s,c}})|\bm{\Vec{X}_{s,c}}={\Vec{X}_{1,c}}]\bm{s}-\mathbb{E}[r(\bm{\Vec{X}_{s,c}})|\bm{\Vec{X}_{s,c}}={\Vec{X}_{1,c}}]\bm{s}|\bm{s}, {\Vec{X}_{1,c}}]=0
% \end{align*}
% $$ \mathbb{E}[\bm{\varepsilon_{1,c}}|\bm{s}, \bm{\Vec{X}_{1,c}}]= \mathbb{E}[0.25\bm{s}-\bm{r}\bm{s}|\bm{s}, \bm{\Vec{X}_{1,c}}]=\mathbb{E}[\mathbb{E}[\bm{r}|\bm{\Vec{X}_{1,c}}]\bm{s}-\mathbb{E}[\bm{r}|\bm{\Vec{X}_{1,c}}]\bm{s}|\bm{s}, \bm{\Vec{X}_{1,c}}]=0.$$
As such, (\ref{C1T4E1}) and (\ref{C1T4E2}) satisfy \textbf{Assumptions 1} by their specifications. In addition, they satisfy \textbf{Assumption 2} because
$$\mathbb{E}[\ln(\bm{I^{2}_{s,c}})| {\Vec{X}_{2,c}}]=0.5 + 0.75 \times 1 = 1 + 0.25 \times 1= \mathbb{E}[\ln(\bm{I^{1}_{s,c}})|{\Vec{X}_{1,c}}].$$
\end{counterexample}
\ \\

\begin{counterexample}\label{Counter2}
Consider the following specifications of a potential growth model $(\ref{POt})$:
\begin{equation}\label{specification_T5E1}
   \ln(\bm{I^2_{s,c}}) = 0.5 + 0.75 {s} + \bm{\varepsilon^2_{s,c}} 
\end{equation}
and 
\begin{equation}\label{specification_T5E2}
    \ln(\bm{I^1_{s,c}}) =1 + 0.25{s}+ \bm{\varepsilon^1_{s,c}},
\end{equation}
where $\bm{\varepsilon^2_{s,c}}:=0.75 s- r(\bm{\Vec{X}_{s,c}}) {s^2}$ and $\bm{\varepsilon^1_{s,c}}:=0.25 s-  r(\bm{\Vec{X}_{s,c}}) {s}$. More precisely, the parameters in this counterexample are specified as ${r(\Vec{X}_{2,c})}=\mathbb{E}[r(\bm{\Vec{X}_{s,c}})|\bm{\Vec{X}_{s,c}}={\Vec{X}_{2,c}}]=0.75$, ${\alpha(\Vec{X}_{2,c})}=\mathbb{E}[{\alpha}(\bm{\Vec{X}_{s,c}})|\bm{\Vec{X}_{s,c}}={\Vec{X}_{2,c}}]=0.5$, ${r(\Vec{X}_{1,c})}=\mathbb{E}[{r}(\bm{\Vec{X}_{s,c}})|\bm{\Vec{X}_{s,c}}={\Vec{X}_{1,c}}]=0.25$,  ${\alpha(\Vec{X}_{1,c})}=\mathbb{E}[{\alpha}(\bm{\Vec{X}_{s,c}})|\bm{\Vec{X}_{s,c}}={\Vec{X}_{1,c}}]=1$. Clearly, this example satisfies \textbf{Assumption 2}:
$$\mathbb{E}[\ln(\bm{I^{2}_{1,c}})| {\Vec{X}_{2,c}}]=0.5 + 0.75 + (0.75-0.75) = 1 + 0.25 + (0.25-0.25)= \mathbb{E}[\ln(\bm{I^{1}_{1,c}})|{\Vec{X}_{1,c}}].$$
However, this example does not satisfy \textbf{Assumption 1} because the expectation of the potential growth model (\ref{specification_T5E1})'s error term conditional on $\bm{\Vec{X}_{s,c}}={\Vec{X}_{2,c}}$ is a function of $s$:
\begin{align*}
    \mathbb{E}[\bm{\varepsilon^2_{2,c}}| {\Vec{X}_{2,c}}]= \mathbb{E}[0.75{s}-r(\bm{\Vec{X}_{s,c}}){s^2}| \bm{\Vec{X}_{s,c}}={\Vec{X}_{2,c}}]=0.75{s}-0.75{s^2}.
\end{align*}
Finally, it is easy to check that \textbf{Theorem 1} is not applicable in this example as 
$$r(\Vec{X}_{2,c})=0.75$$
while
\begin{align*}
    &\mathbb{E}[\ln(\bm{I^2_{2,c}})|\Vec{X}_{2,c}]-\mathbb{E}[\ln(\bm{I^{1}_{1,c}})|{\Vec{X}_{1,c}}]\\
=&\left(0.5 + 0.75\times 2 + 0.75\times 2-0.75\times 4\right)-\left(1 + 0.25 + (0.25-0.25)\right)\\
=& -0.75.
\end{align*}
\end{counterexample}

\begin{counterexample}\label{Counter3}
Consider the following specifications of a potential growth model $(\ref{POt})$:
\begin{equation}\label{AET6E1}
   \ln(\bm{I^2_{s,c}}) = 0.5 + 0.75 {s} + \bm{\varepsilon^2_{s,c}} 
\end{equation}
and 
\begin{equation}\label{AET6E2}
    \ln(\bm{I^1_{s,c}}) =1 + 0.25{s}+ \bm{\varepsilon^1_{s,c}},
\end{equation}
where $\bm{\varepsilon^2_{s,c}}:=0.75 {s}- r(\bm{\Vec{X}_{s,c}}) {s}$ and $\bm{\varepsilon^1_{s,c}}:=0.25 {s}-  r(\bm{\Vec{X}_{s,c}}) {s}$. More precisely, the parameters in this counterexample are specified as ${r(\Vec{X}_{2,c})}=\mathbb{E}[r(\bm{\Vec{X}_{s,c}})|\bm{\Vec{X}_{s,c}}={\Vec{X}_{2,c}}]=0.75$, ${\alpha(\Vec{X}_{2,c})}=\mathbb{E}[{\alpha}(\bm{\Vec{X}_{s,c}})|\bm{\Vec{X}_{s,c}}={\Vec{X}_{2,c}}]=0.5$, ${r(\Vec{X}_{1,c})}=\mathbb{E}[{r}(\bm{\Vec{X}_{s,c}})|\bm{\Vec{X}_{s,c}}={\Vec{X}_{1,c}}]=0.25$,  ${\alpha(\Vec{X}_{1,c})}=\mathbb{E}[{\alpha}(\bm{\Vec{X}_{s,c}})|\bm{\Vec{X}_{s,c}}={\Vec{X}_{1,c}}]=1$.\par

First, to see that OLS Assumption 2 (OLS Assumption 3 in the original manuscript) is not satisfied by this specification, we note that
$$\mathbb{E}[\bm{\varepsilon_{2,c}}|{\Vec{X}_{2,c}}]=\mathbb{E}[\bm{\varepsilon^2_{2,c}}|{\Vec{X}_{2,c}}]=0\neq 0.25-0.75=\mathbb{E}[\bm{\varepsilon^1_{1,c}}|{\Vec{X}_{2,c}}]= \mathbb{E}[\bm{\varepsilon_{1,c}}|{\Vec{X}_{2,c}}].$$

Subsequently, we verify that this specification satisfies \textbf{Assumptions 1}, i.e.,
$$\mathbb{E}[\ln(\bm{I^t_{s,c}})|\Vec{X}_{t,c}]=\alpha(\Vec{X}_{t,c})+ r(\Vec{X}_{t,c}){s}+\mathbb{E}[\bm{\varepsilon^t_{s,c}}|\Vec{X}_{t,c}] \ \ \forall s\in \mathbb{Z}^+,$$
and \textbf{Assumption 2}, i.e.,
$$\mathbb{E}[\ln(\bm{I^t_{s,c}})| \Vec{X}_{t,c}]=\mathbb{E}[\ln(\bm{I^{t-1}_{s,c}})|{\Vec{X}_{t-1,c}}].$$
First, we note that $\forall t^-\in \{1,2\}$,
$$\mathbb{E}[\bm{\varepsilon^{t^-}_{s,c}}| {\Vec{X}_{t^-,c}}]=0,$$
because
\begin{align*}
    \mathbb{E}[\bm{\varepsilon^2_{s,c}}| {\Vec{X}_{2,c}}]= \mathbb{E}[0.75{s}-r(\bm{\Vec{X}_{s,c}}){s}| \bm{\Vec{X}_{s,c}}={\Vec{X}_{2,c}}]=0.75{s}-0.75{s}=0
\end{align*}
and
\begin{align*}
    \mathbb{E}[\bm{\varepsilon^1_{s,c}}| {\Vec{X}_{1,c}}]= \mathbb{E}[0.25{s}-r(\bm{\Vec{X}_{s,c}}){s}| \bm{\Vec{X}_{s,c}}={\Vec{X}_{1,c}}]=0.25{s}-0.25{s}=0.
\end{align*}
Lastly, this specification satisfies \textbf{Assumption 2} because
$$\mathbb{E}[\ln(\bm{I^{2}_{s,c}})| {\Vec{X}_{2,c}}]=0.5 + 0.75 \times 1 = 1 + 0.25 \times 1= \mathbb{E}[\ln(\bm{I^{1}_{s,c}})|{\Vec{X}_{1,c}}].$$
\end{counterexample}

\subsection{Proofs for Results Stated in the Main Text}\label{apdMainTextProof}
%     \bm{r_{t,c}}&=\mathbb{E}[\ln(\frac{\bm{I^t_{s,c}}}{\bm{I^t_{s-1,c}}}) | \bm{\Vec{X}_{t,c}}]\\
\noindent
Proof of Lemma \ref{LemmaPotential}:
\begin{align*}
{r_{t,c}} &= r({\Vec{X}_{t,c}})\\
% &=\mathbb{E}\left[\underset{T\rightarrow \infty}{lim}\frac{1}{T-1}\sum^{T}_{s=2}\left(\ln(\bm{I^t_{s,c}})-\ln(\bm{I^t_{s-1,c}})\right) \bigg| \Vec{X}_{t,c}\right]\\
     % &=\mathbb{E}[\ln(\bm{I^t_{s,c}})-\ln(\bm{I^t_{s-1,c}})| \bm{\Vec{X}_{t,c}}]\\
     &=
     \bigg(\alpha(\Vec{X}_{t,c})+ r(\Vec{X}_{t,c}){t}+\mathbb{E}[\bm{\varepsilon^t_{t,c}}|\Vec{X}_{t,c}]\bigg)-\bigg(\alpha(\Vec{X}_{t,c})+ r(\Vec{X}_{t,c}){(t-1)}+\mathbb{E}[\bm{\varepsilon^t_{t,c}}|\Vec{X}_{t,c}]\bigg)\\
     &=\mathbb{E}\left[\ln(\bm{I^t_{t,c}})-\ln(\bm{I^t_{t-1,c}})\bigg| \Vec{X}_{t,c}\right]\ \ \ \ \text{(by Assumption 1)}.
\end{align*}
%By the definition (\ref{cape}) of ${r_{t,c}}$, we have
%For $\forall s\in \mathbb{Z}^+$, we have
% which implies 
% $$\bm{r_{t,c}}=r(\bm{\Vec{X}_{t,c}})$$
% by Assumption 1 as
% $$\mathbb{E}[\ln(\bm{I^t_{s,c}})|\bm{s},\bm{\Vec{X}_{t,c}}]=\alpha(\bm{\Vec{X}_{t,c}})+ r(\bm{\Vec{X}_{t,c}})t'+\mathbb{E}[\bm{\varepsilon_{t,c}}|\bm{\Vec{X}_{t,c}}]$$
% and 
% $$\mathbb{E}[\ln(\bm{I^t_{s,c}})|\bm{s}=t'-1,\bm{\Vec{X}_{t,c}}]=\alpha(\bm{\Vec{X}_{t,c}})+ r(\bm{\Vec{X}_{t,c}})(t'-1)+\mathbb{E}[\bm{\varepsilon_{t,c}}|\bm{\Vec{X}_{t,c}}].$$
\QED

\noindent
Proof of Theorem \ref{T1}:\\
We aim to establish the equality (\ref{fw}) between the county-level instantaneous exponential growth rate and the observed log case number changes, i.e.,
$$r(\Vec{X}_{t,c})=\mathbb{E}[\ln(\bm{I^t_{t,c}})|\Vec{X}_{t,c}]-\mathbb{E}[\ln(\bm{I^{t-1}_{t-1,c}})|{\Vec{X}_{t-1,c}}].$$
To this end, we utilize the following decomposition:
\begin{align*}
 &\mathbb{E}[\ln(\bm{I^t_{t,c}})|\Vec{X}_{t,c}]-\mathbb{E}[\ln(\bm{I^{t-1}_{t-1,c}})|{\Vec{X}_{t-1,c}}]\\
 =& (\mathbb{E}[\ln(\bm{I^t_{t,c}})|\Vec{X}_{t,c}]- \mathbb{E}[\ln(\bm{I^t_{t-1,c}})|\Vec{X}_{t,c}])\\
 +& (\mathbb{E}[\ln(\bm{I^t_{t-1,c}})|\Vec{X}_{t,c}]-\mathbb{E}[\ln(\bm{I^{t-1}_{t-1,c}})|{\Vec{X}_{t-1,c}}]). \numberthis \label{T1P}
\end{align*}
By Lemma \ref{LemmaPotential}, the county-level instantaneous exponential growth rate ${r_{t,c}}$ is given by the first term in the decomposition (\ref{T1P}), i.e.,  $${r_{t,c}}=\mathbb{E}[\ln(\bm{I^t_{t,c}})|\Vec{X}_{t,c}]- \mathbb{E}[\ln(\bm{I^t_{t-1,c}})|\Vec{X}_{t,c}].$$
The second term in the decomposition (\ref{T1P}) is $0$, i.e.,
$$0=\mathbb{E}[\ln(\bm{I^t_{t-1,c}})|\Vec{X}_{t,c}]-\mathbb{E}[\ln(\bm{I^{t-1}_{t-1,c}})|{\Vec{X}_{t-1,c}}],$$
by Assumption 2 (Overlap). Therefore, the decomposition (\ref{T1P}) establishes the desired equality (\ref{fw}) under Assumptions 1 and 2. \QED\\

\noindent
Proof of Corollary \ref{T2}:

The conditional mean outcome  (\ref{OLSID}) is obtained by plugging in
$$\mathbb{E}[\ln(\bm{I^t_{t-1,c}})|\Vec{X}_{t,c}]=\mathbb{E}[\ln(\bm{I^{t-1}_{t-1,c}})|{\Vec{X}_{t-1,c}}]$$
from Assumption 2 (Overlap) into the equality (\ref{fw}) established by Theorem \ref{T1}. The initial estimator is the sample counterpart of the conditional mean outcome  (\ref{OLSID}). \QED\\

\noindent
Proof of Theorem \ref{T3}:\\
Following \S6.1 by \cite{athey2019generalized}, the estimator derived from the weighted conditional mean outcome (\ref{GRFID}), i.e.,
\begin{equation*}
r({\Vec{X}}_{t,c})= \sum_{t'\in [t]}\sum_{c'\in C}\gamma_{t',c'}(X_{t,c})\mathbb{E}[(\ln(\bm{I_{t',c'}})-\ln(\bm{I_{t'-1,c'}}))|\Vec{X}_{t,c}],
   % \sum\limits_{t'\in [t]}\sum\limits_{c'\in C}\gamma_{t',c'}(x_{t,c})\mathbb{E}[\left(\bm{Y_{t',c'}(s)}-r(x_{t,c}){W_{t',c'}(\bm{s})}\right) {W_{t',c'}(\bm{s})}|\bm{\Vec{X}} =x_{t',c'}]=0,
\end{equation*}
is a consistent estimator of ${r_{t,c}}$ provided that $\forall t'\in [t]$,  $\forall c'\in C$,
$$\mathbb{E}[(\ln(\bm{I_{t',c'}})-\ln(\bm{I_{t'-1,c'}}))|\Vec{X}_{t,c}]$$
is Lipschitz continuous in $\Vec{X}_{t,c}$. Clearly, this condition follows from Assumptions 2 and 3. In addition, the sample counterpart of (\ref{GRFID}) is \eqref{T3r}, i.e.,
\begin{equation*}
\bm{\hat{r}^{TLRF}_{t,c}}= \sum\limits_{t'\in [t]}\sum\limits_{c'\in C}  {\gamma_{t',c'}}(\bm{\Vec{X}_{t,c}}) \left(\ln(\bm{I_{t',c'}})-\ln(\bm{I_{t'-1,c'}})\right).
\end{equation*}  \QED

\noindent
Proof of Proposition \ref{GRFEquivalent}:
\begin{align*}
    &\bm{\hat{r}^{RF}_{t,c}}\\
    =&\frac{1}{|B|}\sum^{|B|}_{b=1} \sum\limits_{t'\in [t]}\sum\limits_{c'\in C} (\ln(\bm{I_{t',c'}})-\ln(\bm{I_{t'-1,c'}}))\frac{\mathbb{1} (\{{\Vec{X}_{t',c'}}\in L_b(\Vec{X}_{t,c})\})}{|L_b(\Vec{X}_{t,c})|}\\
    =&\sum\limits_{t'\in [t]}\sum\limits_{c'\in C} \frac{1}{|B|}\sum\limits_{b=1}^{|B|}\frac{\mathbb{1} (\{\bm{\Vec{X}_{t',c'}}\in L_b(\Vec{X}_{t,c})\})}{|L_b(\Vec{X}_{t,c})|}\left(\ln(\bm{I_{t',c'}})-\ln(\bm{I_{t'-1,c'}})\right)\\
    =& \sum\limits_{t'\in [t]}\sum\limits_{c'\in C}  \bm{\gamma_{t',c'}}(\Vec{X}_{t,c}) \left(\ln(\bm{I_{t',c'}})-\ln(\bm{I_{t'-1,c'}})\right)\\
    =&\bm{\hat{r}^{TLRF}_{t,c}}.
\end{align*}
\QED

\subsection{OLS Identification Assumptions of $r_{t,c}$}\label{vsOLS}

%provides a formal comparison of identification assumptions 1 and 2 with the OLS identification assumptions. We begin by deriving

%Theorem \ref{CompareOLS} establishes that the identification assumptions presented in Theorem \ref{T1} are, in fact, less restrictive than the OLS assumptions introduced in Theorem \ref{L1}. Lastly,

This section derives the OLS estimators of $\bm{r_{t,c}}$ for fixed fitting window sizes under standard OLS identification assumptions, as outlined in Theorem \ref{L1}. Following this, Theorem \ref{Restrict} highlights that the identification assumptions we employ facilitate the development of more versatile estimators of $\bm{r_{t,c}}$ in comparison to those derived from the OLS assumptions. Within this section, we employ the abbreviations $\bm{\alpha}$ and $\bm{r}$ to represent the random coefficients $\alpha(\bm{\Vec{X}})$ and $r(\bm{\Vec{X}})$, correspondingly, i.e.,
$$\bm{\alpha}:=\alpha(\bm{\Vec{X}})\ \ \text{and}\ \ \bm{r}:=r(\bm{\Vec{X}}).$$
The conditional expectations of these coefficients, given the relevant features at county $c$ on day $t$, are denoted as ${\alpha}({\Vec{X}_{t,c}})$ and ${r}({\Vec{X}_{t,c}})$, respectively, i.e.,
$${\alpha}({\Vec{X}_{t,c}}):=\mathbb{E}[\bm{\alpha} |{\Vec{X}_{t,c}}]\ \ \text{and}\ \ {r}({\Vec{X}_{t,c}}):=\mathbb{E}[\bm{r} |{\Vec{X}_{t,c}}].$$

We first state the identification assumptions for OLS estimators:\\
% \textbf{OLS Assumption 1} \textit{(Local Unconfoundedness)} $\forall t^-\in \{t-\delta+1,\dots,t\}$,
% $$\{\bm{\alpha},\bm{r} ,\bm{\varepsilon_{t^-,c}}\} \indep \bm{s}\ |\ \bm{\Vec{X}_{t^- ,c}};$$
%\ \ \mathbb{E}[\bm{s}|\bm{\Vec{X}_{t,c}}]=\mathbb{E}[\bm{s}|\bm{\Vec{X}_{t^-,c}}],
\textbf{OLS Assumption 1} \textit{(Homogeneity)} $\forall t^-\in \{t-\delta+1,\dots,t\}$,
$${\alpha}({\Vec{X}_{t,c}})={\alpha}({\Vec{X}_{t^-,c}}),\ \ {r}({\Vec{X}_{t,c}})={r}({\Vec{X}_{t^-,c}}),\ \ \mathbb{E}[\bm{\varepsilon_{t^-,c}}|{\Vec{X}_{t,c}}]=\mathbb{E}[\bm{\varepsilon_{t^-,c}}|{\Vec{X}_{t^-,c}}];$$
% $$\mathbb{E}[\bm{\alpha}|{\Vec{X}_{t,c}}]=\mathbb{E}[\bm{\alpha}|{\Vec{X}_{t^-,c}}],\ \ \mathbb{E}[\bm{r}|{\Vec{X}_{t,c}}]=\mathbb{E}[\bm{r}|{\Vec{X}_{t^-,c}}],\ \ \mathbb{E}[\bm{\varepsilon_{t^-,c}}|{\Vec{X}_{t,c}}]=\mathbb{E}[\bm{\varepsilon_{t^-,c}}|{\Vec{X}_{t^-,c}}];$$
\textbf{OLS Assumption 2} \textit{(Normalization)} $\forall t^-\in \{t-\delta+1,\dots,t\}$,
$$\mathbb{E}[\bm{\varepsilon_{t,c}}|{\Vec{X}_{t,c}}]=\mathbb{E}[\bm{\varepsilon_{t^-,c}}|{\Vec{X}_{t,c}}].$$
Subsequently, we demonstrate Theorem \ref{L1} that the identification of ${r_{t,c}}$ is also possible through these assumptions:
\begin{theorem}\label{L1}
Let $\delta \in \mathbb{Z}^{+}\backslash\{1\}$ be a fixed fitting window size, and $c \in C$ be a fixed county. 
The county-level instantaneous exponential growth rate ${r_{t,c}}$ is identified as
\[{r_{t,c}} =\frac{\sum_{t^-=t-\delta+1}^{t}(t^--\frac{\sum_{t^-=t-\delta+1}^{t}t^-}{\delta})(\mathbb{E}[\ln(\bm{I_{t^-,c}})|{\Vec{X}_{t,c}}] - \frac{\sum_{t^-=t-\delta+1}^{t}\mathbb{E}[\ln(\bm{I_{t^-,c}})|{\Vec{X}_{t,c}}]}{\delta})}{\sum_{t^-=t-\delta+1}^{t}\left(t^--\frac{\sum_{t^-=t-\delta+1}^{t}t^-}{\delta}\right)^2}.\numberthis\label{OLSEstimator}
\]

% \[
%     {r_{t,c}} =\frac{\mathbb{Cov}[\ln(\bm{I_{s,c}}), \bm{s}|{\Vec{X}_{t,c}}]}{\mathbb{Var}[\bm{s}|{\Vec{X}_{t,c}}]}.\numberthis\label{OLSEstimator}
% \]
% whose sample counterpart is
% \begin{equation}\label{P1P0}
%     \hat{r}_{t,c}^{ols(\delta)}=\frac{\sum\limits_{t^-=t-\delta+1}^{t}\sum\limits_{c\in C} \left[\left(t^--\frac{\sum\limits_{t^-=t-\delta+1}^t t^-}{\delta}\right)\left(ln(I_{t^-,c})-\frac{\sum\limits_{t^-=t-\delta+1}^{t}\sum\limits_{c\in C} ln(I_{t^-,c})}{\delta|C|}\right)\right]}{\sum\limits_{t^-=t-\delta+1}^t\left(t^--\frac{\sum\limits_{t^-=t-\delta+1}^t t^-}{\delta}\right)^2}.
% \end{equation}
%the equivalence between the random coefficient $r({\Vec{X}_{t,c}})$ and the county-level instantaneous exponential growth rate $\bm{r_{t,c}}$ demonstrated in
\end{theorem}

\noindent
Proof of Theorem \ref{L1}:\\
Our goal is to identify ${r_{t,c}}$ using the log case data $\{{ln(I_{t^-,c})}\}_{t^-\in \{t-\delta+1,\dots,t\}}$ from county $c$ within the fitting window of size $\delta$. To achieve this, we first observe that by \text{OLS Assumption 1}, the actual growth model (\ref{observed}) reduces to a homogeneous growth model, i.e., $\forall t^-\in \{t-\delta+1,\dots,t\},$
\begin{align*}
        \ln(\bm{I_{t^-,c}})&= {\alpha_{t,c}} + {r_{t,c}} {t^-} + \bm{\varepsilon_{t^-,c}}.
\end{align*}
By OLS Assumptions 1 and 2, we have
\begin{align*}
\mathbb{E}[ \ln(\bm{I_{t^-,c}})&- \bm{\alpha} - \bm{r} {t^-} - \bm{\varepsilon_{t^-,c}}|{\Vec{X}_{t,c}}]=0\ \ \ \ \forall t^-\in \{t-\delta+1,\dots,t\}\\
       \Rightarrow \sum^t_{t^-=t-\delta+1}t^-\mathbb{E}[\ln(\bm{I_{t^-,c}})|{\Vec{X}_{t,c}}]&= \sum^t_{t^-=t-\delta+1}t^-\mathbb{E}[\bm{\alpha}|{\Vec{X}_{t,c}}]\ + \sum^t_{t^-=t-\delta+1}(t^-)^2\mathbb{E}[\bm{r}|{\Vec{X}_{t,c}}] + \sum^t_{t^-=t-\delta+1}t^-\mathbb{E}[\bm{\varepsilon_{t^-,c}}|{\Vec{X}_{t,c}}].  \numberthis \label{L1E1}
\end{align*}
By subtracting 
\begin{align*}
    \frac{1}{\delta}\sum^t_{t^-=t-\delta+1}\mathbb{E}[\ln(\bm{I_{t^-,c}})|{\Vec{X}_{t,c}}]\sum^t_{t^-=t-\delta+1}t^-&=\frac{1}{\delta} \bigg(\delta \mathbb{E}[\bm{\alpha}|{\Vec{X}_{t,c}}]\sum^t_{t^-=t-\delta+1}t^-\\
    &+ \mathbb{E}[\bm{r}|{\Vec{X}_{t,c}}]\left(\sum^t_{t^-=t-\delta+1}t^-\right)^2 + \sum^t_{t^-=t-\delta+1}\mathbb{E}[\bm{\varepsilon_{t^-,c}}|{\Vec{X}_{t,c}}]\sum^t_{t^-=t-\delta+1}t^-\bigg)
\end{align*}
from (\ref{L1E1}), we have
\begin{align*}
   &\sum^t_{t^-=t-\delta+1}t^-\mathbb{E}[\ln(\bm{I_{t^-,c}})|{\Vec{X}_{t,c}}]-\frac{1}{\delta}\sum^t_{t^-=t-\delta+1}\mathbb{E}[\ln(\bm{I_{t^-,c}})|{\Vec{X}_{t,c}}]\sum^t_{t^-=t-\delta+1}t^-\\
   =&\sum^t_{t^-=t-\delta+1}t^-\mathbb{E}[\bm{\alpha}|{\Vec{X}_{t,c}}]-\frac{1}{\delta} \delta \mathbb{E}[\bm{\alpha}|{\Vec{X}_{t,c}}]\sum^t_{t^-=t-\delta+1}t^-\\
+&\sum^t_{t^-=t-\delta+1}(t^-)^2\mathbb{E}[\bm{r}|{\Vec{X}_{t,c}}]-\delta\mathbb{E}[\bm{r}|{\Vec{X}_{t,c}}]\left(\sum^t_{t^-=t-\delta+1}\frac{t^-}{\delta}\right)^2\\
   +& \sum^t_{t^-=t-\delta+1}t^-\mathbb{E}[\bm{\varepsilon_{t^-,c}}|{\Vec{X}_{t,c}}]- \frac{1}{\delta}\sum^t_{t^-=t-\delta+1}\mathbb{E}[\bm{\varepsilon_{t^-,c}}|{\Vec{X}_{t,c}}]\sum^t_{t^-=t-\delta+1}t^-.   \numberthis \label{L1E2}
\end{align*}

Finally, (\ref{L1E2}) implies (\ref{OLSEstimator}) because
\begin{align*}
&\sum^t_{t^-=t-\delta+1}t^-\mathbb{E}[\ln(\bm{I_{t^-,c}})|{\Vec{X}_{t,c}}]-\frac{1}{\delta}\sum^t_{t^-=t-\delta+1}\mathbb{E}[\ln(\bm{I_{t^-,c}})|{\Vec{X}_{t,c}}]\sum^t_{t^-=t-\delta+1}t^-\\
=& \sum^t_{t^-=t-\delta+1}t^-\mathbb{E}[\ln(\bm{I_{t^-,c}})|{\Vec{X}_{t,c}}]-\delta \sum^t_{t^-=t-\delta+1}\frac{\mathbb{E}[\ln(\bm{I_{t^-,c}})|{\Vec{X}_{t,c}}]}{\delta}\sum^t_{t^-=t-\delta+1}\frac{t^-}{\delta}\\
    =&\sum_{t^-=t-\delta+1}^{t}(t^--\frac{\sum_{t^-=t-\delta+1}^{t}t^-}{\delta})(\mathbb{E}[\ln(\bm{I_{t^-,c}})|{\Vec{X}_{t,c}}] - \frac{\sum_{t^-=t-\delta+1}^{t}\mathbb{E}[\ln(\bm{I_{t^-,c}})|{\Vec{X}_{t,c}}]}{\delta}), \numberthis \label{OLSFact1}
\end{align*}
\[\sum^t_{t^-=t-\delta+1}t^-\mathbb{E}[\bm{\alpha}|{\Vec{X}_{t,c}}]-\frac{1}{\delta} \delta \mathbb{E}[\bm{\alpha}|{\Vec{X}_{t,c}}]\sum^t_{t^-=t-\delta+1}t^-=0,\]
\[\sum^t_{t^-=t-\delta+1}(t^-)^2\mathbb{E}[\bm{r}|{\Vec{X}_{t,c}}]-\delta\mathbb{E}[\bm{r}|{\Vec{X}_{t,c}}]\left(\sum^t_{t^-=t-\delta+1}\frac{t^-}{\delta}\right)^2=\sum_{t^-=t-\delta+1}^{t}\left(t^--\frac{\sum_{t^-=t-\delta+1}^{t}t^-}{\delta}\right)^2r_{t,c},\numberthis \label{OLSFact2}\]
and
\begin{align*}
&\sum^t_{t^-=t-\delta+1}t^-\mathbb{E}[\bm{\varepsilon_{t^-,c}}|{\Vec{X}_{t,c}}]-\frac{1}{\delta}\sum^t_{t^-=t-\delta+1}\mathbb{E}[\bm{\varepsilon_{t^-,c}}|{\Vec{X}_{t,c}}]\sum^t_{t^-=t-\delta+1}t^-\\
=&\sum^t_{t^-=t-\delta+1}t^-\mathbb{E}[\bm{\varepsilon_{t,c}}|{\Vec{X}_{t,c}}]-\frac{1}{\delta}\sum^t_{t^-=t-\delta+1}\mathbb{E}[\bm{\varepsilon_{t,c}}|{\Vec{X}_{t,c}}]\sum^t_{t^-=t-\delta+1}t^-\ \ \ \ \ \ \ \ (\text{by OLS Assumption 2})\\
=&0,
\end{align*}
where (\ref{OLSFact1}) and (\ref{OLSFact2}) make use of the fact that, for $\{(x_i,y_i)\in \mathbb{R}^2|i=1,2,\dots,n\}$, we have 
$$\sum^n_{i=1}(x_i-\bar{x})(y_i-\bar{y})=\sum^n_{i=1} x_i y_i-n\bar{x}\bar{y}.$$
\QED

Subsequently, we show that our identification assumptions 1 and 2 allow for more adaptive estimators of the instantaneous county-level exponential growth rate compared to those derived from the OLS Assumptions 1 and 2.
\begin{theorem}\label{Restrict}
Any OLS estimators (\ref{OLSEstimator}) of ${r_{t,c}}$ with a $\delta$-day fitting window can be expressed as a convex combination of $\delta-1$ two-point estimators in the form (\ref{fw}), i.e., $\forall \delta\in \mathbb{Z}^{+}\backslash\{1\}$,
\begin{align*}
    &\frac{\sum_{t^-=t-\delta+1}^{t}(t^--\frac{\sum_{t^-=t-\delta+1}^{t}t^-}{\delta})(\mathbb{E}[\ln(\bm{I_{t^-,c}})|{\Vec{X}_{t,c}}] - \frac{\sum_{t^-=t-\delta+1}^{t}\mathbb{E}[\ln(\bm{I_{t^-,c}})|{\Vec{X}_{t,c}}]}{\delta})}{\sum_{t^-=t-\delta+1}^{t}\left(t^--\frac{\sum_{t^-=t-\delta+1}^{t}t^-}{\delta}\right)^2}\\
    =&\sum^t_{t^-=t-\delta+2}\hat{\mu}_{t^-,c}\left( \mathbb{E}[\ln(\bm{I_{t^-,c}})|{\Vec{X}_{t,c}}]-\mathbb{E}[\ln(\bm{I_{t^--1,c}})|{\Vec{X}_{t,c}}]\right), \numberthis \label{OLSEQ}
\end{align*}
where $\{\hat{\mu}_{t^-,c}\}_{t^-\in \{t-\delta+2,\dots,t\}}$ are non-negative weights that sum to 1, i.e.,
\begin{equation}\label{sampleweight}
    \hat{\mu}_{t^-,c}=  \frac{\sum^t_{i=t^-}(i-\frac{2t-\delta+1}{2})}{\sum^t_{i=t-\delta + 1}(i-(t-\delta+1))(i-\frac{2t-\delta+1}{2})}.
\end{equation}
As such, the sample counterpart of the OLS estimator (\ref{OLSEstimator}) with a $\delta$-day fitting window is
\begin{equation}\label{rtcOLST5}
    \bm{\hat{r}^{ols(\delta)}_{t,c}}=\sum^t_{t^-=t-\delta+2}\hat{\mu}_{t^-,c}\left(\ln(\bm{I_{t^-,c}})-\ln(\bm{I_{t^--1,c}}) \right).
\end{equation}

%\begin{equation}\label{OLSEQ1}
%     \bm{\mu_{t^-,c}}=\frac{ \sum^{t}_{i=t^-}\mathbb{P}[\bm{s}=i|\bm{\Vec{X}_{t,c}}](i-\mathbb{E}[\bm{s}|\bm{\Vec{X}_{t,c}}])}{\sum^{t}_{i=t-\delta+1}\mathbb{P}[\bm{s}=i|\bm{\Vec{X}_{t,c}}](i-(t-\delta+1))(i-\mathbb{E}[\bm{s}|\bm{\Vec{X}_{t,c}}])}.
% \end{equation}

%\mathbb{Cov}[\ln(\bm{I_{s,c}}),\bm{s}|\bm{\Vec{X}_{t,c}}]&=\mathbb{E}[(\ln(\bm{I_{s,c}})-\mathbb{E}[\ln(\bm{I_{s,c}})|\bm{\Vec{X}_{t,c}}])(\bm{s}-\mathbb{E}[\bm{s}|\bm{\Vec{X}_{t,c}}])|\bm{\Vec{X}_{t,c}}]\\

\end{theorem}
Proof of Theorem \ref{Restrict}:\\
First, the numerator of the OLS estimator (\ref{OLSEstimator}) can be rewritten as
\begin{align*}
&\sum_{t^-=t-\delta+1}^{t}(t^--\frac{\sum_{t^-=t-\delta+1}^{t}t^-}{\delta})(\mathbb{E}[\ln(\bm{I_{t^-,c}})|{\Vec{X}_{t,c}}] - \frac{\sum_{t^-=t-\delta+1}^{t}\mathbb{E}[\ln(\bm{I_{t^-,c}})|{\Vec{X}_{t,c}}]}{\delta})\\
    &=\sum_{t^-=t-\delta+1}^{t}\mathbb{E}[\ln(\bm{I_{t^-,c}})|{\Vec{X}_{t,c}}](t^--\frac{\sum_{t^-=t-\delta+1}^{t}t^-}{\delta}) \numberthis\label{T5E1}
    % &=\mathbb{E}[\mathbb{E}[\ln(\bm{I_{s,c}})|\bm{s},\bm{\Vec{X}_{t,c}}](\bm{s}-\mathbb{E}[\bm{s}|\bm{\Vec{X}_{t,c}}])|\bm{\Vec{X}_{t,c}}] 
\end{align*}
Additionally, by rewriting ${r_{t^-,c}}$ as (\ref{fw}), i.e., $\forall t^-\in \{t-\delta+2,\dots,t\}$,
\begin{equation}\label{OLSEQ2}
    {r_{t^-,c}}=\mathbb{E}[\ln(\bm{I_{t^-,c}})|{\Vec{X}_{t,c}}]-\mathbb{E}[\ln(\bm{I_{t^--1,c}})|{\Vec{X}_{t,c}}],
\end{equation}
we have $\forall \delta\in \mathbb{Z}^{+}\backslash\{1\}$, $\forall i\in \{t-\delta+2,\dots,t\},$
\begin{align*}
    \mathbb{E}[\ln(\bm{I_{i,c}})|{\Vec{X}_{t,c}}]&={r_{i,c}}+\mathbb{E}[\ln(\bm{I_{i-1,c}})|{\Vec{X}_{t,c}}]\\
    &=\sum^{i}_{t^-=t-\delta+2}{r_{t^-,c}}+\mathbb{E}[\ln(\bm{I_{t-\delta+1,c}})|{\Vec{X}_{t,c}}]. \numberthis\label{T5E2}
\end{align*}
By (\ref{T5E2}), (\ref{T5E1}) becomes 
\begin{align*}
& \sum_{t^-=t-\delta+1}^{t}(t^--\frac{\sum_{t^-=t-\delta+1}^{t}t^-}{\delta})(\mathbb{E}[\ln(\bm{I_{t^-,c}})|{\Vec{X}_{t,c}}] - \frac{\sum_{t^-=t-\delta+1}^{t}\mathbb{E}[\ln(\bm{I_{t^-,c}})|{\Vec{X}_{t,c}}]}{\delta})\\
&= \sum^t_{i=t-\delta+2}\left(\sum^{i}_{t^-=t-\delta+2}{r_{t^-,c}}+\mathbb{E}[\ln(\bm{I_{t-\delta+1,c}})|{\Vec{X}_{t,c}}]\right)(i-\frac{\sum_{i=t-\delta+1}^{t}i}{\delta})\\
&+\mathbb{E}[\ln(\bm{I_{t-\delta+1,c}})|{\Vec{X}_{t,c}}](t-\delta+1-\frac{\sum_{i=t-\delta+1}^{t}i}{\delta})\\
&= \sum^t_{i=t-\delta+2}\sum^{i}_{t^-=t-\delta+2}{r_{t^-,c}}(i-\frac{2t-\delta+1}{2})+ \sum^t_{i=t-\delta+1} \mathbb{E}[\ln(\bm{I_{t-\delta+1,c}})|{\Vec{X}_{t,c}}](i-\frac{2t-\delta+1}{2})\\
&=\sum^{t}_{t^-=t-\delta+2} \sum^t_{i=t^-}(i-\frac{2t-\delta+1}{2}){r_{t^-,c}}
% \sum^t_{t^-=t-\delta+2}\sum^{t}_{i=t^-}\mathbb{P}[\bm{s}=i|\bm{\Vec{X}_{t,c}}]\bm{r_{t^-,c}}(i-\mathbb{E}[\bm{s}|\bm{\Vec{X}_{t,c}}])\\
% &+  \mathbb{E}[\ln(\bm{I_{\bm{s},c}})|\bm{s}=t-\delta+1,\bm{\Vec{X}_{t,c}}] \mathbb{E}[\bm{s}-\mathbb{E}[\bm{s}|\bm{\Vec{X}_{t,c}}]|\bm{\Vec{X}_{t,c}}]\\
% &= \sum^t_{t^-=t-\delta+2}\bm{r_{t^-,c}} \sum^{t}_{i=t^-}\mathbb{P}[\bm{s}=i|\bm{\Vec{X}_{t,c}}](i-\mathbb{E}[\bm{s}|\bm{\Vec{X}_{t,c}}]).
\end{align*}
Therefore, the OLS estimator (\ref{OLSEstimator}) can be written as
\begin{align*}
    &\frac{\sum_{t^-=t-\delta+1}^{t}(t^--\frac{\sum_{t^-=t-\delta+1}^{t}t^-}{\delta})(\mathbb{E}[\ln(\bm{I_{t^-,c}})|{\Vec{X}_{t,c}}] - \frac{\sum_{t^-=t-\delta+1}^{t}\mathbb{E}[\ln(\bm{I_{t^-,c}})|{\Vec{X}_{t,c}}]}{\delta})}{\sum_{t^-=t-\delta+1}^{t}\left(t^--\frac{\sum_{t^-=t-\delta+1}^{t}t^-}{\delta}\right)^2}\\
    =& \sum^t_{t^-=t-\delta+2}\frac{ \sum^t_{i=t^-}(i-\frac{2t-\delta+1}{2})}{\sum_{i=t-\delta+1}^{t}\left(i-\frac{\sum_{i=t-\delta+1}^{t}i}{\delta}\right)^2}{r_{t^-,c}}\\
    =&  \sum^t_{t^-=t-\delta+2}\frac{ \sum^t_{i=t^-}(i-\frac{2t-\delta+1}{2})}{\sum_{i=t-\delta+1}^{t}i\left(i-\frac{2t-\delta+1}{2}\right)}{r_{t^-,c}}\\
    =&  \sum^t_{t^-=t-\delta+2}\frac{ \sum^t_{i=t^-}(i-\frac{2t-\delta+1}{2})}{\sum_{i=t-\delta+2}^{t}(i-(t-\delta+1))\left(i-\frac{2t-\delta+1}{2}\right)}{r_{t^-,c}}.
\end{align*}
As such, by defining $\{\hat{\mu}_{t^-,c}\}_{t^-\in \{t-\delta+2,\dots,t\}}$ and $\{{r_{t^-,c,c}}\}_{t^-\in \{t-\delta+2,\dots,t\}}$ according to (\ref{sampleweight}) and (\ref{OLSEQ2}), we have (\ref{OLSEQ}) and its sample counterpart (\ref{rtcOLST5}). Notably, the sample weights (\ref{sampleweight}) are fixed and independent of feature realizations.\par

% do not depend on features because OLS estimators with fixed window sizes are estimated using uniform sampling. That is, to estimate the instantaneous case growth rate of county $c$ on day $t$ using the OLS estimator with fixed window size $\delta$, the incident case number on each past $\delta$ day is sampled with probability $\frac{1}{\delta}$, i.e., $\forall t^-\in\{t-\delta+1,\dots,t\},\ \mathbb{P}[\bm{s}=t^-|\bm{\Vec{X}_{t,c}}]=\frac{1}{\delta}.$\par

Lastly, we show that $\{\hat{\mu}_{t^-,c}\}_{t^-\in \{t-\delta+2,\dots,t\}}$ are non-negative weights that sum to 1. First, $\sum^t_{t^-=t-\delta+2}\hat{\mu}_{t^-,c}=1$ because
\begin{align*}
    \sum^t_{t^-=t-\delta+2}\hat{\mu}_{t^-,c}&=\sum^t_{t^-=t-\delta+2}\frac{ \sum^t_{i=t^-}(i-\frac{2t-\delta+1}{2})}{\sum_{i=t-\delta+2}^{t}(i-(t-\delta+1))\left(i-\frac{2t-\delta+1}{2}\right)}\\
    &= \frac{ \sum^t_{i=t-\delta+2}\sum^{i}_{t^-=t-\delta+2}(i-\frac{2t-\delta+1}{2})}{\sum_{i=t-\delta+2}^{t}(i-(t-\delta+1))\left(i-\frac{2t-\delta+1}{2}\right)}\\
    &= \frac{ \sum^t_{i=t-\delta+2}(i-(t-\delta+1))(i-\frac{2t-\delta+1}{2})}{\sum_{i=t-\delta+2}^{t}(i-(t-\delta+1))\left(i-\frac{2t-\delta+1}{2}\right)}\\
    &=1.
\end{align*}
Second, to see that $\{\hat{\mu}_{t^-,c}\}_{t^-\in \{t-\delta+2,\dots,t\}}$ are non-negative, we first demonstrate that their numerators are non-negative for any realized features $x_{t,c}$.\par
Case 1: $\forall t^-\in \{t-\delta+2,\dots,t\}$ where $t^--1 \leq \frac{2t-\delta+1}{2}$, the numerator of $\hat{\mu}_{t^-,c}$ can be expressed as
\begin{align*}
    &\sum^{t}_{i=t^-}(i-\frac{2t-\delta+1}{2})\\
    =&\sum^{t}_{i=t-\delta+1} \left(i-\frac{2t-\delta+1}{2}\right)-\sum^{t^--1}_{i=t-\delta+1}\left(i-\frac{2t-\delta+1}{2}\right)\\
    =&\sum^{t^--1}_{i=t-\delta+1}\left(\frac{2t-\delta+1}{2}-i\right)\\
    \geq& 0,
\end{align*}
which is non-negative.\par
Case 2: $\forall t^-\in \{t-\delta+2,\dots,t\}$ where $t^--1 >  \frac{2t-\delta+1}{2}$, the numerator of $\hat{\mu}_{t^-,c}$ is positive, i.e.,
$$\sum^{t}_{i=t^-}(i-\frac{2t-\delta+1}{2})>0.$$
%because $\mathbb{P}[\bm{s}=i](i-\mathbb{E}[\bm{s}])>0,\ \ \forall i\in \{t-\delta+1,\dots,t\}.$\par
By combining Case 1 and Case 2, we thus prove that the numerator of each $\hat{\mu}_{t^-,c}$ is non-negative for any $t^-\in \{t-\delta+2,\dots,t\}$.
In addition, since $\sum^t_{t^-=t-\delta+2}\hat{\mu}_{t^-,c}=1$, the denominator of each $\hat{\mu}_{t^-,c}$, as a sum of non-negative numerators, is also non-negative. Hence $\{\hat{\mu}_{t^-,c}\}_{t^-\in \{t-\delta+2,\dots,t\}}$ are non-negative. \QED

Lastly, we state and prove a useful lemma that clarifies the distribution of OLS weights (\ref{sampleweight}).
\begin{lemma}\label{OLSWeightLemma}
    The OLS weights $\{\hat{\mu}_{t^-,c}\}_{t^-\in \{t-\delta+2,\dots,t\}}$, where 
    $$ \hat{\mu}_{t^-,c}=  \frac{\sum^t_{i=t^-}(i-\frac{2t-\delta+1}{2})}{\sum^t_{i=t-\delta + 1}(i-(t-\delta+1))(i-\frac{2t-\delta+1}{2})},$$
    are concentrated towards the center, i.e.,\\ $\hat{\mu}_{t,c}=1$ when $\delta=2$, $\hat{\mu}_{t,c}=\hat{\mu}_{t-1,c}$ when $\delta=3$, and
$$\hat{\mu}_{t,c}<\dots<\hat{\mu}_{\lfloor \frac{2t-\delta+2}{2}\rfloor,c}=\hat{\mu}_{\lceil \frac{2t-\delta+2}{2}\rceil,c}>\dots>\hat{\mu}_{t-\delta+2,c}$$
when $\delta>3$.
\end{lemma}
Proof of Lemma \ref{OLSWeightLemma}:\\
The first two claims, i.e., $\hat{\mu}_{t,c}=1$ when $\delta=2$, $\hat{\mu}_{t,c}=\hat{\mu}_{t-1,c}$ when $\delta=3$, can be directly verified.\\
To see that the third claim is true, we first verify $\hat{\mu}_{\lfloor \frac{2t-\delta+2}{2}\rfloor,c}=\hat{\mu}_{\lceil \frac{2t-\delta+2}{2}\rceil,c}$. When $\delta$ is an even number, we have $\lfloor \frac{2t-\delta+2}{2}\rfloor=\lceil \frac{2t-\delta+2}{2}\rceil$, which directly implies $\hat{\mu}_{\lfloor \frac{2t-\delta+2}{2}\rfloor,c}=\hat{\mu}_{\lceil \frac{2t-\delta+2}{2}\rceil,c}$. When $\delta$ is an odd number, we have
\begin{align*}
    \hat{\mu}_{\lceil \frac{2t-\delta+2}{2}\rceil,c}=&\frac{\sum^t_{i=\lceil \frac{2t-\delta+2}{2}\rceil}(i-\frac{2t-\delta+1}{2})}{\sum^t_{i=t-\delta + 1}(i-(t-\delta+1))(i-\frac{2t-\delta+1}{2})}\\
    =&\frac{\sum^t_{i=\lfloor \frac{2t-\delta+2}{2}\rfloor}(i-\frac{2t-\delta+1}{2})-(\frac{2t-\delta+1}{2}-\frac{2t-\delta+1}{2})}{\sum^t_{i=t-\delta + 1}(i-(t-\delta+1))(i-\frac{2t-\delta+1}{2})}\\
    =&\frac{\sum^t_{i=\lfloor \frac{2t-\delta+2}{2}\rfloor}(i-\frac{2t-\delta+1}{2})}{\sum^t_{i=t-\delta + 1}(i-(t-\delta+1))(i-\frac{2t-\delta+1}{2})}\\
    =&\hat{\mu}_{\lfloor \frac{2t-\delta+2}{2}\rfloor,c}.
\end{align*}
Additionally, for all $\eta\in \mathbb{Z}^+$ such that $t-\delta+2< \lfloor \frac{2t-\delta+2}{2}\rfloor-\eta<\lceil \frac{2t-\delta+2}{2}\rceil+\eta< t$, we have $\hat{\mu}_{\lceil \frac{2t-\delta+2}{2}\rceil+\eta,c}>\hat{\mu}_{\lceil \frac{2t-\delta+2}{2}\rceil+\eta+1,c}$, i.e.,
\begin{align*}
    \hat{\mu}_{\lceil \frac{2t-\delta+2}{2}\rceil+\eta+1,c}=&\frac{\sum^t_{i=\lceil \frac{2t-\delta+2}{2}\rceil+\eta+1}(i-\frac{2t-\delta+1}{2})}{\sum^t_{i=t-\delta + 1}(i-(t-\delta+1))(i-\frac{2t-\delta+1}{2})}\\
    =&\frac{\sum^t_{i=\lceil \frac{2t-\delta+2}{2}\rceil+\eta}(i-\frac{2t-\delta+1}{2})-(\lceil \frac{2t-\delta+2}{2}\rceil+\eta+1-\frac{2t-\delta+1}{2})}{\sum^t_{i=t-\delta + 1}(i-(t-\delta+1))(i-\frac{2t-\delta+1}{2})}\\
    <&\frac{\sum^t_{i=\lceil \frac{2t-\delta+2}{2}\rceil+\eta}(i-\frac{2t-\delta+1}{2})}{\sum^t_{i=t-\delta + 1}(i-(t-\delta+1))(i-\frac{2t-\delta+1}{2})}\\
    =&\hat{\mu}_{\lceil \frac{2t-\delta+2}{2}\rceil+\eta,c},
\end{align*}
and $\hat{\mu}_{\lfloor \frac{2t-\delta+2}{2}\rfloor-\eta,c}>\hat{\mu}_{\lfloor \frac{2t-\delta+2}{2}\rfloor-\eta-1,c}$, i.e.,
\begin{align*}
    \hat{\mu}_{\lfloor \frac{2t-\delta+2}{2}\rfloor-\eta-1,c}=&\frac{\sum^t_{i=\lfloor \frac{2t-\delta+2}{2}\rfloor-\eta-1}(i-\frac{2t-\delta+1}{2})}{\sum^t_{i=t-\delta + 1}(i-(t-\delta+1))(i-\frac{2t-\delta+1}{2})}\\
    =&\frac{\sum^t_{i=\lfloor \frac{2t-\delta+2}{2}\rfloor-\eta}(i-\frac{2t-\delta+1}{2})+(\lfloor \frac{2t-\delta+2}{2}\rfloor-\eta-1-\frac{2t-\delta+1}{2})}{\sum^t_{i=t-\delta + 1}(i-(t-\delta+1))(i-\frac{2t-\delta+1}{2})}\\
    <&\frac{\sum^t_{i=\lfloor \frac{2t-\delta+2}{2}\rfloor-\eta}(i-\frac{2t-\delta+1}{2})}{\sum^t_{i=t-\delta + 1}(i-(t-\delta+1))(i-\frac{2t-\delta+1}{2})}\\
    =&\hat{\mu}_{\lfloor \frac{2t-\delta+2}{2}\rfloor-\eta,c}.
\end{align*}
As such, we have shown that for $\delta>3$,
$$\hat{\mu}_{t,c}<\dots<\hat{\mu}_{\lfloor \frac{2t-\delta+2}{2}\rfloor,c}=\hat{\mu}_{\lceil \frac{2t-\delta+2}{2}\rceil,c}>\dots>\hat{\mu}_{t-\delta+2,c}.$$
\QED

\subsection{Analytical Results of the Bias-variance Trade-off for Instantaneous Growth Rate Estimations}\label{AppendixBiasVariance}

This section examines how the choice of fitting window size affects the bias-variance trade-off in estimating instantaneous growth rates.\par

First, we obtain the standard bias-variance decomposition of an estimator's MSE:
\begin{theorem}\label{BVTradeoff}
    For any estimator $\bm{\hat{r}_{t,c}}$ of the county-level instantaneous exponential growth rate $r(\Vec{X}_{t,c})$, its Mean-squared error (MSE) has the following bias-variance decomposition: 
    \begin{equation}\label{MSEGrowthRate}
        \mathbb{E}[\left(r(\Vec{X}_{t,c})-\bm{\hat{r}_{t,c}}\right)^2]=(\underbrace{r(\Vec{X}_{t,c})-\mathbb{E}[\bm{\hat{r}_{t,c}}]}_{\textbf{Bias}})^2
    +\underbrace{\mathbb{Var}[\bm{\hat{r}_{t,c}} ]}_{\textbf{Variance}}.
    \end{equation}
\end{theorem}

Second, We provide an analytical result for how the bias-variance trade-off of OLS estimators is affected by their fitting window sizes:
\begin{corollary}\label{OLSTradeoff}
    For an OLS estimator with $\delta$-day fitting window ($\bm{\hat{r}^{ols(\delta)}_{t,c}}$), its MSE, i.e.,
    \begin{align*}
    &\mathbb{E}[\left(r(\Vec{X}_{t,c})-\bm{\hat{r}^{ols(\delta)}_{t,c}}\right)^2 ]\\
    =&\left(\underbrace{r(\Vec{X}_{t,c})-\mathbb{E}[\sum^t_{t^-=t-\delta+2}\hat{\mu}_{t^-,c}\left(\ln(\bm{I_{t^-,c}})-\ln(\bm{I_{t^--1,c}}) \right) |\Vec{X}_{t,c}]}_{\textbf{Bias}}\right)^2\\
    +&\underbrace{\mathbb{Var}[\sum^t_{t^-=t-\delta+2}\hat{\mu}_{t^-,c}\left(\ln(\bm{I_{t^-,c}})-\ln(\bm{I_{t^--1,c}}) \right)|\Vec{X}_{t,c} ]}_{\textbf{Variance}}, \numberthis \label{R2MSE}
\end{align*}
has a zero bias term when $\delta=2$ and has an asymptotically vanishing variance term when $\delta=t$.
\end{corollary}
Notably, changing the fitting window size $\delta$ cannot strike the right balance for the bias-variance tradeoff in (\ref{R2MSE}) due to the non-adaptive weighting scheme $\{{\hat{\mu}_{t^-,c}}\}_{t^-\in \{t-\delta+2,\dots,t\}}$.\par

In contrast, through adaptive weighting, the \texttt{TLRF} estimator $\bm{\hat{r}^{TLRF}_{t,c}}$ can more effectively balance the bias and variance tradeoff in its MSE:
\begin{align*}
    &\mathbb{E}[\left(r(\Vec{X}_{t,c})-\bm{\hat{r}^{TLRF}_{t,c}}\right)^2 ]\\
    =&\left(\underbrace{r(\Vec{X}_{t,c})-\mathbb{E}[\sum\limits_{t'\in [t]}\sum\limits_{c'\in C}  \bm{\gamma_{t',c'}}(\Vec{X}_{t,c}) \left(\ln(\bm{I_{t',c'}})-\ln(\bm{I_{t'-1,c'}})\right)|\Vec{X}_{t,c}]}_{\textbf{Bias}}\right)^2\\
    +&\underbrace{\mathbb{Var}[\sum\limits_{t'\in [t]}\sum\limits_{c'\in C}  \bm{\gamma_{t',c'}}(\Vec{X}_{t,c}) \left(\ln(\bm{I_{t',c'}})-\ln(\bm{I_{t'-1,c'}})\right)|\Vec{X}_{t,c}]}_{\textbf{Variance}}.\numberthis \label{R2TLGRFMSE}
\end{align*}
That is, by selecting $\{\bm{\gamma_{t',c'}}(\Vec{X}_{t,c})\}_{t'\in [t],c'\in C}$ to minimize (\ref{R2TLGRFMSE}), one can achieve better or at least equal MSE compared to selecting $\delta$ to minimize (\ref{R2MSE}). More precisely, (\ref{R2TLGRFMSE}) would yield the same value as (\ref{R2MSE}) if we choose the weights $\bm{\gamma_{t',c'}}(X_{t,c})=\hat{\mu}_{t',c}\ \ \forall t'\in \{t-\delta+2,\dots,t\}$ and  $\bm{\gamma_{t',c'}}(X_{t,c})=0\ \ \forall t'\in \{2,\dots,t-\delta+1\}\ \text{or}\ \forall c'\in C\backslash\{c\}$. The key insight here is that \texttt{TLRF} offers a generalized approach to solving the fitting window selection problem by transforming it into a weight assignment problem.\par

Third, we discuss how \texttt{TLRF} manages its bias-variance trade-off. Notably, balancing the bias-variance trade-off for \texttt{TLRF} essentially involves choosing $\bm{\hat{r}^{TLRF}_{t,c}}$ to minimize (\ref{R2TLGRFMSE}). However, since the ground truth, i.e., growth rate $r(\Vec{X}_{t,c})$, is not observable, optimizing (\ref{R2TLGRFMSE}) empirically requires certain transformations. For example, choosing $\bm{\hat{r}^{TLRF}_{t,c}}$ to minimize
(\ref{R2TLGRFMSE}) is equivalent to choosing $\bm{\hat{r}^{TLRF}_{t,c}}$ to maximize
\begin{equation}\label{TransformMax}
    r^2(\Vec{X}_{t,c})-\mathbb{E}[\left(r(\bm{\Vec{X}})-\bm{\hat{r}^{TLRF}_{t,c}}\right)^2].
\end{equation}
\cite{athey2016recursive} observed that if $\mathbb{E}[\bm{\hat{r}^{TLRF}_{t,c}}]=r(\Vec{X}_{t,c})$, then the sample counterpart of (\ref{TransformMax}) reduces to $(\bm{\hat{r}^{TLRF}_{t,c}})^2$, which is directly measurable and thus can be optimized empirically. Subsequently, \cite{athey2019generalized} further relaxed this assumption and demonstrated that their GRF method achieves consistency if the parameter of interest is identifiable by a conditional mean outcome. Therefore, to balance the bias-variance trade-off of \texttt{TLRF}, we first find a conditional mean outcome that identifies $r(\Vec{X}_{t,c})$ in \S\ref{4.1} and subsequently apply the random forest algorithm to obtain a consistent final estimator in \S\ref{4.2}.\\

% \begin{corollary}
%     The bias term of \texttt{TLRF}'s MAE is zero
% \end{corollary}

%Use (\ref{GRFEst})

\noindent
\textbf{Proofs for this section:}\\

\noindent
Proof of Theorem \ref{BVTradeoff}:
\begin{align*}
&\mathbb{E}[\left(r(\Vec{X}_{t,c})-\bm{\hat{r}_{t,c}}\right)^2  ]\\
    &= r^2(x_{t,c})-2r(x_{t,c})\mathbb{E}[\bm{\hat{r}_{t,c}}  ]+\mathbb{E}[\bm{\hat{r}^2_{t,c}}  ]\\
    % \mathbb{E}[(r(\bm{\Vec{X}_{t,c}})-\mathbb{E}[\underset{c'\in C}{\sum}\sum^t_{t'=2}w_{t',c'} \left(\ln(\bm{I_{t',c'}})-\ln(\bm{I_{t'-1,c'}})\right)|\bm{\Vec{X}_{t,c}}=x_{t,c}])^2|\bm{\Vec{X}_{t,c}}=x_{t,c}]\\
    % +&\mathbb{E}[(\underset{c'\in C}{\sum}\sum^t_{t'=2}w_{t',c'} \left(\ln(\bm{I_{t',c'}})-\ln(\bm{I_{t'-1,c'}})\right))^2|\bm{\Vec{X}_{t,c}}=x_{t,c}]\\
    &= r^2(x_{t,c})-2r(x_{t,c})\mathbb{E}[\bm{\hat{r}_{t,c}}  ] +\mathbb{E}[\bm{\hat{r}^2_{t,c}}  ] + (\mathbb{E}[\bm{\hat{r}_{t,c}}  ])^{2} - (\mathbb{E}[\bm{\hat{r}_{t,c}}  ])^{2}\\
    &= r^2(x_{t,c})-2r(x_{t,c})\mathbb{E}[\bm{\hat{r}_{t,c}}  ] + (\mathbb{E}[\bm{\hat{r}_{t,c}}  ])^{2}+\mathbb{E}[\bm{\hat{r}^2_{t,c}}  ] - (\mathbb{E}[\bm{\hat{r}_{t,c}}  ])^{2}\\
    &=(\underbrace{r(x_{t,c})-\mathbb{E}[\bm{\hat{r}_{t,c}}  ]}_{\textbf{Bias}})^2
    +\underbrace{\mathbb{Var}[\bm{\hat{r}_{t,c}}  ]}_{\textbf{Variance}}.
\end{align*}
\QED

\noindent
Proof of Corollary \ref{OLSTradeoff}:\\
By Theorem \ref{Restrict}, an OLS estimator with $\delta$-day fitting window can be expressed as:
$$\bm{\hat{r}^{ols(\delta)}_{t,c}}=\sum^t_{t^-=t-\delta+2}\hat{\mu}_{t^-,c}\left(\ln(\bm{I_{t^-,c}})-\ln(\bm{I_{t^--1,c}}) \right),$$
where
\begin{equation*}
    \hat{\mu}_{t^-,c}=  \frac{\sum^t_{i=t^-}(i-\frac{2t-\delta+1}{2})}{\sum^t_{i=t-\delta + 1}(i-(t-\delta+1))(i-\frac{2t-\delta+1}{2})}.
\end{equation*}

First, when $\delta=2$, the bias term of its MSE (\ref{MSEGrowthRate}), i.e.,
$$r(\Vec{X}_{t,c})-\mathbb{E}[\ln(\bm{I_{t,c}})-\ln(\bm{I_{t-1,c}})|\Vec{X}_{t,c}],$$
is zero by Theorem \ref{T1}.

Second, when $\delta=t$, we show that the variance term of its MSE (\ref{MSEGrowthRate}) vanishes asymptotically, i.e.,
\begin{equation}\label{VarianceVanish}
    lim_{t\rightarrow \infty}\mathbb{Var}[\bm{\hat{r}^{ols(\delta)}_{t,c}}]=0.
\end{equation}
To see this, We first note that the variance term can be upper-bounded in the following fashion:
\begin{align*}
&\mathbb{Var}[\bm{\hat{r}^{ols(\delta)}_{t,c}}]\\
    =&\mathbb{Var}[\sum^t_{t^-=2}\hat{\mu}_{t^{-},c} \left(\ln(\bm{I_{t^-,c}})-\ln(\bm{I_{t^--1,c}})\right)|\Vec{X}_{t,c}] \\
    =& \sum^t_{t'=2}\sum^t_{t''=2}\hat{\mu}_{t',c}\hat{\mu}_{t'',c} \mathbb{Cov}[\ln(\bm{I_{t',c}})-\ln(\bm{I_{t'-1,c}}),\ln(\bm{I_{t'',c}})-\ln(\bm{I_{t''-1,c}})|  \Vec{X}_{t,c}] \\
    \leq&\sum^t_{t'=2}\sum^t_{t''=2}(\hat{\mu}_{\lfloor\frac{t+2}{2}\rfloor,c})^2 \bigg|\mathbb{Cov}[\ln(\bm{I_{t',c}})-\ln(\bm{I_{t'-1,c}}),\ln(\bm{I_{t'',c}})-\ln(\bm{I_{t''-1,c}})|  \Vec{X}_{t,c}] \bigg| \numberthis \label{variance_3_ols_weight_centered}\\
    =&(\hat{\mu}_{\lfloor\frac{t+2}{2}\rfloor,c})^{2} \sum^t_{t'=2}\sum^t_{t''=2}\bigg|\mathbb{Cov}[\ln(\bm{I_{t',c}})-\ln(\bm{I_{t'-1,c}}),\ln(\bm{I_{t'',c}})-\ln(\bm{I_{t''-1,c}})|  \Vec{X}_{t,c}]\bigg|,  \numberthis \label{variance_4_bounded}
\end{align*}
where
(\ref{variance_3_ols_weight_centered}) follows from Lemma \ref{OLSWeightLemma}, i.e.,
$$\hat{\mu}_{t,c}<\dots<\hat{\mu}_{\lfloor \frac{t+2}{2}\rfloor,c}=\hat{\mu}_{\lceil \frac{t+2}{2}\rceil,c}>\dots>\hat{\mu}_{t-\delta+2,c}.$$
Therefore, to show (\ref{VarianceVanish}), it suffices to show (\ref{variance_4_bounded}) approaches zero as $t$ approaches infinity.\par
In addition, as a sufficient condition for ergodicity of second moments, it is common to assume that the autocovariance is absolutely summable, i.e.,
$$lim_{t\rightarrow \infty}\sum^t_{t'=2}\sum^t_{t''=2}\bigg|\mathbb{Cov}[\ln(\bm{I_{t',c}})-\ln(\bm{I_{t'-1,c}}),\ln(\bm{I_{t'',c}})-\ln(\bm{I_{t''-1,c}})|  \Vec{X}_{t,c}]\bigg|<\infty,$$
when proving limit theorems for serially dependent observations \citep{hamilton2020time}.\par
As such, to show (\ref{variance_4_bounded}) approaches zero as $t$ approaches infinity, it suffices to show that
\begin{equation}\label{OLSlimit}
    lim_{t\rightarrow \infty}(\hat{\mu}_{\lfloor\frac{t+2}{2}\rfloor,c})^{2}=0.
\end{equation}
In fact, by fixing the fitting window size $\delta=t$, we have
\begin{align*}
    \hat{\mu}_{\lfloor\frac{t+2}{2}\rfloor,c} &=  \frac{\sum^t_{i=\lfloor\frac{t+2}{2}\rfloor}(i-\frac{2t-t+1}{2})}{\sum^t_{i=t-t + 1}(i-(t-t+1))(i-\frac{2t-t+1}{2})}\\
    &= \frac{\sum^t_{i=\lfloor\frac{t+2}{2}\rfloor}(i-\frac{t+1}{2})}{\sum^t_{i=1}(i-1)(i-\frac{t+1}{2})}\\
    &=\frac{\frac{1}{2} \left(t-\left\lfloor \frac{t}{2}\right\rfloor \right) \left\lfloor \frac{t}{2}\right\rfloor}{\frac{1}{12} t \left(t^2-1\right)}\\
    % &= \frac{\sum^t_{i= t^{-} - 1}(i-\frac{t-1}{2})}{\sum^t_{i=0}i(i-\frac{t-1}{2})}\\
    % &= \frac{\sum^t_{i= \lfloor \frac{t + 2}{2} \rfloor - 1}(i-\frac{t-1}{2})}{\sum^t_{i=0}i(i-\frac{t-1}{2})}\\
    % &=  \frac{\sum^t_{i= \lfloor \frac{t + 2}{2} \rfloor - 1}(i-\frac{t-1}{2})}{\frac{1}{12}(t)(t-1)(t+1)}\\
% &= \begin{cases}
%     \frac{\sum^{t-1}_{i= \frac{t}{2}}\big(i-\frac{t-1}{2}\big)}{\frac{1}{12}(t)(t-1)(t+1)} & \text{, when $t$ even}\\
%     \frac{ \sum^{t-1}_{i= \frac{t-1}{2}}\big(i-\frac{t-1}{2}\big)}{\frac{1}{12}(t)(t-1)(t+1)} & \text{, when $t$ odd}
% \end{cases}\\
&= \begin{cases}
     \frac{ \frac{1}{8}t^{2}}{\frac{1}{12}(t)(t-1)(t+1)} & \text{, when $t$ even}\\
    \frac{ \frac{1}{8}(t-1)(t+1)}{\frac{1}{12}(t)(t-1)(t+1)} & \text{, when $t$ odd}\\
\end{cases}\\
&= \begin{cases}
        \frac{3}{2}(\frac{t}{t^{2}-1})  & \text{, when $t$ even}\\
        \frac{3}{2}(\frac{1}{t})  & \text{, when $t$ odd},
    \end{cases}
\end{align*}
which implies (\ref{OLSlimit}). \QED

\section{Synthetic Experiment}\label{adp:synthetic}
We propose a synthetic experiment as a sensitivity analysis for how a small model misspecification could affect the final results.

Put simply, it assumes that the outcome variable and dependent variable follow a locally (dependent on $\vec{X}_{t,c}$) log-linear relation.

For our experiment, we consider a scenario where the true growth rate is locally linear instead:
\begin{align}\label{TrueDGP}
    \mathbb{E}[\bm{I^t_{s,c}}|\Vec{X}_{t,c}] =\alpha(\Vec{X}_{t,c})+ r(\Vec{X}_{t,c}){s}+\mathbb{E}[\bm{\varepsilon^t_{t,c}}|\Vec{X}_{t,c}] \ \ \forall s\in \mathbb{Z}^+.
\end{align}

\textbf{Experiment Setup:}\\
Following our real life use case, we index the data observations by the tuple $(t,c)$, where $t \in [T]:=\{1, \cdots, 365\}$ is the progress of time, and $c \in C:=\{1,\cdots,1000\}$ are indexes for 1000 different counties.

At each time county pair index, the observed outcome variable is $I_{t,c} \in \mathbb{Z}^{+}_{0}$ and the observable feature variable is $\vec{X}_{t,c} \in \mathbb{R}^{6}$ i.e., a vector with 6 continuous features.

Let the true data generating process (DGP) for the incident case numbers be specified below:
\begin{align}
    I_{1,c} &= 100, \forall c\in C\\
    I_{t,c} &= \underbrace{10\bigg(\sum_{k=1}^{2}X_{t,c,k}\bigg)}_{r(\vec{X}_{t,c})} + I_{t-1,c}, \forall  t \geq 1, c\in C
\end{align}

In short, the true growth rate parameter is the determined by the sum of the first 2 features, with the other 4 features being irrelevant.
We simulate the features and random noise as follows:
\begin{align}
    X_{t,c,k} &\sim \text{Uniform}(0,1), \forall t\in [T], c\in C, k\in\{1,\dotsc,6\}\\
\end{align}

If our model is correctly specified as linear, then the projected forecast 7 days later from today is: 
\begin{align*}
    (I_{t+7,c}) &= (I_{t,c}) + 7\hat{r}_{t,c}\\
\end{align*}
If not correctly specified as exponential:
\begin{align*}
    (I_{t+7,c}) &= (I_{t,c})\exp\{7\hat{r}_{t,c}\}\\
    \ln(I_{t+7,c}) &= \ln(I_{t,c}) + 7\hat{r}_{t,c}
\end{align*}

To allow both models to have enough time to accumulate data for training, we start benchmarking them daily from $t=30$ onwards. 

\textbf{Results:}\\
Figures \ref{fig:R2C1_Parameter_MAE} and \ref{fig:R2C1_Parameter_RMSE} present the daily MAE and RMSE of \texttt{Exponential\ TLRF} and \texttt{Linear\ TLRF} estimators, respectively. The median MAE and RMSE of these estimators are presented in Table \ref{tab:daily_parameter_mean_mae_rmse}. Compared to \texttt{Linear\ TLRF}, the misspecified exponential model estimator \texttt{Exponential\ TLRF} is consistently outperformed as expected.

\begin{figure}[H]
\centering
\includegraphics[width=\textwidth]{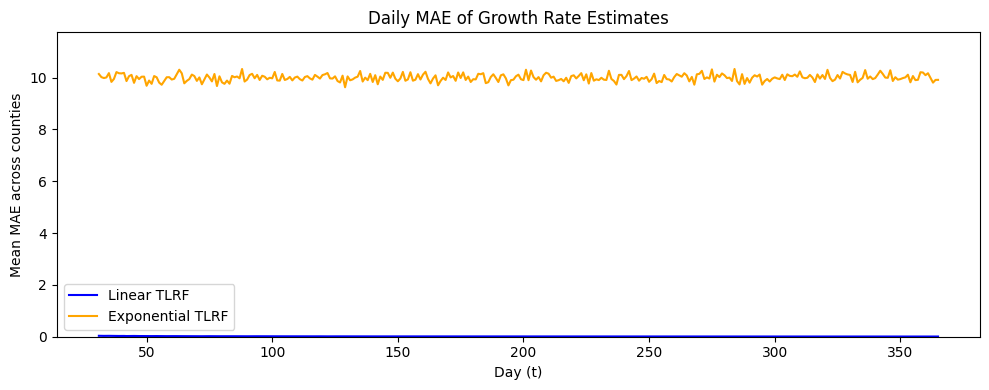}
\caption{MAE plot of parameter estimation accuracy (\texttt{Exponential\ TLRF} vs. \texttt{Linear\ TLRF}): Since synthetic data is generated from the locally linear model (\ref{TrueDGP}), the \texttt{Exponential\ TLRF} estimator suffers from the misspecification bias, while the \texttt{Linear\ TLRF} estimator is unbiased.}
\label{fig:R2C1_Parameter_MAE}
\end{figure}

\begin{figure}[H]
\centering
\includegraphics[width=\textwidth]{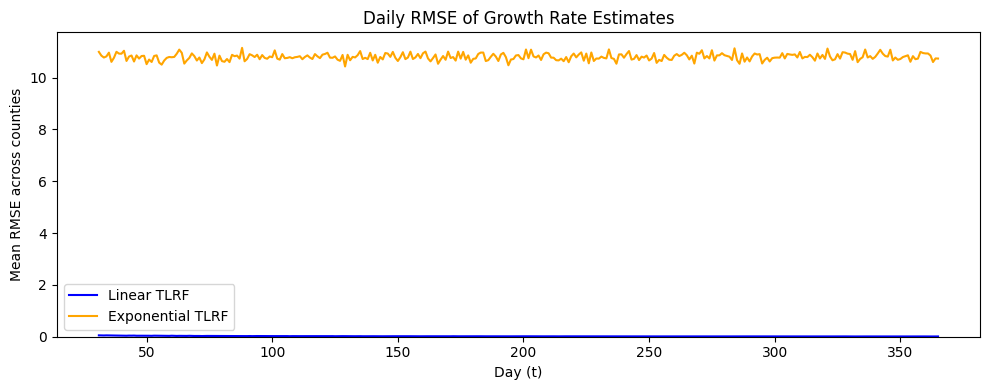}
\caption{RMSE plot of parameter estimation accuracy (\texttt{Exponential\ TLRF} vs. \texttt{Linear\ TLRF}): Since synthetic data is generated from the locally linear model (\ref{TrueDGP}), the \texttt{Exponential\ TLRF} estimator suffers from the misspecification bias, while the \texttt{Linear\ TLRF} estimator is unbiased.}
\label{fig:R2C1_Parameter_RMSE}
\end{figure}

In addition we present the overall median of the daily MAE and RMSE of the growth rate estimates of both models in Table \ref{tab:daily_parameter_mean_mae_rmse}.

\begin{table}[!htpb]
\centering
\begin{tabular}{|l|c|c|}
\hline
\textbf{Method} & \textbf{MAE} & \textbf{RMSE} \\
\hline
\texttt{Exponential\ TLRF} & 10.00 & 10.80 \\
\texttt{Linear\ TLRF} & 0.013 & 0.018\\
%Linear Model & 9.99 & 10.01 \\
\hline
\end{tabular}
\caption{Median MAE and RMSE of parameter estimation accuracy (\texttt{Exponential\ TLRF} vs. \texttt{Linear\ TLRF}): Since synthetic data is generated from the locally linear model (\ref{TrueDGP}), the \texttt{Exponential\ TLRF} estimator suffers from the misspecification bias, while the \texttt{Linear\ TLRF} estimator is unbiased.}
\label{tab:daily_parameter_mean_mae_rmse}
\end{table}

Subsequently, we compare the 7-day-ahead prediction accuracy between \texttt{Exponential\ TLRF} and \texttt{Linear\ TLRF}, a benchmarking methodology applied throughout the manuscript. Figures \ref{fig:R2C1_MAE} and \ref{fig:R2C1_RMSE} present the daily MAE and RMSE of these two forecasting models, respectively. The median MAE and RMSE of both models' forecasts are presented in Table \ref{tab:daily_mean_mae_rmse}. Interestingly, despite parameter estimation errors, forecast accuracy between \texttt{Exponential\ TLRF} and \texttt{Linear\ TLRF}  remains comparable, with the gap narrowing as training data accumulates.

\begin{figure}[H]
\centering
\includegraphics[width=\textwidth]{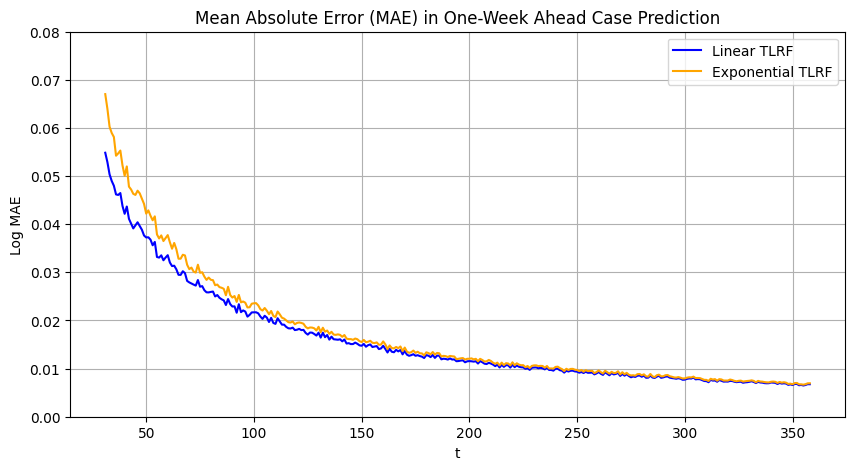}
\caption{MAE plot of prediction accuracy (\texttt{Exponential\ TLRF} vs. \texttt{Linear\ TLRF}): Since synthetic data is generated from the locally linear model (\ref{TrueDGP}), the \texttt{Exponential\ TLRF} estimator suffers from the misspecification bias, while the \texttt{Linear\ TLRF} estimator is unbiased.}
\label{fig:R2C1_MAE}
\end{figure}

\begin{figure}[H]
\centering
\includegraphics[width=\textwidth]{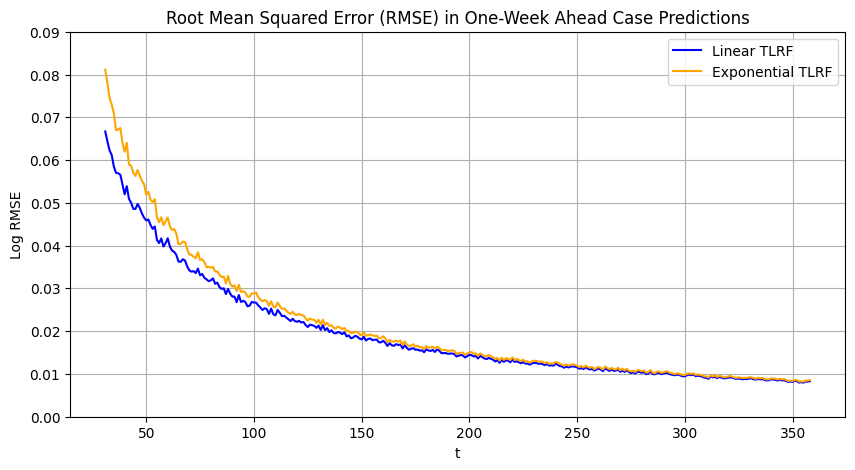}
\caption{RMSE plot of prediction accuracy (\texttt{Exponential\ TLRF} vs. \texttt{Linear\ TLRF}): Since synthetic data is generated from the locally linear model (\ref{TrueDGP}), the \texttt{Exponential\ TLRF} estimator suffers from the misspecification bias, while the \texttt{Linear\ TLRF} estimator is unbiased.}
\label{fig:R2C1_RMSE}
\end{figure}

\begin{table}[!htpb]
\centering
\begin{tabular}{|l|c|c|}
\hline
\textbf{Method} & \textbf{MAE} & \textbf{RMSE} \\
\hline
\texttt{Exponential\ TLRF} & 0.012 & 0.015 \\
\texttt{Linear\ TLRF} & 0.012 & 0.014 \\
%Mis-specified Exponential Growth Rate Model & 0.114 & 0.139 \\
%Linear Model & 0.115 & 0.142 \\
\hline
\end{tabular}
\caption{Median MAE and RMSE of prediction accuracy (\texttt{Exponential\ TLRF} vs. \texttt{Linear\ TLRF}): Since synthetic data is generated from the locally linear model (\ref{TrueDGP}), the \texttt{Exponential\ TLRF} estimator suffers from the misspecification bias, while the \texttt{Linear\ TLRF} estimator is unbiased.}
\label{tab:daily_mean_mae_rmse}
\end{table}

\textit{Experiment findings.} 

This synthetic experiment reveals two important insights about \texttt{TLRF}'s behavior under model misspecification. First, \texttt{TLRF} exhibits a large error in parameter estimation under model misspecification, as expected. Second, surprisingly, this parameter bias translates to only modest degradation in forecast accuracy with the performance gap narrowing as training data accumulates. This demonstrates that while Assumption 1 is crucial for the correct interpretation of parameters as exponential growth rates (fundamental for epidemiological models like SIR/SEIR), \texttt{TLRF} maintains practical robustness for prediction tasks even under misspecification.\par

The code for this benchmark can be found on our GitHub repository at: \url{https://github.com/wangzilongri/COVID-tracking/tree/TLGRF_major_revision/analysis/A1_Robustness_Check}.

%\section{Implementation Details and Pseudocode for Estimation Algorithms}
\section{Supplementary Benchmark Implementation Details and Results}\label{apd1}
This section is dedicated to providing additional implementation details and results of the benchmarks discussed in the main text.
Appendix \ref{apd1.dynamic} discusses the window fitting based estimators discussed in \S\ref{PE}.
Appendix \ref{apd:LASSOTL} then provides further implementation details regarding the \texttt{LASSOTL} benchmark discussed in \S\ref{benchmark_TL}.

\subsection{Window Based Methods}\label{apd1.dynamic}
In this subsection, we provide the implementation details and additional performance metrics of the  Fixed Windows estimators as referenced in \S\ref{subsec.fixed}, \texttt{tcv} and \texttt{ctcv} as referenced in \S\ref{subsec.flexible}, and the K-Means Dynamic Window procedure as described in \S\ref{kmeans}.

\subsubsection{Fixed Window Size Estimators}\label{supplementary_fixed}
This subsubsection contains the additional RSME plot (Figure \ref{fig:RMSE4_together}) of \texttt{TLRF} against fixed window size estimators mentioned in \S\ref{subsec.fixed}. We can see from Figure \ref{fig:RMSE4_together} that no clear winner emerges between various fixed window size choices and that they are also uniformly outperformed by \texttt{TLRF} when evaluated upon the RMSE metric.

\begin{figure}[H]
  \includegraphics[width=\linewidth, height=0.6\textwidth]{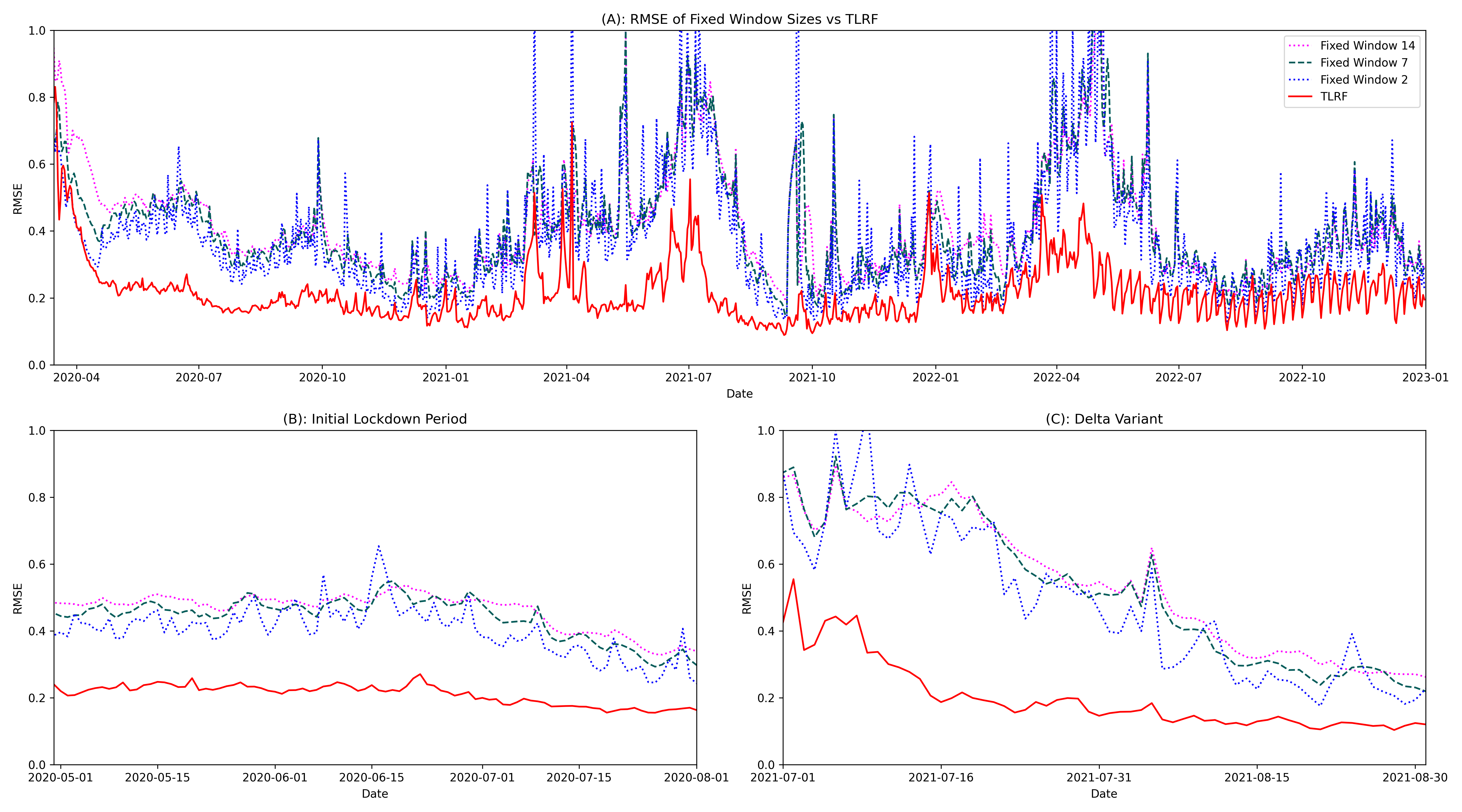}
  \caption{RMSE plot of prediction accuracy (\texttt{TLRF} vs. Fixed 2-, 7-, and 14-Day Fitting Window): Plot A (Top) shows the RMSE comparison across the study period. Plot B (bottom left) zooms in on the initial lockdown period. Plot C (bottom right) zooms in on the period during the outbreak of the Delta variant.}
  \label{fig:RMSE4_together}
\end{figure}

\subsubsection{tcv}\label{apd1.tcv}
The \texttt{tcv} estimator (\ref{tcv}) can be directly computed from the case incidence data $\{(s, \ln({I}_{s,c})) : \forall s \in \{t-\delta_{t}+1,\dotsc,t\}, c \in C\}$ for any fitting window size hyperparameter  as:
\begin{equation}\label{tcv_regression_form}
    \bm{\hat{r}^{ols(\delta_{t})}_{t,c}}=\frac{\sum_{s=t-\delta_{t}+1}^{t}(s-\frac{\sum_{s=t-\delta_{t}+1}^{t}s}{\delta_{t}})(\ln(\bm{I_{s,c}}) - \frac{\sum_{s=t-\delta_{t}+1}^{t}\ln(\bm{I_{s,c}})}{\delta_{t}})}{\sum_{s=t-\delta_{t}+1}^{t}\left(s-\frac{\sum_{s=t-\delta_{t}+1}^{t}s}{\delta_{t}}\right)^2}.
\end{equation} 
This encompasses the incidence case data of every county $c$ from the beginning of the chosen fitting window ($t-\delta_{t}+1$) to the current time of query ($t$).
As described in \S\ref{subsec.flexible}, the key characteristic of the \texttt{tcv} estimator is that at each time point $t$, all the counties are constrained to use the same fitting window size hyperparameter $\delta_{t}$, i.e., the entire nation shares the same fitting window hyperparameter at every time step.
We will now describe the cross-validation procedure to obtain the best $\delta_{t}$, and how the final test-time estimator is computed and evaluated.

Our goal at test time is to forecast the log-incident case numbers of each county ($\ln(I_{t+7,c})$) seven days ahead from our current time of query $t$.
Our search range for our hyperparameter ranges from $\{2,\dotsc,14\}$.
To prevent data leakage from the future, we follow the standard implementation of the rolling window time series cross validation procedure \citep[c.f.][]{scikit-learn}. 
Specifically, the training-validation splits on each day are generated successively to ensure that the training set is always of an earlier date than that of the validation set. 
This procedure is designed to be consistent with the accumulative nature of time series data to avoid leaking information from the future. 

More precisely, in our validation procedure, we generate $t-20$ folds of training and validation pairs $\{(training_n, validation_n): \forall n \in \{1,2,\dotsc,t-20\}\}$, where the $n$-th fold consists of training data from day $n$ to $n+13$: $training_n := \{(s, \ln(\bm{I}_{s,c})): s \in \{n,n+1,\dotsc,n+13\}, c \in C\}$ and validation ground truth of all counties on the $n + 20$-th day: $validation_n:=\{(s, \ln(\bm{I}_{s,c})): s \in \{n+20\}, c \in C\}$.
A detailed illustration of the training and validation data used in each fold when choosing the best $\delta_{t}$ is presented in Table \ref{tab:cross_validation_data_tcv}.

\begin{table}[ht]
\centering
\begin{tabular}{|c|c|c|}
\hline
Fold Index & Training Data & Validation Data \\
\hline
1 & $\{(1, \ln(\bm{I}_{1,c})), \ldots, (14, \ln(\bm{I}_{14,c}))| \forall c \in C\}$ & $\{(21, \ln(\bm{I}_{21,c})) | \forall c \in C\}$ \\
2 & $\{(2, \ln(\bm{I}_{2,c})), \ldots, (15, \ln(\bm{I}_{14,c}))| \forall c \in C\}$ & $\{(22, \ln(\bm{I}_{22,c})) | \forall c \in C\}$ \\
$\vdots$ & $\vdots$ & $\vdots$ \\
$n$ & $\{(n, \ln(\bm{I}_{n,c})), \ldots, (n+13, \ln(\bm{I}_{n+13,c}))| \forall c \in C\}$ & $\{(n+20, \ln(\bm{I}_{n+20,c})) | \forall c \in C\}$ \\
$\vdots$ & $\vdots$ & $\vdots$ \\
$t-20$ & $\{(t-20, \ln(\bm{I}_{t-20,c})), \ldots, (t-7, \ln(\bm{I}_{t-7,c}))| \forall c \in C\}$ & $\{(t, \ln(\bm{I}_{t,c})) | \forall c \in C\}$ \\
\hline
\end{tabular}
\caption{Training and validation data of each fold for selecting hyperparameter $\delta_t \in \{2,\dotsc,14\}$ for the \texttt{tcv} estimator by cross-validation}
\label{tab:cross_validation_data_tcv}
\end{table}

For each of the folds, we then train the \texttt{tcv} estimators using candidate window sizes $\{2,\dotsc,14\}$, forecast the incident case numbers of each county in the fold, and assess the validation error of each hyperparameter choice in each fold.
The validation errors for each choice of the hyperparameter are then averaged across all folds. 
The hyperparameter choice with the lowest validation error is assigned to $\delta_t$ for test time evaluation (forecasting $\ln(\bm{I}_{t+7,c})$).
This cross validation procedure and test-time evaluation is repeated for all $\delta_{t}: t \in \{21,\dotsc, T\}$.\\

\subsubsection{ctcv}\label{apd1.ctcv}
For each county $c$, the \texttt{ctcv} estimator (\ref{ctcv}) can be directly computed from the case incidence data $\{(s, \ln({I}_{s,c})) : \forall s \in \{t-\delta_{t,c}+1,\dotsc,t\}\}$ for any fitting window size $\delta_{t,c}$ hyperparameter, in the following fashion:
\begin{equation}\label{ctcv_regression_form}
    \bm{\hat{r}^{ols(\delta_{t,c})}_{t,c}}=\frac{\sum_{s=t-\delta_{t,c}+1}^{t}(s-\frac{\sum_{s=t-\delta_{t,c}+1}^{t}s}{\delta_{t,c}})(\ln(\bm{I_{s,c}}) - \frac{\sum_{s=t-\delta_{t,c}+1}^{t}\ln(\bm{I_{s,c}})}{\delta_{t,c}})}{\sum_{s=t-\delta_{t,c}+1}^{t}\left(s-\frac{\sum_{s=t-\delta_{t,c}+1}^{t}s}{\delta_{t,c}}\right)^2}.
\end{equation} 
This encompasses the incidence case data of the county $c$ from the beginning of the chosen fitting window ($t-\delta_{t,c}+1$) to the current time of query ($t$).
In contrast to \texttt{tcv}, the key characteristic of the \texttt{ctcv} estimator is that at each time point $t$, all the counties are free to use their own fitting window size hyperparameter $\delta_{t,c}$.
We will now describe the cross-validation procedure to obtain the best $\delta_{t,c}$, and how the final test-time estimator is computed and evaluated.

The procedure for selecting the best hyperparameter at each time point is done independently for each county $c$. 
The test time evaluation for forecasting $\ln(\bm{I}_{t+7,c})$ is consistent with \texttt{tcv} as described in Appendix \ref{apd1.tcv}.
Therefore, for each county $c$ at time point $t$, we will generate $t-20$ folds of training and validation pairs as per Table \ref{tab:cross_validation_data_ctcv}.

\begin{table}[ht]
\centering
\begin{tabular}{|c|c|c|}
\hline
Fold Index & Training Data & Validation Data \\
\hline
1 & $\{(1, \ln(\bm{I}_{1,c})), \ldots, (14, \ln(\bm{I}_{14,c}))\}$ & $\{(21, \ln(\bm{I}_{21,c})) \}$ \\
2 & $\{(2, \ln(\bm{I}_{2,c})), \ldots, (15, \ln(\bm{I}_{14,c}))\}$ & $\{(22, \ln(\bm{I}_{22,c})) \}$ \\
$\vdots$ & $\vdots$ & $\vdots$ \\
$n$ & $\{(n, \ln(\bm{I}_{n,c})), \ldots, (n+13, \ln(\bm{I}_{n+13,c}))\}$ & $\{(n+20, \ln(\bm{I}_{n+20,c})) \}$ \\
$\vdots$ & $\vdots$ & $\vdots$ \\
$t-20$ & $\{(t-20, \ln(\bm{I}_{t-20,c})), \ldots, (t-7, \ln(\bm{I}_{t-7,c}))\}$ & $\{(t, \ln(\bm{I}_{t,c})) \}$ \\
\hline
\end{tabular}
\caption{Training and validation data of each fold for selecting hyperparameter $\delta_{t,c} \in \{2,\dotsc,14\}$ for the \texttt{ctcv} estimator by cross-validation}
\label{tab:cross_validation_data_ctcv}
\end{table}
The hyperparameter selection process, followed by the test time evaluation (forecasting $\ln(\bm{I}_{t+7,c})$), will proceed similarly to the rolling window time series cross-validation procedure described in Appendix \ref{apd1.tcv}. The only difference is that each county $c$ will now choose its own best $\delta_{t,c}$ independently.

%\subsubsection{TLRF vs Dynamic Window Comparison}\label{supplementary_tcvctcv}
%This subsection contains the additional RMSE plots of the Dynamic Window Size estimators in %\S\ref{subsec.flexible}. A review of Figures \ref{fig:updated_tcv_ctcv_mae_appendix} and %\ref{fig:updated_tcv_ctcv_rmse_appendix} reveals that the outperformance of \texttt{TLRF} over %\texttt{tcv} and \texttt{ctcv} is consistent across days throughout the study period considered.
%\input{Plots/updated_tcv_ctcv_mae_appendix}
%\input{Plots/updated_tcv_ctcv_rmse_appendix}

\subsubsection{K-Means Dynamic Window}\label{apd1.kdw}
The K-Means Dynamic Window method obtains its estimator in two steps. 
First, utilizing spatial information (e.g., HSA regions, Census information, Proportion of vulnerable populations), it conducts K-Means clustering on all counties in the U.S., where $K$ is the total number of clusters.
Second, once the $K$ clusters $\{C_{1},C_{2},\dotsc, C_{K}\}$ have been generated,  
it chooses a cluster-specific fitting window for counties within the same cluster via time-based cross-validation.

We will first describe the clustering process in detail.
We then describe the hyperparameter selection process for the dynamic window estimators within a given cluster.

\textbf{Clustering Implementation}\\
The goal of the clustering step is to partition all the counties of the US into $K \in \{100, 200,\dotsc,\}$ clusters using geographic (i.e., time-invariant) information from our datasets. 
We first determined at the dataset level what features are time-invariant within our study:
\begin{enumerate}
    \item The 2019 United States Census Gazetteer Files \citep[c.f.][]{USCensus2019}. Sample features included are:
    \begin{itemize}
        \item Geographic Centers (Latitude, Longitude) of Counties
    \end{itemize}
    \item The Centers for Disease Control and Prevention Social Vulnerability Index 2018 Database \citep[c.f.][]{CDCSVI2018}. Sample features included are:
    \begin{itemize}
        \item Population below the poverty line
        \item Unemployment Rate
        \item Proportion of Elderly ($>65$ years of age)
    \end{itemize}
    \item  The CDC SARS-CoV2 Variant Proportions Dataset \citep[c.f.][]{cdc_variants}. We only utilize the information of which county belongs to which Health and Human Services (HHS) Region.
\end{enumerate}
The $K$ clusters are then randomly initialized using the \texttt{k-means$++$} procedure by \cite{1283494}, where the clustering procedure was run for at most $10,000$ iterations. 

This parallel implementation can be found on our GitHub repo: \url{https://github.com/wangzilongri/COVID-tracking/tree/TLGRF_major_revision/analysis/benchmark_tcv_kmeans_code}

\textbf{Dynamic Window Selection per Cluster}\\
For a given choice of $K$, once we have obtained the $K$ clusters $C_1,\dotsc,C_K$, we can proceed with fitting the dynamic window estimator within each cluster in a similar fashion to the previously described estimators \texttt{tcv,ctcv}.
For each county $c$ in the $k$-th cluster $C_{k}$, at time point $t$, the K-Means Dynamic Window estimator (\ref{kmeanstcv}) can be directly computed from the case incidence data $\{(s, \ln({I}_{s,c})) : \forall s \in \{t-\delta_{t,k}+1,\dotsc,t\}, \forall c \in C_{k}\}$ for any fitting window size hyperparameter $\delta_{t,k}$, in the following fashion:
\begin{equation}\label{kmeans_regression_form}
    \bm{\hat{r}^{ols(\delta_{t,k})}_{t,k}}=\frac{\sum_{s=t-\delta_{t,k}+1}^{t}(s-\frac{\sum_{s=t-\delta_{t,k}+1}^{t}s}{\delta_{t,k}})(\ln(\bm{I_{s,c}}) - \frac{\sum_{s=t-\delta_{t,k}+1}^{t}\ln(\bm{I_{s,c}})}{\delta_{t,k}})}{\sum_{s=t-\delta_{t,k}+1}^{t}\left(s-\frac{\sum_{s=t-\delta_{t,k}+1}^{t}s}{\delta_{t,k}}\right)^2}.
\end{equation} 

%We now proceed to describe the test time evaluation, and cross validation procedure to select the best $\delta_{t,k}$ hyperparameter for each county in the $k$-th cluster at time $t$.

The procedure for selecting the best hyperparameter at each time point is done independently for each cluster $C_k$. 
The test time evaluation for forecasting $\ln(\bm{I}_{t+7,c})$ is consistent with \texttt{tcv} and \texttt{ctcv} as described in Appendix \ref{apd1.tcv} and Appendix \ref{apd1.ctcv} respectively.
Each cluster $C_k$ independently selects its best hyperparameter $\delta_{t,k}$ at every time point $t$, which will be shared for all counties $c \in C_k$.
Therefore, at time point $t$, for each cluster $C_k$, we will generate $t-20$ folds of training and validation pairs as per Table \ref{tab:cross_validation_data_kmeans}.
Once each cluster obtains its hyperparameter choices from the rolling window time series cross-validation procedure, we will proceed with the test time evaluation of forecasting $\ln(\bm{I}_{t+7,c})$ for all time points and all counties across all clusters.\par

Notably, this estimator reduces to the \texttt{tcv} estimator (\ref{tcv})
if its fitting window size is the same for all U.S. counties each day $t$, i.e., $\delta_{t,k}=\delta_{t}$, by assigning all counties to the same cluster. 
In contrast, this estimator becomes the \texttt{ctcv} estimator (\ref{ctcv}) if each county is assigned to its own distinct cluster, i.e., $\delta_{t,k}=\delta_{t,c}$.

\begin{table}[ht]
\centering
\begin{tabular}{|c|c|c|}
\hline
Fold Index & Training Data & Validation Data \\
\hline
1 & $\{(1, \ln(\bm{I}_{1,c})), \ldots, (14, \ln(\bm{I}_{14,c}))| \forall c \in C_k\}$ & $\{(21, \ln(\bm{I}_{21,c})) | \forall c \in C_k\}$ \\
2 & $\{(2, \ln(\bm{I}_{2,c})), \ldots, (15, \ln(\bm{I}_{14,c}))| \forall c \in C_k\}$ & $\{(22, \ln(\bm{I}_{22,c})) | \forall c \in C_k\}$ \\
$\vdots$ & $\vdots$ & $\vdots$ \\
$n$ & $\{(n, \ln(\bm{I}_{n,c})), \ldots, (n+13, \ln(\bm{I}_{n+13,c}))| \forall c \in C_k\}$ & $\{(n+20, \ln(\bm{I}_{n+20,c})) | \forall c \in C_k\}$ \\
$\vdots$ & $\vdots$ & $\vdots$ \\
$t-20$ & $\{(t-20, \ln(\bm{I}_{t-20,c})), \ldots, (t-7, \ln(\bm{I}_{t-7,c}))| \forall c \in C_k\}$ & $\{(t, \ln(\bm{I}_{t,c})) | \forall c \in C_k\}$ \\
\hline
\end{tabular}
\caption{Training and validation data of each fold for selecting hyperparameter $\delta_{t,k} \in \{2,\dotsc,14\}$ for all counties in the $k$-th cluster at time $t$}
\label{tab:cross_validation_data_kmeans}
\end{table}

The parallel implementation of this procedure can be found on our GitHub repository at: \url{https://github.com/wangzilongri/COVID-tracking/tree/TLGRF_major_revision/analysis/benchmark_tcv_kmeans_code/cumsum_code}.

\subsection{LASSO Based Transfer Learner}\label{apd:LASSOTL}
In this subsection, we provide the implementation details of the \texttt{LASSOTL} benchmark derived from \cite{bastani2021predicting} along with additional performance benchmarks in comparison to \texttt{TLRF}.
 
We utilize the identification result from Corollary (\ref{OLSTradeoff}): where the county-level instantaneous exponential growth rate ${r_{t,c}}$ is identified as 
\[
{r_{t,c}}=\mathbb{E}[\ln(\bm{I_{t,c}})-\ln(\bm{I_{t-1,c}})|\Vec{X}_{t,c}].
\]
As such, by denoting 
$${Y_{gold}}=\ln({I_{t,c}})-\ln({I_{t-1,c}}),\ \ \Vec{X}_{gold}=\Vec{X}_{t,c},$$
and
$${\vec{Y}_{proxy}}=[\ln({I_{t',c'}})-\ln({I_{t'-1,c'}})]_{(t',c')\neq (t,c)},\ \ \mathcal{X}_{proxy}=[\Vec{X}_{t',c'}]_{(t',c')\neq (t,c)},$$
we implemented the \texttt{LASSOTL} estimator in two stages:
\begin{align*}
\mathbf{Step\ 1:}&\ \text{given regularization penalty parameter $\mu$,}\\
&\hat{\vec{\beta}}_{proxy} = \text{arg}\min_{\vec{\beta}} \left\{
    \frac{1}{t|C|-1} \left\| {\vec{Y}_{proxy}} - \mathcal{X}^T_{proxy} \vec{\beta} \right\|_2^2 + \mu\|\vec{\beta}\|_{1}
\right\} \\
\mathbf{Step\ 2:}&\ \text{given regularization penalty parameter $\lambda$,}\\
&\hat{\vec{\beta}}_{joint} = \text{arg}\min_{\vec{\beta}} \left\{
    \left\| {Y_{gold}} - \Vec{X}^T_{gold} \vec{\beta} \right\|_2^2
    + \lambda \left\| \vec{\beta} - \hat{\vec{\beta}}_{proxy} \right\|_1
\right\},
\end{align*}
and obtain the \texttt{LASSOTL} estimator for the exponential growth rate:
$${\hat{r}^{LASSOTL}_{t,c}}=\Vec{X}^T_{gold} \hat{\vec{\beta}}_{joint}.$$
This was then used to generate the 7 days ahead prediction of the log incident cases $\ln(\hat{I}^{LASSOTL}_{t,c})$ and evaluated for test-time accuracy via the RMSE and MAE metric to remain consistent with our benchmarking approaches.

For each query time $t$, our approach involves one Step 1 estimator and $|C|$ separate Step 2 estimators, where $C$ represents the set of counties or subgroups being studied. The hyperparameters tuned were the LASSO penalties $\mu$ for Step 1 and $\lambda$ for Step 2. To select the optimal hyperparameters, we performed grid searches over the range $\{10^{-5}, 10^{-4}, \dotsc, 10^{0}, 10^{1}, 10^{2}\}$. The tuning process was carried out independently for each step using 5-fold cross-validation to ensure robust performance. Both Step 1 and Step 2 were implemented using the \texttt{Python} library \texttt{scikit-learn} \citep{scikit-learn}, leveraging its \texttt{Lasso} and \texttt{GridSearchCV} functionalities for model fitting and hyperparameter optimization.

%\subsubsection{TLRF vs LASSOTL Comparison}\label{supplementary_LASSOTL}
%This subsection contains the additional MAE and RMSE plots of the \texttt{LASSOTL} estimator.

\begin{figure}[H]
\centering
\includegraphics[width=\textwidth]{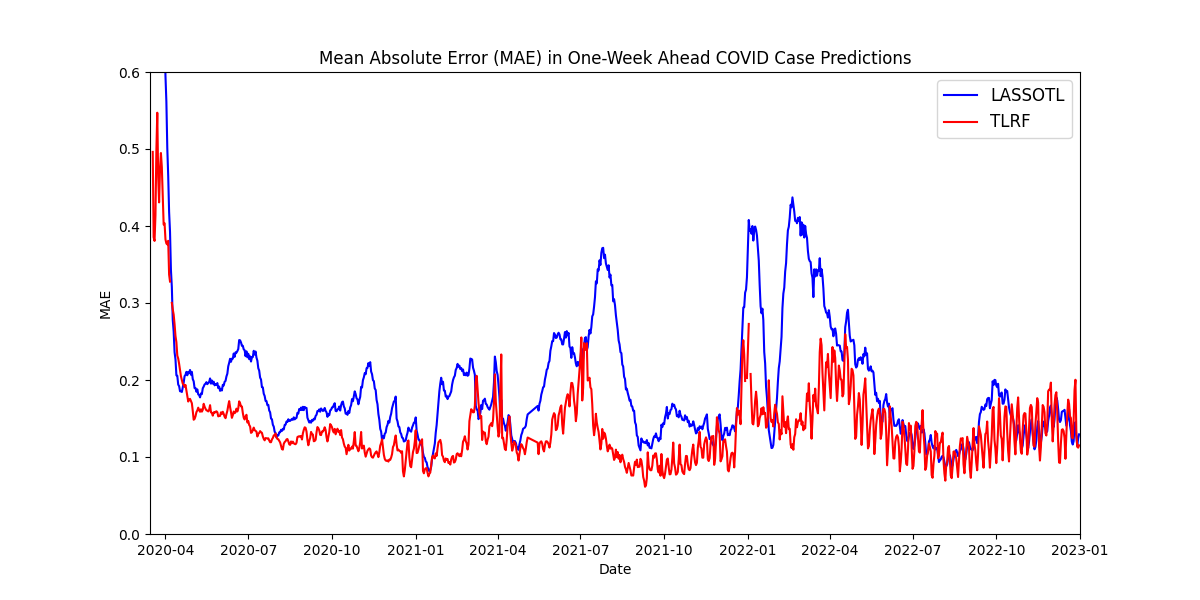}
    \caption{MAE plot of prediction accuracy (\texttt{TLRF} v.s. \texttt{LASSOTL})}
\label{fig:apd_Benchmark_Stage2_MAE}
\end{figure}
\begin{figure}[H]
\centering
\includegraphics[width=\textwidth]{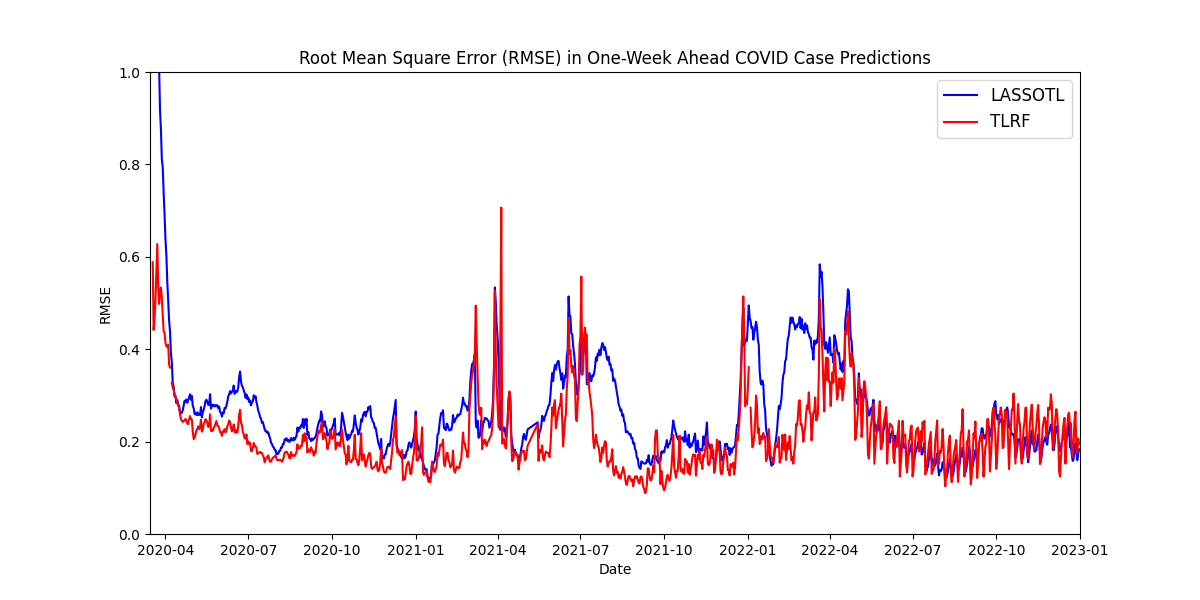}
    \caption{RMSE plot of prediction accuracy (\texttt{TLRF} v.s. \texttt{LASSOTL})}
\label{fig:apd_Benchmark_Stage2_RMSE}
\end{figure}

We also present supplementary results in the form of daily MAE and RMSE plots in Figures \ref{fig:apd_Benchmark_Stage2_MAE} and \ref{fig:apd_Benchmark_Stage2_RMSE} respectively, which demonstrate that \texttt{TLRF} outperforms \texttt{LASSOTL} for this use case.

The code for this benchmark can be found on our GitHub repository at: \url{https://github.com/wangzilongri/COVID-tracking/tree/TLGRF_major_revision/analysis/benchmark_transfer_learning}.

\section{Details Regarding Dataset}\label{apd:dataset}

This section provides an overview of the dataset used by \texttt{TLRF} in our study.
First, in Appendix \ref{apdC}, we describe the various datasets employed in our study.
Next, in Appendix \ref{apdA}, we explain the methodology used to define the incident case numbers, which serve as the dependent variable in our task.
Finally, in Appendix \ref{apd:feature_importance}, we introduce the concept of feature importance and discuss the key features identified by \texttt{TLRF}.

\subsection{Feature Data}\label{apdC}

All feature data we used, i.e., $\{\Vec{X}_{t,c}\}_{t\in \mathbb{Z}^+, c\in C}$, are publicly available online. In this section, we briefly describe what these datasets are and how we incorporated them in our work. 

%\subsection{New York Times COVID-19 Dataset}
%\cite{NYTimes2020} dataset is the primary panel dataset, where county level cumulative case numbers and deaths are updated every day from their first recorded appearance. 

\subsubsection{2019 US Census Gazetteer Files}
The 2019 United States Census Gazetteer Files \citep[c.f.][]{USCensus2019} were used to obtain the geographic locations (latitude-longitude centroids) of each county officially registered by the United States Census Bureau. This provides a spatial feature space for the random forest algorithm to further split upon.   

\subsubsection{Centers for Disease Control and Prevention Social Vulnerability Index 2018 Database}
The Social Vulnerability Index (SVI) database \citep[c.f.][]{CDCSVI2018} is a compilation of socio-economic factors such as unemployment rate, poverty rate, education attainment level etc. at the county level. These features are also included in our methodology.

\subsubsection{COVID-19 US state policy database (CUSP)}
The CUSP database \citep[c.f.][]{julia_2020} tracks when each state implemented and ended policies such as mask mandates, lockdowns, economic policies in response to the COVID-19 pandemic. As such, it is a vector of features capturing daily policy changes in each state. As these policies is state-wide, we naturally extend them to the respective county level.     

\subsubsection{The COVID Tracking Project }
From the COVID Tracking Project \citep[c.f.][]{COVIDTracking}, we obtained the daily numbers of PCR, Antibody and Antigen tests performed and their positivity rates at each state. These are used as features in our estimation algorithm as well. 

\subsubsection{SARS-CoV2 Variant Proportions}
This dataset provided by the CDC \citep[c.f.][]{CDC_COVID_Tracker} contains the mean estimate (along with its lower and upper estimation bounds) of the share of variants, e.g., ``B.1", ``P.1", ``XBB.1.5", ``others" of each Health and Human Services (HHS) region every week (or biweekly if indicated) starting from the 2nd of January 2021. Each date and region record can potentially have multiple published estimates, and we decided to take the latest published estimates (not later than the day of estimation). For data imputation, each county can be mapped to its HHS region, and for the days in between every reporting interval, we decided to forward fill the missing entries (i.e., for the days between week $T$ to week $T+1$, we will use the entries from week $T$). For the dates beyond the last reporting date, we also decided to forward fill. For the dates before the first record, we decided to backfill, i.e., for dates before the 2nd of January 2021, we will use the entries for the 2nd of January 2021. The variants' proportions are then normalized into percentages and joined with our current dataset to be used as features for \texttt{TLRF}

\subsection{COVID-19 Case Incidence}\label{apdA}

In epidemiology, case incidence, or incident case number, measures the number of new occurrences of a disease in a population over a specified period of time \citep{Incidence}. More precisely, the incident case number at county $c$ on day $t$ is defined as $I_{t,c}:=C_{t,c}-C_{t-\Delta t, c}\  \text{for}\  t > \Delta t,$ where $C_{t,c}$ and $C_{t-\Delta t, c}$ are the cumulative case number at county $c$ up to day $t$ and $t-\Delta t$, respectively. Case incidence describes how fast a disease spreads in a population, and thus is commonly used to guide policy decisions such as school closures \citep{Kittitas}.\par

Another useful measure in epidemiology is the active case number, which captures the number of still ``infectious" cases. More precisely, the active case number at county $c$ on day $t$ is defined as
\begin{equation}\label{acn}
    A_{t,c} := C_{t,c} - D_{t,c} - R_{t,c},
\end{equation}
where $C_{t,c}$, $D_{t,c}$ and $R_{t,c}$ are the cumulative number of cases , deaths and recoveries at county $c$ up to day $t$. Clearly, calculating the active case number requires more information than calculating the incident case number, and thus may not always be feasible. In fact, we are unable to directly calculate the number of active cases as in (\ref{acn}), because
the number of recovered cases $R_{t}$ are not reported at county-level in the U.S.. In addition, we find that the cumulative number of case and death numbers are not always consistent in our dataset \citep[c.f.][]{NYTimes2020}, i.e., $\exists c\in C, t\in \mathbb{Z}^+ \ s.t.\  C_{t,c} < D_{t,c}$.\par

To better capture COVID-19 case growth trajectories,  we focus on analyzing the 22-day incident case number in this paper, i.e.
\begin{equation}
    I_{t,c} := \begin{cases}
                C_{t,c} - C_{t-22,c} &\text{, if } t \geq 23\\
                C_{t,c} &\text{, if }  1 \leq t < 23.
             \end{cases} \label{eq:3}
\end{equation} 
Here, defining the incident case using a period of 22 days was based on the early practices observed during the pandemic. Specifically, patients were recommended to undergo a self-quarantine period of 17-22 days to confirm recovery from COVID-19 \citep[c.f.][]{straits2021singapore}. Therefore, we chose the most stringent duration of 22 days to account for potential delays in resolving active cases. Notably, other studies also use a similar time to resolution of COVID-19 \citep[c.f.][]{MEL1}. As such, (\ref{eq:3}) measures how quickly COVID-19 spreads in a county. In addition, (\ref{eq:3}) provides a good proxy for the active case number (\ref{acn}) as well, assuming that a COVID-19 case takes at most 22 days to resolve. Lastly, we use the 7-day moving average to smooth out oscillations in case number reporting \citep[c.f.][]{bergman2020oscillations} and exclude $I_{t,c}$ observations with value below 20 in this study.

\subsection{Feature Importance}\label{apd:feature_importance}

\subsubsection{Feature Importance by Splitting Frequency and Depth}\label{apd:feature_importancec}
The native \texttt{variable\_importance} method of the \texttt{grf} package calculates the importance of features by tallying how often the individual trees split upon said feature weighted by how early they are split within each tree. The earlier the split, the more informative the feature is and hence it holds higher weight. 
More formally, if we wish to compute the feature importance of feature $f$, for each tree in the grf, we traverse the tree and check each node to see if said node split on $f$. The level of the node increments by 1 as we descend the tree, with the root node having a level of $0$. For every level we descend, we halve the importance of the contribution of the feature. This can be succinctly represented in equation (\ref{eqn:feature_importance}),

\begin{align}
    \texttt{Importance}(f) = \sum_{b \in B}\sum_{node \in b} \mathbb{1}\{\text{\texttt{node} split on $f$}\}^{-2\times \texttt{level(node)}}\label{eqn:feature_importance}
\end{align}
where $B$ is a forest consisting of tress $b$; each tree $b$ consists of nodes (representing splits); the level of the node is given by \texttt{level(node)}.

\subsubsection{Top 20 Most Important Features}\label{apd:feature_importance20}
We show the Top 20 features which we presented in Appendix \ref{subsec.feature} and explain what each feature means in descending order of importance.

For the purposes of our algorithm, if the feature is a date, we transformed it into a vector representing how many days have elapsed since the policy was implemented, with all days before it being $0$, and the start date itself being $1$. For example, if a county $c$'s feature $f$ date is $2020-02-05$ i.e., policy $f$ was implemented for that county on the 5th of May 2020, the feature for county $c$ would look like:
\[
    \begin{bmatrix}
    0,&\dotsc ,& 0 ,& \underbrace{1}_{2020-02-05} ,& 2 ,&3 ,&\dotsc
    \end{bmatrix} 
\]
Top 20 Features in Descending Order of Importance
\begin{enumerate}
    \item Stopped Personal Visitation in State Prisons
    \begin{itemize}
        \item The date a state stopped personal visitations to state prisons
    \end{itemize}
    \item State of emergency issued
    \begin{itemize}
        \item The date a state first issued any type of emergency declaration
    \end{itemize}
    \item Expanding Supplemental Nutrition Assistance Program (SNAP)
    \begin{itemize}
        \item The date a state was approved the use of a waiver to provide many SNAP households with emergency supplementary benefits up to the maximum benefit a household can receive 
    \end{itemize}
    \item{Closed K-12 Public Schools}
    \begin{itemize}
        \item The date states closed K-12 public schools
    \end{itemize}
    \item Closed Bars
    \begin{itemize}
        \item The date states closed bars statewide. Unless otherwise noted, bars are defined as establishments that derive more than 50 percent of gross revenue from the sales of alcoholic beverages
    \end{itemize}
    \item Expanding Medicaid benefits (1135 Waivers)
    \begin{itemize}
        \item The date a state used a 1135 waiver to modify or waive Medicaid requirements
    \end{itemize}
    \item Closed Gyms
    \begin{itemize}
        \item The date states closed gyms statewide.
    \end{itemize}
    \item Closed Restaurants Except Takeout
    \begin{itemize}
        \item The date when restaurants are closed for in person dining with the exception of takeout orders
    \end{itemize}
    \item Closed Other Non-Essential Businesses
    \begin{itemize}
        \item The date a state closed non-essential businesses statewide. Only included directives/orders
    \end{itemize}
    \item Allow/Expand Medicaid Telehealth Coverage
    \begin{itemize}
        \item The date a state expanded Telehealth coverge for Medicaid recipients
    \end{itemize}
    \item Closed Movie Theaters
    \begin{itemize}
        \item The date a state closed cinemas and theaters
    \end{itemize}
    \item Variant - B.1.617.2 
    \begin{itemize}
        \item Proportion of positive COVID test cases being of the `B.1.617.2' strain, updated bi-weekly by the CDC
    \end{itemize}
    \item Ratio of Positive Tests
    \begin{itemize}
        \item Ratio of Positive COVID-19 tests across a rolling 7 day average
    \end{itemize}
    \item Variant - Other
    \begin{itemize}
        \item Proportion of positive COVID test cases with undetermined/unrecorded strains, udpated bi-weekly by the CDC
    \end{itemize}
    \item Variant - AY.3
    \begin{itemize}
        \item Proportion of positive COVID test cases being of the `AY.3' strain, updated bi-weekly by the CDC
    \end{itemize}
    \item Date General Public Became Eligible for COVID-19 Vaccination
    \begin{itemize}
        \item The date in which a state made the general public eligible for COVID-19 vaccination.
    \end{itemize}
    \item Date Adults Ages 30+ Became Eligible for Covid-19 Vaccination
    \begin{itemize}
        \item The date in which a state made adults ages 30+ eligible for COVID-19 vaccination
    \end{itemize}
    \item Allowed Restaurants to Sell Takeout Alcohol
    \begin{itemize}
        \item The date when restaurants, not classified as bars by percentage of revenue, are allowed to sell takeout alcohol
    \end{itemize}
    \item Variant - BA.1.1
    \begin{itemize}
        \item Proportion of positive COVID test cases being of the `BA.1.1' strain, updated bi-weekly by the CDC
    \end{itemize}
    \item Date Adults Ages 80+ Became Eligible for Covid-19 Vaccination
    \begin{itemize}
        \item The date in which a state made adults ages 80+ eligible for COVID-19 vaccination
    \end{itemize}
    
\end{enumerate}

\subsubsection{Feature Importance and Transfer Learning}\label{subsec.feature}
%\zw{are the feature importance updated?}

As shown in Appendix \ref{apd:TLGRF_Analysis}, \texttt{TLRF} derives much of its predictive power from transfer learning. This finding suggests that instead of relying on possibly limited historical data of a county in isolation for training, we can pool together data of counties which experienced similar events to generate a more robust and accurate rate estimate. A remaining question though is ``what are the most important determinants of `similarity' across counties''. Motivated by this question, in this section, we analyze the most important features used by \texttt{TLRF} in determining similarity across time and space.\par

We discovered that \texttt{TLRF} assigns non-zero importance to 398 out of the 468 of features incorporated in our dataset. In other words, our algorithm uses at least 398 features to determine which proxy data should be pooled together with the target data for estimation purposes.\par

\begin{figure}[ht]
\centering  
\includegraphics[width=\textwidth]{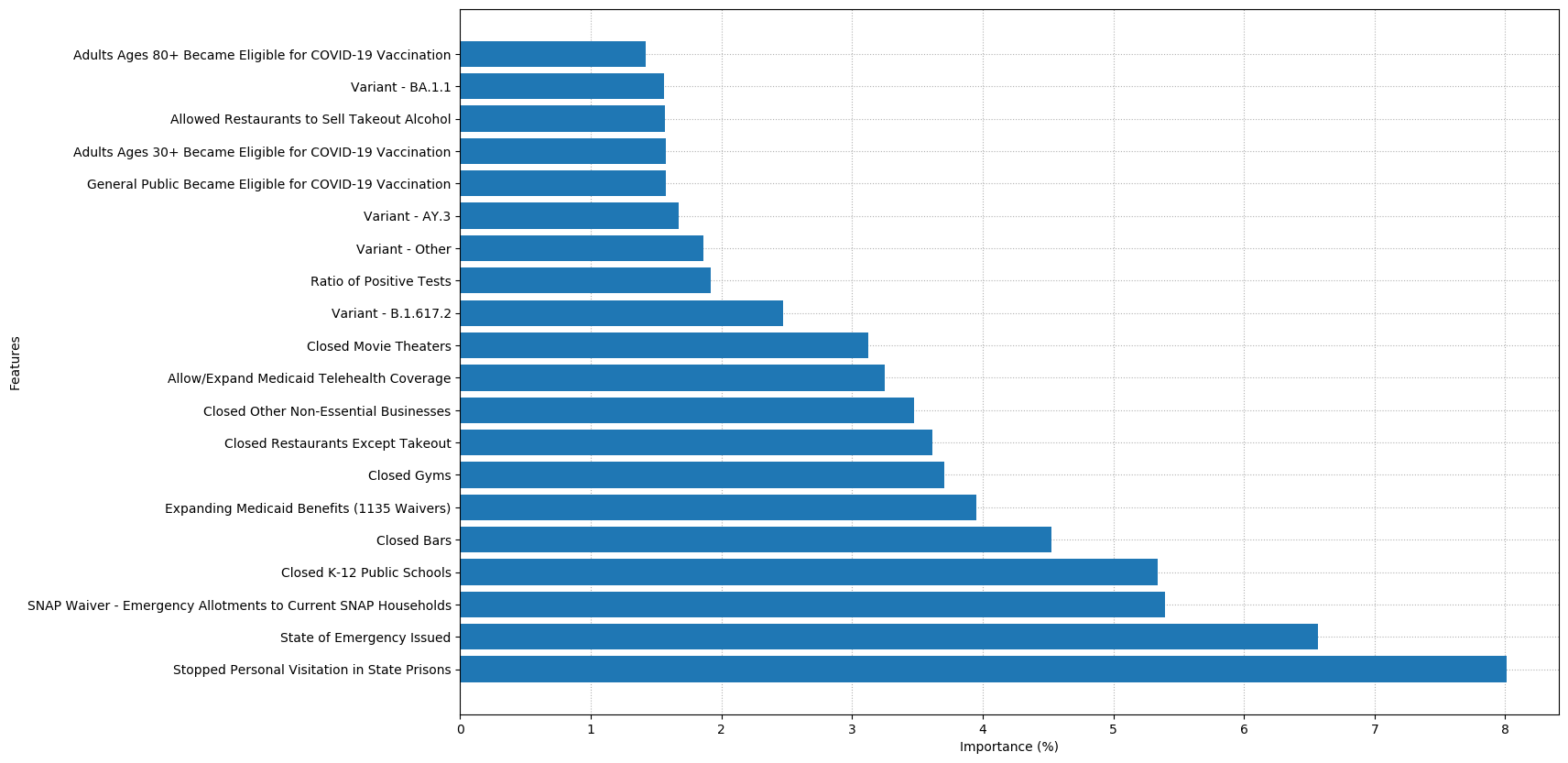}
\caption{Top 20 most important features of \texttt{TLRF} based on weighted split frequency }

\label{fig:GRF_TOP20}

\end{figure}

Figure \ref{fig:GRF_TOP20} presents the top 20 most important features used by \texttt{TLRF} in determining similarity in terms of COVID-19 spreading patterns. 
The top 20 most important features account for over $66.6\%$ of the total feature importance. Amongst them, we first observe that policies that aim to reduce social interactions (e.g., closing of bars, closing of restaurants, closing of non-essential businesses, stopping personal visitation to prisons, state of emergency issued, closing of K12 public schools etc.) collectively account for a total of $43.7$ out of $100$ percent of the weighted splits. Next, safety net policies (Expanding Supplementary Nutrition Benefits, Expanding Medicaid Benefits w/ 1135 Waiver, Expanding Teleheath Coverage) account for a total of $12.6\%$. We then see that the proportion of positive COVID-19 tests and various strains account for $9.49\%$.  Finally, we see that policies pertaining to the availability of vaccinations account for a total of $4.6\%$. These findings indicate that isolation policies, safety net policies, COVID-19 strain information, and vaccination availability provide the strongest signals in determining experienced growth rates and county similarity for our algorithm.

%\clearpage

\section{Additional Analyses for the CDPHE Case Study}\label{apd:Case_Study}
In this appendix subsection, we present additional information complementing our case study in \S\ref{CS}. 
First, in Appendix \ref{apd:Case_Study_DP}, we provide further details regarding the number of decision points and the counties investigated. 
Following that, in Appendix \ref{apd:Case_Study_populous}, we conduct a more in-depth analysis comparing the performance of \texttt{TLRF} and CDPHE. 
Then, in Appendix \ref{apd:Case_Study_threshold}, we consider an alternative use case of \texttt{TLRF} by considering a threshold based policy for allocating investigation resources.
Finally, in Appendix \ref{apd:coronaSEIR_R0}, we consider plugging in our instantaneous growth rate estimates from \texttt{TLRF} into a compartmental model and show that it can be used to improve incident case estimates in Colorado.
% examine the implications of adopting an absolute threshold based policy instead, in contrast to the relative rank-based threshold policy used by CDPHE and \texttt{TLRF}.

\subsection{Decision Points and Outbreak Investigations}\label{apd:Case_Study_DP}
\ \\
Our study period, which spans 2020-03-17 to 2022-12-31, has a total of 159 decision points (DPs) and 182 investigated counties.
A plot of how many investigations were made at each date is given below in Figure \ref{fig:CPDHE_capacity_appendix} and in Table \ref{table:CPDHE_capacity_appendix}.

\begin{figure}[H]
\centering
\resizebox{\linewidth}{!}{\includegraphics{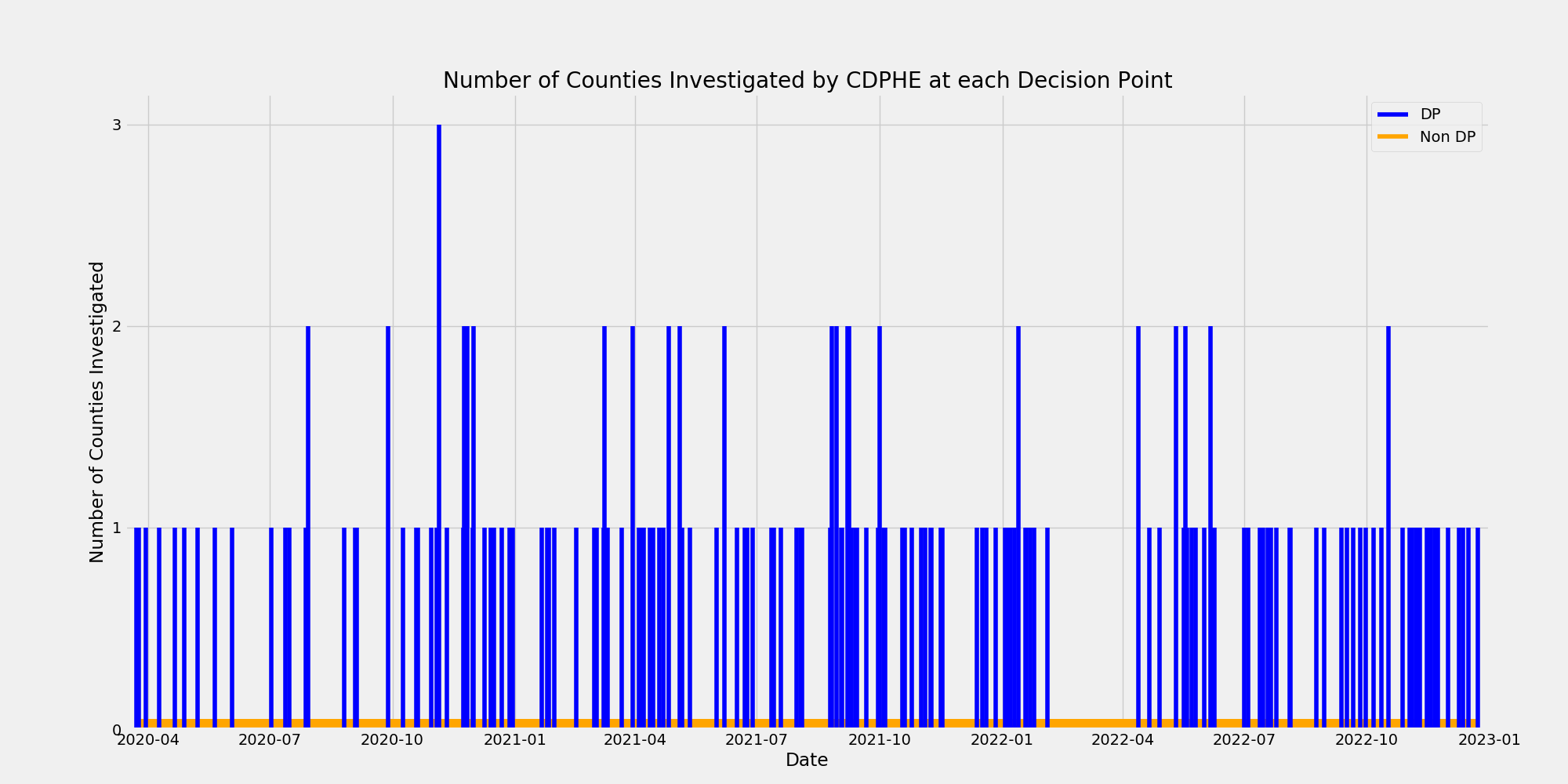}}
\caption{Daily Number of New Investigations Initiated by CDPHE}
\label{fig:CPDHE_capacity_appendix}
\end{figure}

\begin{table}[h]
\centering
\begin{tabular}{c|c|c|}
\cline{2-3}
& \textbf{\# of Days} & \textbf{Daily \# of New Investigations} \\
\cline{2-3}
Non-decision points & 861  & 0 \\
\cline{2-3}
\multirow{3}{*}{Decision points} & 137 & 1 \\
& 21 & 2 \\
& 1 & 3 \\
\cline{2-3}
\end{tabular}
\caption{Distribution of Daily Number of New Investigations Initiated by CDPHE}
\label{table:CPDHE_capacity_appendix}
\end{table}

\subsection{TLRF vs. CDPHE Further Analyses}\label{apd:Case_Study_populous}
\ \\
We further examine if there are systematic differences in counties chosen for investigations by CDPHE and \texttt{TLRF}.
We also investigate what types of counties are most frequently misclassified by the CDPHE and \texttt{TLRF}, respectively.

%if there is significant evidence to demonstrate that

%in response to the suggestion of weighting the outcomes by population

First, we compare CDPHE to \texttt{TLRF}'s performance in the more populous counties in Table \ref{tab:population_binary_appendix}. 
We observe that both the CDPHE and \texttt{TLRF} exhibit higher PPV and F1 Scores in this analysis when we focus on the more populous subset of counties in Colorado. 
However, \texttt{TLRF} still consistently outperforms CDPHE across all subsets and metrics.
We provide an intuitive explanation behind this.
As we explained \S\ref{CS}, the \texttt{TLRF} decision criterion is implicitly weighted by the population of counties as it prioritizes counties with high current reported incident case numbers.
Therefore, the \texttt{TLRF} algorithm is inherently capturing CDPHE's investigation prioritization of outbreaks in more populous counties, but in a more targeted manner.\\

% Next, to see if there are systematic differences in the counties investigated, we present the t-test conducted on the counties chosen by CDPHE and \texttt{TLRF} in Table \ref{tab:ranking_table_rev1}.
% We found that on average, there is no significant evidence of systematic differences in total population size between counties investigated by CDPHE and \texttt{TLRF}.
% We also found no systematic differences in statistics related to the at-risk populations that CDPHE focuses on, such as the number of nursing home beds and the population of incarcerated.\\

\begin{table}[H]
    \centering
    \begin{tabular}{cccccc}
        \hline
        \multirow{2}{*}{Most Populous Counties} & \multicolumn{2}{c}{CDPHE} & \multicolumn{2}{c}{\texttt{TLRF}} \\
        \cmidrule(lr){2-3} \cmidrule(lr){4-5}
         & PPV & F1 & PPV & F1 \\
        \midrule
        Top 10\% & 0.3333 & 0.3636 & 1.0000 & 1.0000 \\
        Top 20\%  & 0.3571 & 0.2020 & 0.6974 & 0.7211 \\
        Top 50\%  & 0.2185 & 0.2023 & 0.6268 & 0.6357 \\
        All & 0.1813 & 0.1813 & 0.5879 & 0.5879 \\
        \hline
    \end{tabular}
    \caption{Positive Predictive Value (PPV) and F1 Score of CDPHE and \texttt{TLRF} on the more populous counties of Colorado}
    \label{tab:population_binary_appendix}
\end{table}

% We also present the findings of our analysis on the misclassification of counties by both CDPHE and \texttt{TLRF} in Table \ref{tab:CDPHE_vs_TLGRF_vs_Overall_rev1}.
% Notably, our investigation revealed that the most commonly misclassified counties, on average, exhibit a lower total population size and other notable census statistics (e.g., individuals facing poverty, unemployment, or disabilities) in comparison to the broader demographic metrics of all Colorado counties.
% We also find significant overlap in the most commonly misclassified counties by the CDPHE and by \texttt{TLRF}.
% These findings suggest that CDPHE and \texttt{TLRF} have similar prioritization in terms of counties investigated.

Finally, Tables \ref{tab:CDPHE_Correct_Incorrect_Overall_rev1} and \ref{tab:TLGRF_Correct_Incorrect_Overall_rev1} present the summary statistics of counties most frequently misclassified by CDPHE and \texttt{TLRF}, respectively. Notably, our investigation reveals significant overlap in the commonly misclassified counties by CDPHE and \texttt{TLRF}. Specifically, the most frequently misclassified counties, on average, exhibit a lower total population size and other notable census statistics (e.g., individuals facing poverty, unemployment, or disabilities) in comparison to the state average. These findings suggest that CDPHE and \texttt{TLRF} share similar empirical patterns in terms of their investigation priorities.\par

\begin{table}[htbp]
\centering
\resizebox{\textwidth}{!}{
\begin{tabular}{llrrrrrrr}
\toprule
        &     &  Total Population &  Poverty &  Unemployed &   Age $> 65$ & Age $< 17$ &  Disabled &  Uninsured \\
Counties of: & Statistics &            &         &          &          &          &          &          \\
\midrule
\multirow{2}{*}{CDPHE Incorrect} & Mean & 35495.4 & 3572.0 & 839.3 & 5491.2 & 7386.2 & 3635.9 & 3265.0 \\
        & Median & 18325.0 & 2084.0 & 531.0 & 3459.0 & 3381.0 & 2332.0 & 1864.0 \\
\cline{1-9}
\multirow{2}{*}{CDPHE Correct} & Mean &  69929.6 & 8138.9 &  1602.4 &  9320.1 & 14571.1 &   7039.1 &    7057.1 \\
                          & Median &  27754.0 & 3296.0 &   736.0 &  5076.0 &  5845.0 &   2990.0 &    2414.5 \\
\cline{1-9}
\multirow{2}{*}{All Colorado} & Mean & 43319.7 & 4568.9 & 1037.7 & 6468.0 & 9004.7 & 4379.4 & 4226.2 \\
        & Median & 25162.0 & 3075.0 & 589.0 & 3689.0 & 5514.0 & 2939.0 & 2395.0 \\
\bottomrule
\end{tabular}
}
\caption{Comparison of Mean and Median statistics of the Top 10 most frequently misclassified and correctly classified Colorado counties by CDPHE}
\label{tab:CDPHE_Correct_Incorrect_Overall_rev1}
\end{table}

\begin{table}[htbp]
    \centering
    \resizebox{\textwidth}{!}{%
    \begin{tabular}{llrrrrrrr}
\toprule
                      &     &  Total Population &  Poverty &  Unemployed &   Age $> 65$ & Age $< 17$ &  Disabled &  Uninsured \\
Counties of: & Statistics &            &         &          &          &          &          &          \\
\midrule
\multirow{2}{*}{\texttt{TLRF} Incorrect} & Mean & 25699.0 & 2447.4 & 601.3 & 4073.5 & 5181.0 & 2570.6 & 2606.6 \\
        & Median & 17909.0 & 1890.0 & 500.0 & 3253.0 & 2954.0 & 2332.0 & 1762.0 \\
\cline{1-9}
\multirow{2}{*}{\texttt{TLRF} Correct} & Mean &  58924.5 & 6003.8 &  1338.8 &  8490.7 & 11784.9 &   5618.5 &    5637.7 \\
                          & Median &  55101.0 & 4609.0 &  1250.0 &  8634.0 & 10565.0 &   4854.0 &    4993.0 \\
\cline{1-9}
\multirow{2}{*}{All Colorado} & Mean & 43319.7 & 4568.9 & 1037.7 & 6468.0 & 9004.7 & 4379.4 & 4226.2 \\
        & Median & 25162.0 & 3075.0 & 589.0 & 3689.0 & 5514.0 & 2939.0 & 2395.0 \\
\bottomrule
\end{tabular}
}
\caption{Comparison of Mean and Median statistics of the Top 10 most frequently misclassified and correctly classified Colorado counties by \texttt{TLRF}}
\label{tab:TLGRF_Correct_Incorrect_Overall_rev1}
\end{table}

% \begin{table}[htbp]
% \centering
% \resizebox{\textwidth}{!}{%
% \begin{tabular}{llrrrrrrr}
% \toprule
%         &     &  Total Population &  Poverty &  Unemployed &   Age $> 65$ & Age $< 17$ &  Disabled &  Uninsured \\
% Counties of: & Statistics &            &         &          &          &          &          &          \\
% \midrule
% \multirow{2}{*}{CDPHE} & Mean & 35495.4 & 3572.0 & 839.3 & 5491.2 & 7386.2 & 3635.9 & 3265.0 \\
%         & Median & 18325.0 & 2084.0 & 531.0 & 3459.0 & 3381.0 & 2332.0 & 1864.0 \\
% \cline{1-9}
% \multirow{2}{*}{\texttt{TLRF}} & Mean & 25699.0 & 2447.4 & 601.3 & 4073.5 & 5181.0 & 2570.6 & 2606.6 \\
%         & Median & 17909.0 & 1890.0 & 500.0 & 3253.0 & 2954.0 & 2332.0 & 1762.0 \\
% \cline{1-9}
% \multirow{2}{*}{Overall} & Mean & 43319.7 & 4568.9 & 1037.7 & 6468.0 & 9004.7 & 4379.4 & 4226.2 \\
%         & Median & 25162.0 & 3075.0 & 589.0 & 3689.0 & 5514.0 & 2939.0 & 2395.0 \\
% \bottomrule
% \end{tabular}%
% }
% \caption{Comparison of Mean and Median statistics of the Top 10 most commonly misclassified counties by CDPHE and \texttt{TLRF} against Colorado counties in general (Overall) \zw{Replace this table with two tables}}
% \label{tab:CDPHE_vs_TLGRF_vs_Overall_rev1}
% \end{table}

\subsection{A Threshold Policy Derived from \texttt{TLRF}}\label{apd:Case_Study_threshold}\par
\ \\
This section delves into an alternative use case of \texttt{TLRF}, building on the use case discussed in \S\ref{CS}. Specifically, \texttt{TLRF} can be utilized to configure an absolute threshold policy. That is, instead of choosing a relative threshold as in \S\ref{CS}, we can select a fixed threshold. This absolute threshold policy would then recommend investigating all counties that have caseload increases beyond this threshold projected by \texttt{TLRF}. In other words, the less stringent or lower this predetermined threshold is, the more counties would be investigated under this policy. 
To quantify the number of investigations recommended by absolute threshold policies, we have experimented with different thresholds and plotted two of them in Figure \ref{fig:CDPHE_vs_ThresholdDP_vs_ThresholdAll_appendix}.\par

We now discuss the practical implications of implementing policies based on absolute thresholds. 
As shown in Figure \ref{fig:CDPHE_vs_ThresholdDP_vs_ThresholdAll_appendix}, it is difficult to integrate CDPHE's investigation resource constraint into absolute threshold-based policies. On one hand, when a lower threshold is chosen (e.g., Threshold=10), this policy would recommend a large daily investigation number that exceeds the capacity constraint of CDPHE. 
On the other hand, when a higher threshold is chosen (e.g., Threshold=30), this policy would recommend too few investigations on some days when investigation resources are available. 
In general, it can be challenging for an absolute threshold policy to match the investigation frequency of CDPHE. 
Therefore, recommendations made by an absolute threshold policy may not always be feasible for CDPHE.\\

\begin{figure}[H]
    \centering
    \resizebox{\linewidth}{!}{\includegraphics{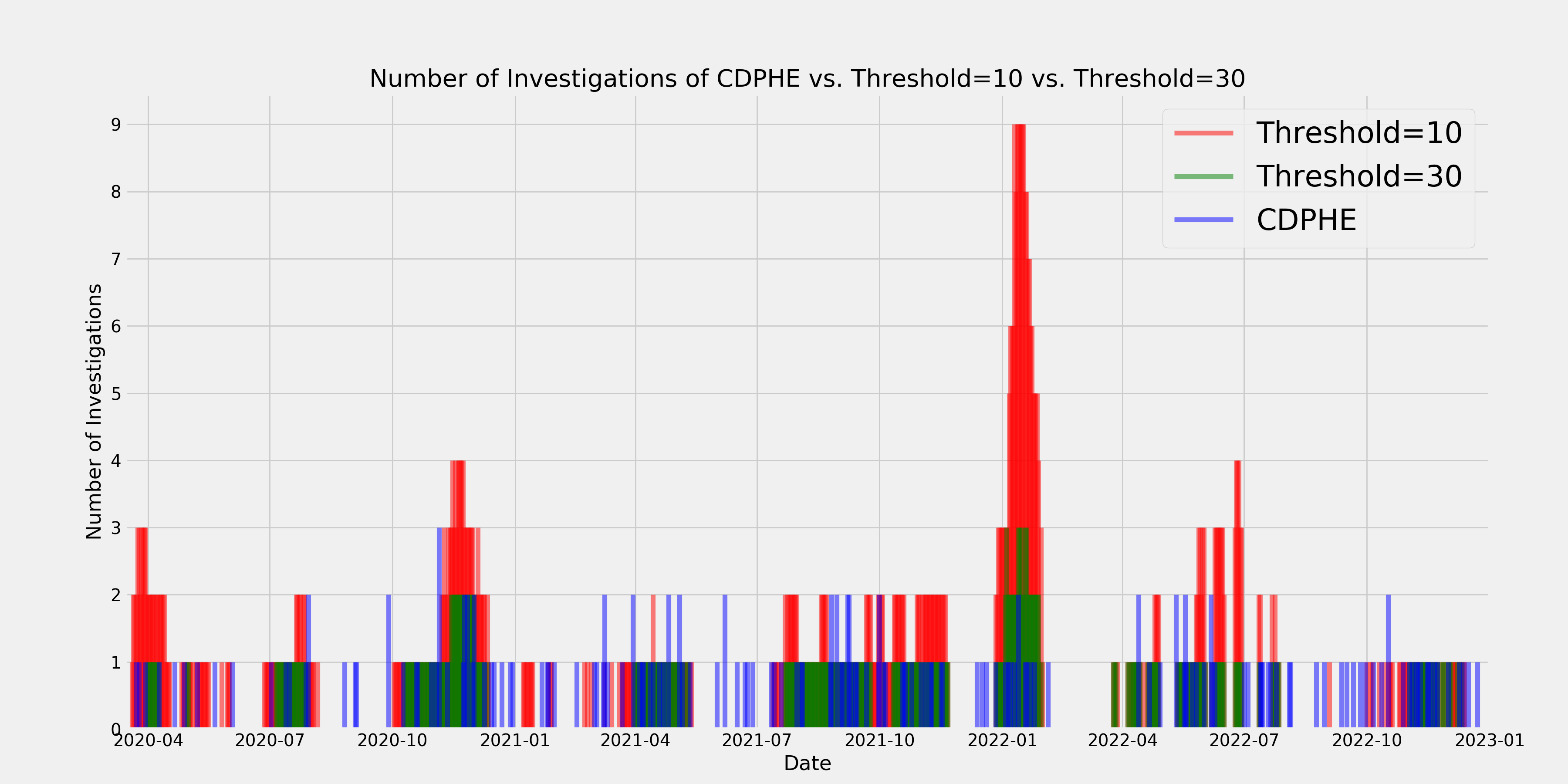}}
    \caption{Daily Number of New  Investigations Initiated by CDPHE\\ vs.\\ 
 Daily Number of New  Investigations Recommended by the Absolute Threshold Policy with Threshold=10 and Threshold=30}
    \label{fig:CDPHE_vs_ThresholdDP_vs_ThresholdAll_appendix}
\end{figure}

% We next evaluated the PPV, TNR, and F1 scores of both variations of the threshold policy and report their performance in Table \ref{tab:threshold_perforamnce_appendix}. 

% \begin{table}[!htpb]
% \centering
% \begin{tabular}{|c|c|c|}
% \hline
% \textbf{Metric} & \textbf{Threshold: all dates} & \textbf{Threshold: DPs only} \\
% \hline
% PPV & 0.749 & 0.775  \\
% TNR & 0.972 & 0.978 \\
% F1 & 0.749 & 0.775  \\
% \hline
% \end{tabular}
% \caption{Performance of variations of the threshold policy. Note that we should not compare them directly due to both having different decision sets.}
% \label{tab:threshold_perforamnce_appendix}
% \end{table}

% While we do note that it is not possible to do a proper apples to apples comparison against CDPHE's historical decisions due to different numbers of counties investigated at each step and even between each other due not having the same decision sets, the threshold policy does appear to work reasonably well given their high F1 scores and number of investigation resources allocated as per Table \ref{tab:Threshold_Table_appendix} do not look out of place given CDPHE's historical decisions (c.f. Table \ref{table:CPDHE_capacity_appendix}).  

\subsection{Using TLRF to Improve SEIR Estimates of Colorado Counties}\label{apd:coronaSEIR_R0}
In this subsection, we integrate our growth rate estimates into established epidemiological models like that of the SEIR framework.

We implemented an SEIR model with the following structure:
\begin{align*}
\frac{dS}{dt} &= -\frac{\beta S I}{N}\\
\frac{dE}{dt} &= \frac{\beta S I}{N} - \sigma E\\
\frac{dI}{dt} &= \sigma E - \gamma I\\
\frac{dR}{dt} &= \gamma I, \numberthis \label{R2RFP}
\end{align*}
where $S, E, I,$ and $R$ represent Susceptible, Exposed, Infectious, and Recovered populations, respectively, and $N$ represents the total population, which is the sum of the previous four terms. Following \cite{ma2020estimating}, the exponential growth rate $r$ relates to model parameters of (\ref{R2RFP}) through
\begin{align}
    r &= \frac{-(\sigma + \gamma) + \sqrt{(\sigma - \gamma)^{2} + 4\sigma\beta}}{2},
\intertext{which implies that}
     \beta &= \frac{(2r + (\sigma + \gamma))^{2} - (\sigma - \gamma)^2}{4\sigma} \label{eq:SEIR_beta}
\end{align}
where $\beta$ is the transmission rate; $\sigma$ is the incubation rate; $\gamma$ is the recovery rate. This relation allows us to incorporate our \texttt{TLRF} growth rate estimates into the SEIR framework. 
As we have independently estimated the exponential growth rate $r_{t,c}$ for each time $t$ and county $c$. Equation (\ref{eq:SEIR_beta}) allows us to reduce the free parameters to $\sigma$ and $\gamma$, which we estimate non-parametrically through a data-driven method described below.  

In this setup, the parameters $(\beta_{t,c}, \sigma_{t,c}, \gamma_{t,c})$ are allowed to vary over time and across counties, with $\beta_{t,c}$ being derived from Equation (\ref{eq:SEIR_beta}). To manage computational costs, we constrain $(\sigma_{t,c}, \gamma_{t,c})$ to remain constant within each county $c$ for a given calendar month. We search for the optimal values of $\sigma_{t,c}$ and $\gamma_{t,c}$ from the set $\{0.01, 0.02, \dots, 0.99, 1.0\}$, while allowing $\beta_{t,c}$ to vary on a daily basis.

Our benchmark criterion remains consistent: we aim to forecast the log of incident case numbers, $\ln(I_{t+7,c})$, seven days ahead for counties in Colorado. The study period spans from April 1, 2020, to December 31, 2022.

For each county in Colorado, we generate a separate SEIR trajectory to reflect the local dynamics of disease spread. To avoid data leakage from future time points, we use an iterative procedure based on retrospective data to select the optimal parameters $(\sigma_{t,c}, \gamma_{t,c})$:

\begin{enumerate}
    \item \textbf{Validation Phase:} For month $T$, we search for the best values of $(\sigma_{t,c}, \gamma_{t,c})$ that minimize the root mean squared error (RMSE) between the SEIR model’s daily incident case numbers $I_{t,c}$ and the actual observed values.
    \item \textbf{Prediction Phase:} To make forecasts for month $T+1$, we fix the parameter estimates $(\sigma_{t,c}, \gamma_{t,c})$ using the best-performing values from the validation phase in month $T$. At each time point $t$, we use the current state $(S_{t,c}, E_{t,c}, I_{t,c}, R_{t,c})$ and the fixed parameters $(\beta_{t,c}, \sigma_{t,c}, \gamma_{t,c})$ to generate the SEIR trajectory 7 days ahead. We then compare the predicted $\ln(I_{t,c})$ with the actual values using RMSE and MAE error metrics. This procedure is repeated for each subsequent month, leveraging temporal consistency while allowing for periodic adjustments to reflect evolving epidemiological conditions.
\end{enumerate}

To evaluate the impact of our method on SEIR model disease forecasting, we compare two versions of Equation (\ref{R2RFP}):
\begin{itemize}
    \item \texttt{SEIR-TLRF}: using our adaptive growth rates
    \item \texttt{SEIR-tcv}: using traditional time-based cross-validation
\end{itemize}
For both methods, we import the $r_{t,c}$ estimates generated by the benchmarks into the forecasting procedure, then derive $\beta_{t,c}$ via Equation (\ref{eq:SEIR_beta}). The results are presented in Table \ref{tab:seir_metrics} and Figures \ref{fig:apd_SEIR_MAE} and \ref{fig:apd_SEIR_RMSE}. The results thus show that \texttt{TLRF}'s growth rate estimates can be plugged into a compartmental model and improve its predictive performance. Specifically, our adaptive approach reduces the Median MAE by 12.7\% (from 0.395 to 0.345) and the RMSE by 13.6\% (from 0.537 to 0.464).
These improvements are substantial, especially considering that only about one-third of models evaluated in \cite{cramer2022evaluation} achieved a greater than 10\% improvement in relative MAE compared to a simple baseline model. 

\begin{table}[H]
\centering
\begin{tabular}{lcc}
\hline
\textbf{Method}  & \textbf{MAE} & \textbf{RMSE}\\
\hline
SEIR-\texttt{tcv}   & 0.395  & 0.537     \\
SEIR-\texttt{TLRF}    & \textbf{0.345}   & \textbf{0.464}    \\
\hline
\end{tabular}
\caption{Median MAE and RMSE of SEIR-\texttt{TLRF} v.s. SEIR-\texttt{tcv}}
\label{tab:seir_metrics}
\end{table}

\begin{figure}[H]
\centering
\includegraphics[width=\textwidth]{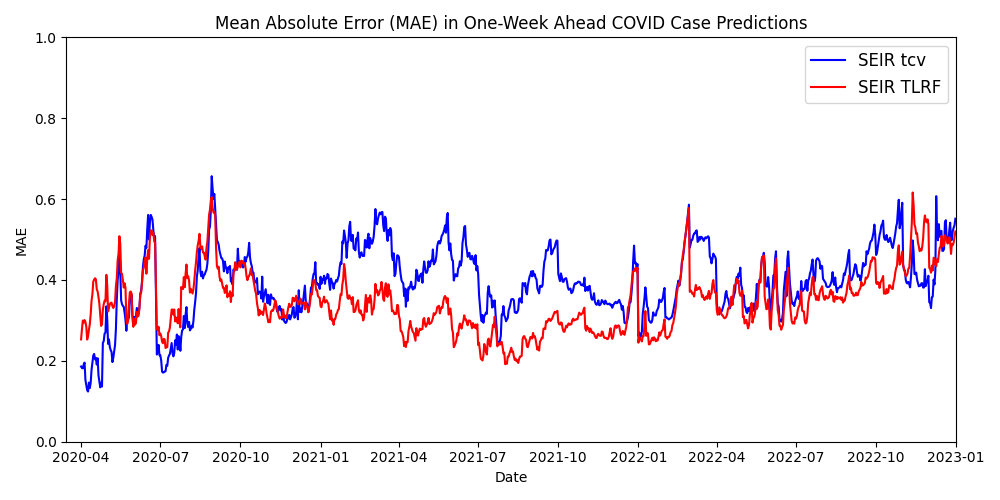}
\caption{ MAE plot of prediction accuracy (SEIR-\texttt{TLRF} v.s. SEIR-\texttt{tcv})}
\label{fig:apd_SEIR_MAE}
\end{figure}
\begin{figure}[H]
\centering
\includegraphics[width=\textwidth]{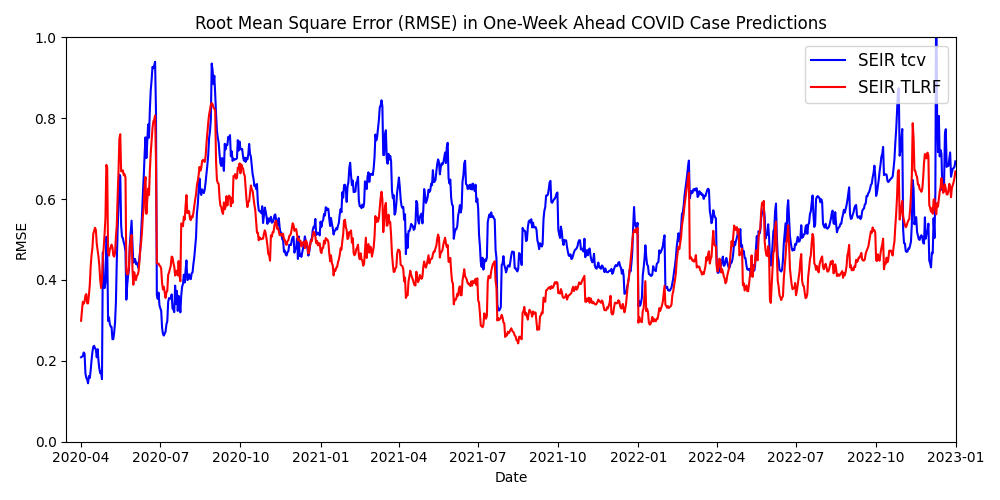}
\caption{ RMSE plot of prediction accuracy (SEIR-\texttt{TLRF} v.s. SEIR-\texttt{tcv})}
\label{fig:apd_SEIR_RMSE}
\end{figure}

The parallel implementation of this procedure can be found on our GitHub repository at: \url{https://github.com/wangzilongri/COVID-tracking/tree/TLGRF_major_revision/analysis/coronaSEIR}.

\section{COVID-19 Outbreak Detection Tool: Additional Details}\label{apd:covid19_outbreak_detection_appendix}
The outbreak detection tool reports county-level doubling times on a weekly basis throughout the course of the pandemic. Doubling time, defined as $\frac{\ln(2)}{\bm{r_{t,c}}}$, converts the instantaneous county-level growth rate $\bm{r_{t,c}}$ estimated from TLRF to an intuitive metric that is easier to explain to the general public. Specifically, doubling time captures the rate of spread of COVID-19 in each county: a longer doubling time means that COVID-19 is spreading slower, and vice versa. For example, Figure \ref{fig:chloroplast} shows the COVID-19 doubling times of each county in the U.S. during the Omicron wave in December 2021 to January 2022. Just before the spread of Omicron variant in early-December 2021, COVID-19 cases were declining, which is evident from the increase in the doubling time from December 05, 2021, to December 26, 2021. However, the Omicron variant started emerging in December end. By early January, the doubling time in most counties was less than 2 weeks. The latest COVID-19 doubling time in each county can be seen at \url{www.covid19sim.org}.
\begin{figure}[H]
\centering
\includegraphics[width=\linewidth]{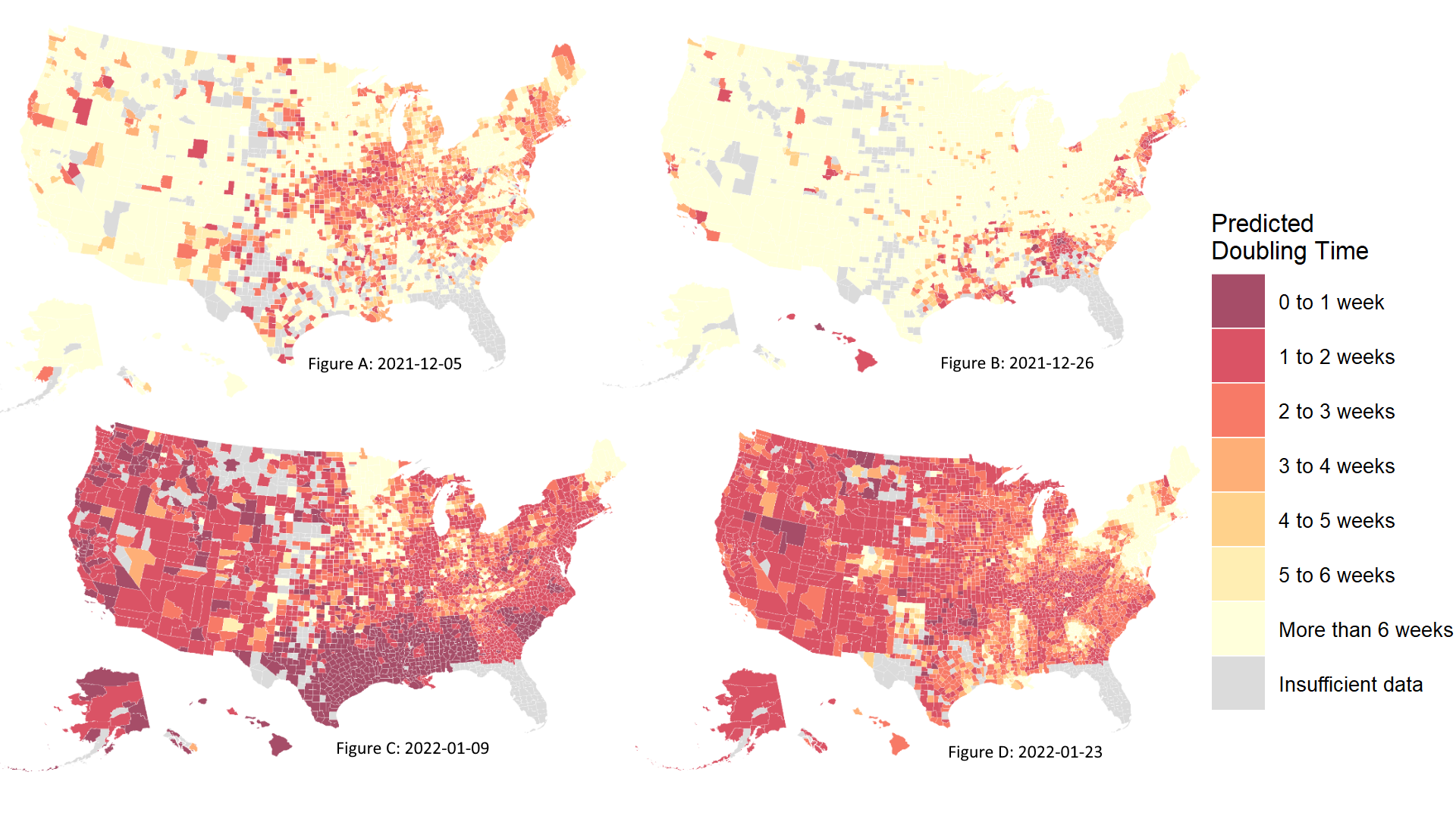}
\caption{Snapshots of the chloroplast of county-level doubling days on our website at various points in time from 2021-12-05 to 2022-01-23.
The doubling day of incident COVID-19 case number at county $c$ on day $t$ is derived from the instantaneous county-level growth rate $\bm{r_{t,c}}$ as $\frac{ln(2)}{\bm{r_{t,c}}}$.}
\label{fig:chloroplast}
\end{figure}
Furthermore, we feature a data explorer (see Figure \ref{fig:data_explorer}): an interactive spreadsheet of doubling days, recorded cases and deaths, vaccination adoption etc. for the more inquisitive members of the public, where they can filter, sort, and search through the data as desired and download it as a csv file. 
Detailed documentation is publicly accessible on \url{www.covid19sim.org} and all code and data are open source.
\begin{figure}[!h]
\centering
\includegraphics[width=\textwidth]{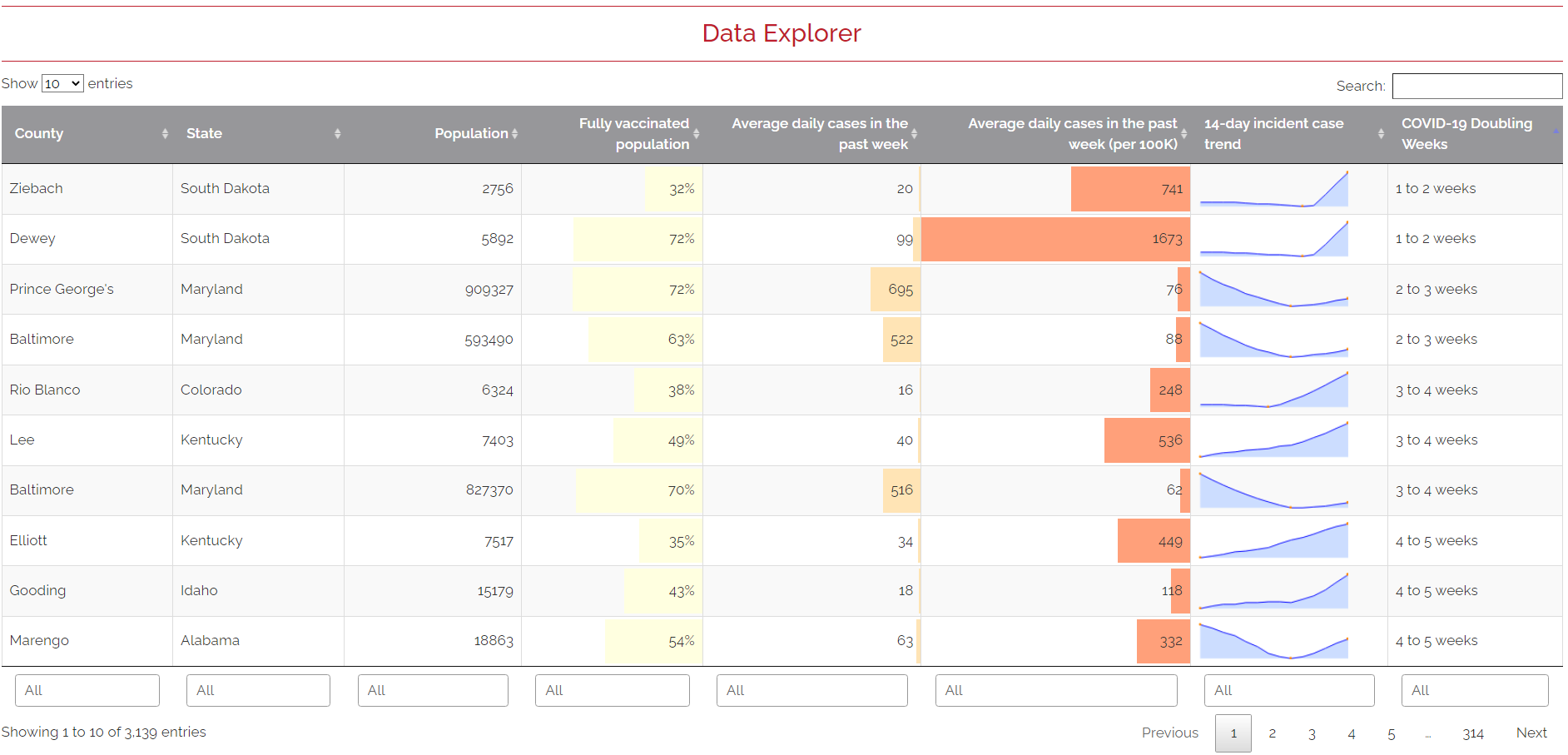}
\caption{Snapshot of the data explorer on our website, where the user can interactively filter, sort, and search through the relevant county level data and download it as a csv file if desired}
\label{fig:data_explorer}
\end{figure}
\par 

The web platform of this decision support tool, ``COVID-19 Outbreak Detection Tool", was released on the 15th of September 2020. 

Our outbreak detection tool has been well received since its launch. The website had around 105 daily visitors during the 2020 winter COVID-19 wave, where web traffic peaked at around 2000 visitors a day, and has an accumulated total of 20,000 authentic users from all 50 states. We have also been consulted by public health officials from the states of Maryland, Michigan, and Massachusetts, and helped them use this outbreak detection tool to guide their decision-making during the pandemic. Finally, our outbreak detection tool has also attracted significant public attention and was featured in leading media outlets, including The Voice of America and National Public Radio.

\end{APPENDICES}

\end{document}